%% file: _arxiv_main.tex
\documentclass{article}
\usepackage{arxiv1}

\usepackage[utf8]{inputenc}
\usepackage[T1]{fontenc}
\usepackage{lmodern}

\usepackage{graphicx}
\usepackage{subcaption}
\usepackage[colorlinks=true, linkcolor=blue, citecolor=blue, urlcolor=blue]{hyperref}
\usepackage{booktabs}
\usepackage{multirow}
\usepackage{makecell}
\usepackage{listings}
\usepackage{enumitem}
\usepackage{cite}
\usepackage{xcolor}
\usepackage{amssymb,amsmath}
\usepackage{mathabx}
\usepackage{CJKutf8}

\newcommand{\notforsubmission}[1]{}
\input{commands}

\renewcommand{\runningheader}{Fragile Preferences: Order Effects in LLMs}

\newif\ifpnas
\pnasfalse

\title{Fragile Preferences: A Deep Dive into Order Effects in Large Language Models}

\author{
  Haonan Yin\textsuperscript{1,*} \quad
  Shai Vardi\textsuperscript{2,*} \quad
  Vidyanand Choudhary\textsuperscript{3} \\
   \textsuperscript{1}Ivy College of Business, Iowa State University \\
  \texttt{haonany@iastate.edu}  \\
  \textsuperscript{2}Muma College of Business, University of South Florida \\
  \texttt{vardi@usf.edu} \\
  \textsuperscript{3}Paul Merage School of Business, University of California-Irvine \\
  \texttt{veecee@uci.edu}  \\
  \textsuperscript{*}\textit{These authors contributed equally.}
}

\date{}

\begin{document}
\maketitle

\begin{abstract}

\input{abstract}
\end{abstract}

\clearpage

\section{Introduction}
\input{_intro}

\section{Results}

\input{_results}

\section{Discussion}

\input{_discussion}

\section{Materials and Methods}
\label{sec:materials}
\input{_methods_pnas}

\bibliographystyle{unsrt}
\bibliography{pnas-sample}

\appendix

\input{_z_appendix_color_sets}
\clearpage
\input{_z_appendix_resume_generation}
\clearpage
\input{_z_appendix_pairwise_colors}

\clearpage

\input{_z_appendix_pairwise_resume}

\clearpage
\input{_z_appendix_true_from_tie_breaking}

\clearpage
\input{_z_appendix_triplewise_colors}

\clearpage
\input{_z_appendix_triplewise_resumes}
\clearpage
\input{_z_appendix_four_colors}
\clearpage
\input{_z_appendix_name}

\clearpage
\input{_z_appendix_gender}
\clearpage
\input{_z_appendix_additional_models}

\clearpage
\input{_z_appendix_additional_languages}

\end{document}

%% file: commands.tex
\newcommand{\robust}{\ggcurly}
\newcommand{\fragile}{\succcurlyeq}

\newcommand{\indiff}{\sim}  
\newcommand{\unsure}{\mathrel{\overset{\scriptscriptstyle ?}{\smash{\lower0.40ex\hbox{$\sim$}}}}}

\newcommand{\ignore}[1]{}

\newcommand{\Bestcolor}{Ideal}

\newcommand{\Goodcolor}{Fair}

\newcommand{\Medcolor}{Plain}
\newcommand{\Worstcolor}{Harsh}

\newcommand{\betweentableandcap}{\vspace{0.5em}}

\newcommand{\subsec}[1]{\ifpnas
\subsection*{#1}
\else
\subsection{#1}
\fi }

\newcommand{\subsubsec}[1]{\ifpnas
\subsubsection*{#1}
\else
\paragraph{#1.}
\fi }

%% file: abstract.tex
Large language models (LLMs) are increasingly deployed in decision-support systems for high-stakes domains such as hiring and university admissions, where choices often involve selecting among competing alternatives. While prior work has noted position biases in LLM-driven comparisons, these biases have not been systematically analyzed or linked to underlying preference structures.
We present the first comprehensive study of position biases across multiple LLMs and two distinct domains: resume comparisons, representing a realistic high-stakes context, and color selection, which isolates position effects by removing confounding factors. We find strong and consistent order effects, including a quality-dependent shift: when all options are high quality, models favor the first option, but when quality is lower, they favor later options. We also identify  a previously undocumented bias: a \emph{name bias}, where certain names are favored despite controlling for demographic signals.
To separate superficial tie-breaking from genuine distortions of judgment, we extend the rational choice framework to classify pairwise preferences as robust, fragile, or indifferent. Using this framework, we show that order effects can lead models to select strictly inferior options. These results indicate that LLMs exhibit distinct failure modes not documented in human decision-making. We also propose targeted mitigation strategies, including a novel use of the temperature parameter, to recover underlying preferences when order effects distort model behavior.

%% file: _intro.tex
\ifpnas{\dropcap{L}arge }\else{Large }\fi 
language models (LLMs) are becoming increasingly integrated into high-stakes decision-support systems. 
For instance, in healthcare, LLMs are being used to assist in differential diagnosis and support triage decisions, where biases could have life-altering consequences~\cite{giuffre2024systematic}; in finance, LLMs support fraud detection and automated customer service, where biases could lead to significant economic harm~\cite{cook2025social}; in hiring, biases could systematically exclude qualified candidates~\cite{wilson2024gender}.
In such contexts, ensuring fairness and transparency is not just an ethical imperative but often a legal requirement, as recent regulations emphasize algorithmic accountability and nondiscrimination~\cite{smuha2025regulation, groves2024auditing, Fabris_2025}.

While studies have found that LLMs outperform humans in many domains such as social situational judgments, diagnosis, prediction, and persuasion~\cite{mittelstadt2024large,bejan2025large, luo2025large, matz2024potential}, a growing literature highlights substantial risks associated with deploying these systems in real-world decision contexts. In particular, LLMs can exhibit systematic biases related to gender, race, and other demographic attributes~\cite{an2025measuring, wen2025faireassessingracialgender, templin2025framework}, which can be subtle and, in some cases, even stronger than those observed in human decision-making~\cite{hofmann2024ai}. Beyond demographic bias, recent work documents linguistic stereotyping based on dialect~\cite{fleisig2024linguistic} as well as cultural favoritism toward groups that are dominant within a given language context~\cite{levy2023comparing}, raising concerns about fairness across diverse populations. Other work has documented sensitivity to prompt framing and contextual cues, which can lead to inconsistent or unstable decisions across otherwise equivalent inputs \cite{valkanova2024irrelevant,yang2024llm, ceballos2024open}. Because these systems are increasingly embedded in automated pipelines, even small contextual effects may propagate into large-scale downstream outcomes. 

The impact of these issues is further compounded by the interaction between LLM outputs and human decision-makers. Studies show that LLM-generated recommendations can increase user overconfidence and amplify existing human biases~\cite{sun2025large, cheung2025large}, with downstream effects in which humans adopt or inherit AI-induced biases~\cite{wilson2025no}. More broadly, reliance on LLMs can lead to forms of cognitive surrender, where users defer to models' outputs even when they are flawed~\cite{shaw2026thinking}. Together, these findings highlight that LLM outputs are not only subject to bias, but can also shape and reinforce human decision-making in ways that may degrade outcomes in practice.

In many real-world decision-making settings, LLMs are used to compare alternatives, requiring models to choose among or rank competing options rather than evaluate each in isolation.
Such comparisons arise, for example, in medical triage~\cite{ramaswamy2026chatgpt}, LLM-as-a-judge frameworks~\cite{shi2025judging}, and hiring decisions~\cite{wang2024jobfair}. While alternatives could in principle be evaluated independently using absolute scores, a growing body of work advocates for pairwise or setwise comparisons, arguing that relative judgments are more consistent, robust, and aligned with human evaluation processes~\cite{liu2024aligning, zhuang2024setwise, qin2023large, zheng2023judging}. However, although comparative judgments can improve reliability relative to evaluating items independently, they implicitly assume that preferences are invariant to irrelevant aspects of presentation, such as the order in which alternatives are shown. This makes them inherently sensitive to contextual and positional effects. In human decision-making, the order in which alternatives are presented can systematically influence judgments, as documented in extensive behavioral research on primacy and recency effects~\cite{webster1964decision, springbett1958factors, rey2020primacy, murphy2006primacy, london1974effects, carlson1971effect}, anchoring~\cite{tversky1974judgment, furnham2011literature}, and contrast effects~\cite{wexley1973training, latham1975training}. When several similar alternatives are arranged in an array, people tend to choose the central options \cite{christenfeld1995choices,shaw2000centrality,valenzuela2009position}. Preferences can also be shaped by the composition of the choice set, for example via decoy effects~\cite{HIGHHOUSE199668, PETTIBONE2000300, ISRdecoy}.  These biases lead to instability  across equivalent presentations and can undermine consistency and lead to irrational preferences.

Recent studies show that LLMs are vulnerable to similar biases. Order effects have been documented in multiple-choice settings~\cite{pezeshkpour202, zhenglarge},  
classification tasks~\cite{guo2024serialpositioneffectslarge, wang2023primacy}, conversational summarization~\cite{sun2025posum},  and pairwise evaluations~\cite{zheng2023judging, rozado2025gender}. Other context-related effects, such as anchoring and decoy effects, have also been observed~\cite{lou2024anchoring, valkanova2024irrelevant}. Sensitivity to prompt ordering has also been observed in reasoning tasks~\cite{chen2024premise, lu2022fantastically}. Yet this literature remains largely descriptive: we know position biases exist, but not how different contexts influence them, the extent to which they are model-specific, or how they interact with other biases. It is also unclear whether they reflect tie-breaking or true preference distortions, or how they differ from position biases in humans.

 In this work, we systematically investigate positional biases in LLMs and their implications for comparative decision-making.\footnote{These should be distinguished from a different form of position bias arising from interactions between attention mechanisms and positional embeddings \cite{wu2025emergence}. In contrast, we study biases in which models favor alternatives based on their position within a set of choices.} We uncover novel behavioral patterns, demonstrate that these effects can meaningfully distort underlying preferences (which we operationalize using random utility theory), and identify important differences from human biases. We examine these effects across five LLM families, comprising nine widely used models, and evaluate them in two fundamentally different domains: hiring decisions and children's room color selection.
 The hiring scenario is a high-stakes application with a concrete link to existing literature on LLM biases. In contrast, color selection is a low-stakes, previously unexplored application that is minimalistic, neutral and easily replicable. Because it is largely free of other well-documented biases such as gender or race, it allows us to isolate position  effects more precisely than in complex domains. For each domain, we conduct experiments using four quality tiers. We uncover a consistent, quality-dependent pattern in pairwise comparisons across both domains: when both options are of relatively high quality, LLMs tend to favor the first candidate or color; when the options are of lower quality, they tend to favor the second. Although the trends are similar across models, what constitutes ``high'' and ``low'' quality may vary across models. In pairwise resume comparisons for instance, Llama 3 8B favors the first candidate in the top two tiers and the second in the bottom two, while GPT-4o-mini favors the first candidate in the top tier, and the second in the bottom three.  This quality-dependent shift appears to be a distinct pattern in LLM decision-making that, to our knowledge, has not been documented in human or prior LLM studies. 
 
The pattern extends to triplewise comparisons: when all three options are high-quality, LLMs tend to prefer the first option, whereas for lower-quality options, they favor later ones.
We further show that position biases are, for the most part, stronger than gender biases in our experimental settings, and uncover a significant and previously undocumented source of distortion: a bias favoring certain names over others. 
These patterns suggest that LLMs are not merely inheriting human-like heuristics from their training data,  but are manifesting new failure modes with distinct underlying mechanisms. We further find that these effects persist across languages (Mandarin Chinese and Hebrew), even though the underlying preferences are language-dependent.

In addition to  these novel patterns, we highlight a conceptual distinction that has been largely overlooked in prior work on order effects in both humans and LLMs. Although rarely distinguished in prior work, position biases can arise from two fundamentally different phenomena: tie-breaking heuristics and true preference distortions. Tie-breaking occurs when the agent (human or AI) has no clear preference and defaults to a fixed positional rule (e.g., always guessing option C on a multiple-choice test). In contrast, a distortion occurs when the order of presentation alters an otherwise strict preference. This distinction is critical for interpreting observed order effects. While tie-breaking reflects indifference and is relatively benign, preference distortion implies that the model may select inferior options due solely to presentation order. Without distinguishing between these mechanisms, observed position effects may be misinterpreted as harmless randomness when they in fact reflect systematic distortions of underlying preferences.

Disentangling these mechanisms is difficult, especially in domains where ground-truth preferences are ambiguous. Many prior studies suggest that order effects emerge mainly when options are similar~\cite{pezeshkpour202, zheng2023judging, zhao2024measuring}, consistent with the tie-breaking view. However, we show that position  effects in LLMs can cause distortions that lead to genuine preference reversals. In other words, even though the LLM strictly prefers option $a$ to option $b$, it might nevertheless choose option $b$ when the two are presented in a certain order. To address this, we introduce a simple yet effective mitigation strategy: querying the model numerous times at higher temperature settings.

At low temperatures, models' outputs are nearly deterministic. Assume that an LLM prefers option $a$ to option $b$ but there is a primacy bias strong enough to reverse this preference. Then, when the model is given a choice between $a$ and $b$ at a low temperature, it will always select the first option presented. Here,  position biases can fully determine which option is selected, masking the model's underlying preferences.

When the temperature parameter (denoted $T$) increases, the models' outputs become more stochastic; repeated sampling shows more variation. While
$T=0$ produces (nearly) deterministic outputs, higher values such as $T=1$ yield greater randomness. By repeatedly querying the model at higher temperatures and  examining selection frequencies across permutations, we can recover the model's underlying preference structure even when lower temperatures obscure it.

To formalize the  preference  structure, 
we extend the classical rational choice framework, which represents preferences using the binary relations \( a \succ b \) (strict preference) and \( a \indiff b \) (indifference). We introduce a refined notation that distinguishes between two types of strict preferences. We write \( a \robust b \) to indicate that the preference for \( a \) over \( b \) is \emph{robust} and stable across contexts. In contrast, \( a \fragile b \) denotes a \emph{fragile} preference: a genuine but unstable preference that may be reversed by order effects.
The distinction between fragile and robust preferences is important because fragile preferences imply that LLM decisions may depend on arbitrary aspects of prompt structure leading the model to make suboptimal decisions.

We define these preferences more formally within our setting  using random utility theory~\cite{manski1977structure}, where one defines preferences using selection probabilities. Let $\Pr[a \mid \{a,b\}, t]$ denote the overall probability of selecting $a$ when both permutations $(a,b)$ and $(b,a)$ are presented an equal number of times at temperature $t$. Intuitively, if $\Pr[a \mid \{a,b\}, t]= 0.5$ for all temperatures $t$, then the model is indifferent between $a$ and $b$. If there exists some temperature $t$ such that $\Pr[a \mid \{a,b\}, t] >  0.5$, then the model strictly prefers $a$ to $b$. We note that we assume there do not exist $t_1$ and $t_2$ such that $\Pr[a \mid \{a,b\}, t_1] > 0.5$ and $\Pr[a \mid \{a,b\}, t_2] < 0.5$, i.e., we rule out preference reversals across temperatures. This is consistent with the role of temperature in language models, which adjusts the randomness of outputs while generally preserving the relative likelihood of alternatives.

At $T=0$, if the model's output were perfectly deterministic, there would be only two possible behaviors: (i) the same option is selected in both orders, giving selection probabilities of $\{0,1\}$, (ii) the model consistently picks one position (e.g., the first option), so each option is chosen half of the time, giving $\{0.5, 0.5\}$. In practice, $T=0$ is not perfectly deterministic;  we use `$\approx$' to indicate probabilities close to, but not exactly, $0$, $0.5$, or $1$.\footnote{Because $T=0$ does not always guarantee perfectly deterministic outputs, we fixed the random seed whenever preliminary runs exhibited non-negligible variability at $T=0$. We classify a frequency as approximately equal to one of $0$, $0.5$, or $1$ if it falls within $0.05$ of that value; in practice, all models tested deviated by at most $0.02$ from the relevant benchmark value.} Preferences are classified as follows:

\begin{itemize}[itemsep=0pt]
    \item 
    If $\Pr[a | \{a,b\}, 0]\!\approx\!1$ and $\Pr[a | \{a,b\}, 1]\!>\!0.5$ then \( a \robust b \).
    \item 
    If $\Pr[a | \{a,b\}, 0]\!\approx\!0.5$ and $\Pr[a | \{a,b\}, 1]\!>\!0.5$ then \( a \fragile b \).
    \item 
    If $\Pr[a | \{a,b\}, 0]\!\approx\!0.5$ and $\Pr[a | \{a,b\}, 1]\!=\!0.5$ then \( a \indiff b \).
\end{itemize}

Preferences that do not fit into any of these cases (for example, if  $\Pr[a \mid \{a,b\}, 0]\!\approx\!1$ while $\Pr[a \mid \{a,b\}, 1]\!\approx\!0.5$) are deemed \emph{inconsistent}; we do not observe such cases in our experiments.

Simply put, if the model selects the same option (say $a$) when queried with $(a,b)$ and $(b,a)$ at $T=0$, then $a \robust b$; the model robustly prefers $a$ to $b$. If the model's choice flips when the presentation order is reversed at $T=0$, then the model may either be indifferent between the two options ($a \indiff b$) or have a fragile preference for one of them ($a \fragile b$ or $b \fragile a$). These cases cannot be distinguished based on $T=0$ alone.

To differentiate between them, we query the model at $T=1$, using an equal number of each permutation. If one option is selected more frequently---for example, if $\Pr[a \mid \{a,b\}, 1] > 0.5$---we interpret this as a fragile preference ($a \fragile b$), meaning that the order-dependent reversal at $T=0$ reflects a genuine preference distortion. If both options are selected with approximately equal frequency (i.e., $\Pr[a \mid \{a,b\}, 1] \approx 0.5$), we interpret this as indifference ($a \indiff b$), in which case the observed order effect is consistent with tie-breaking.  Taken together, this provides a simple procedure: query at $T=0$ to detect order sensitivity, and, when present, use repeated sampling at higher temperatures to determine whether the effect reflects indifference or a distorted preference.

In practice, distinguishing between fragile preference and indifference may be difficult when $\Pr[a \mid \{a,b\}, 1]$ is close to $0.5$. However, in such cases the underlying preference is necessarily weak, and the distinction has limited practical impact.  In practice, this suggests fixing a reasonable number of samples and testing whether selection frequencies differ significantly from $0.5$; if not, the underlying preference is weak and the distinction between fragile preference and indifference is unlikely to be consequential.


When order effects at low temperatures obscure underlying preferences, increasing temperature serves as a diagnostic tool, revealing these latent preferences through repeated sampling. By introducing controlled stochasticity, higher temperatures allow us to estimate the underlying preference distribution rather than observe only a single, potentially biased outcome. As a result, in such settings, higher temperatures can improve preference detection accuracy, contrary to the prevailing understanding that greater accuracy is attained at lower temperature settings~\cite{xu2022systematic,renze2024effect}.

Our findings suggest that comparative decisions in LLMs are shaped by a combination of familiar and novel biases, and that these biases can systematically distort underlying preferences. Importantly, these distortions are not merely artifacts of indifference or randomness: they often reflect genuine preference reversals driven by presentation order. This has practical implications for how LLM-based comparisons should be interpreted and used. In particular, positional effects can outweigh demographic biases and cannot be reliably mitigated by simple strategies such as averaging or randomizing presentation. Instead, careful analysis of selection patterns across permutations and temperatures is required to recover and interpret underlying preferences.

%% file: _results.tex
We selected two domains to balance realism and control over confounding factors: resume screening and color selection.  In the hiring domain, we evaluated three large language models: GPT-4o-mini, Claude 3 Haiku, and Llama 3 8B. In the more controlled color-selection setting, we examined position biases across nine widely used LLMs spanning five families: OpenAI (GPT-4o-mini and GPT-4.1-nano), Anthropic (Claude 3 Haiku and Claude Sonnet 4), Meta (Llama 3 8B and Llama 4 Scout), Google (Gemini 2.5 Flash and Gemini 3 Flash), and Alibaba (Qwen 3 32B).

As order effects are most likely when alternatives are similar in quality~\cite{zheng2023judging, pezeshkpour202}, we created item sets close in quality and grouped them into four tiers: \emph{Best}, \emph{Good}, \emph{Mediocre}, and \emph{Weak} for resumes, and \emph{Ideal}, \emph{Fair}, \emph{Plain}, and \emph{Harsh} for colors. In the hiring domain, we used LLM-generated resumes for four professions (Mechanical Engineer, Registered Nurse, Journalist, and Real Estate Agent). In the color domain, we curated sets of wall paint colors grouped by suitability for a child's room.

For each model and domain, we conducted exhaustive pairwise and triplewise comparisons, testing all permutations of item order. Prompts included ``Select the strongest candidate'' and ``Which color is best for a kid's room?'' Each model was evaluated at multiple temperature settings. At \(T=0\), models behaved with near-determinism, selecting the most likely answer. At \(T=1\), responses varied across repeated queries, allowing statistical analysis of position effects; each \(T=1\) response was treated as an independent sample from the model's output distribution.

We define position biases as follows. A model exhibits a \emph{primacy bias} if there exists a set of options such that the model disproportionately prefers whichever option is presented first. It exhibits a \emph{recency bias} if there exists a set of options such that it disproportionately prefers whichever option is presented last. In triplewise comparisons, it exhibits a \emph{centrality bias} if there exists a set of three options such that it disproportionately prefers whichever option is presented second. A model may exhibit more than one bias depending on the context.

In the main text, we present results from three representative models --- GPT-4o-mini, Claude 3 Haiku, and Llama 3 8B. Full experimental details appear in the Materials and Methods section and SI Appendices. The results for the remaining models are presented in \ifpnas SI Appendix~K \else Appendix~\ref{app:additional_models} \fi.

\subsec{Order Effects in Pairwise Comparisons}

We find a consistent, quality-dependent position effect across models and domains: higher-quality options tend to exhibit primacy effects, while lower-quality options exhibit recency effects.  
For instance, in color selection \ifpnas {(Figure~\ref{fig:pairwise_both_temp_1}A)} \else {(\autoref{fig:pairwise_temp1})}\fi, GPT-4o-mini displayed a primacy effect in the top two tiers (Ideal and Fair) and a recency effect in the bottom two (Plain and Harsh). Claude 3 Haiku and Llama 3 8B showed a primacy effect in the top one and three tiers, respectively, and a recency effect in the bottom three and one tiers, respectively. In \ifpnas {Figure~\ref{fig:pairwise_both_temp_1}B} \else {\autoref{fig:chosenpair_temp1}}\fi, the resume selection results followed a similar overall pattern (though GPT-4o-mini and Claude 3 Haiku showed primacy only in the top tier, and Llama 3 8B in the top two tiers).  This pattern suggests that position bias interacts systematically with perceived quality; position effects depend on the underlying preference structure.

Full statistics for these three models, including those for additional temperature settings, appear in \ifpnas{SI Appendices~C and~D}\else{Appendices~\ref{app:pairwise_colors} and~\ref{app:pairwise_resumes}}\fi. Details for additional models are presented in \ifpnas SI Appendix~K\else Appendix~\ref{app:additional_models}\fi. 


\ifpnas {} \else {
\begin{figure*}[th]
  \centering

  \begin{subfigure}[b]{0.95\textwidth}
    \includegraphics[width=\textwidth]{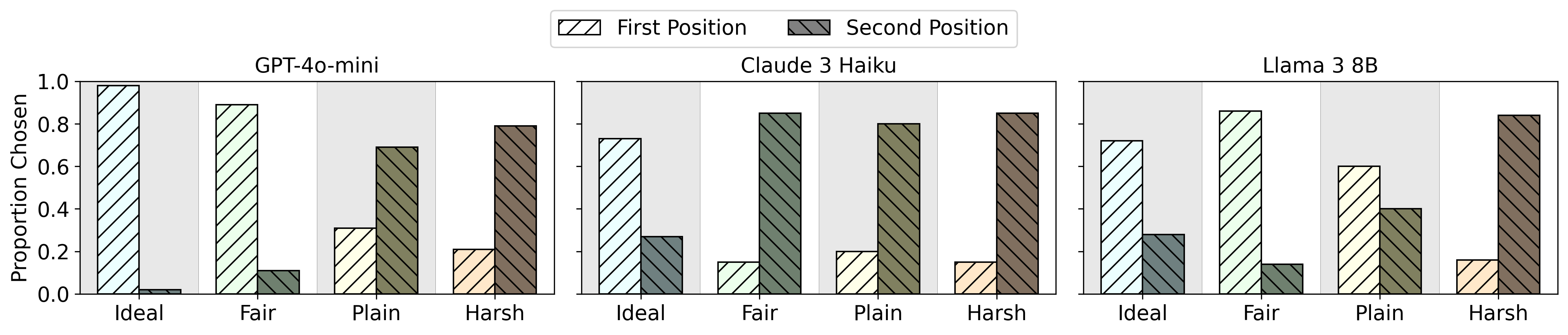}
    \caption{Proportion of color choices by position across 4 quality tiers  in pairwise comparisons. Tier quality is decreasing from left to right: `Ideal' is the highest quality tier and `Harsh' is the lowest.}
    \label{fig:pairwise_temp1}
  \end{subfigure}

  \begin{subfigure}[b]{0.95\textwidth}
    \includegraphics[width=\textwidth]{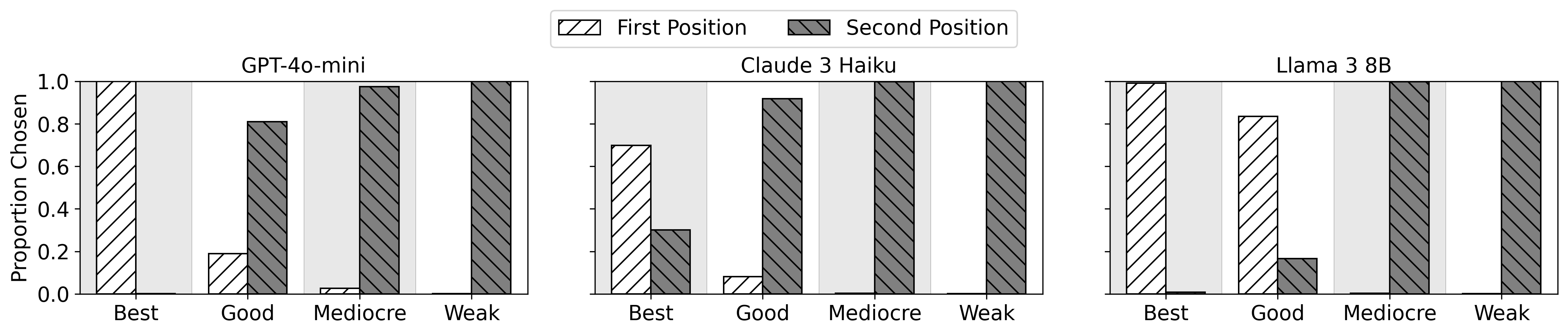}
    \caption{Position bias in pairwise resume selection,  aggregated across professions.}
    \label{fig:chosenpair_temp1}
  \end{subfigure}

  \caption{{\bf Position bias in pairwise comparisons at \boldmath \(T=1\).} 
  (a) shows the effect of presentation order on color selection in pairwise comparisons for each quality tier. 
  (b) shows aggregate positional effects in resume evaluations across professions.
  In both tasks, higher-quality options tend to exhibit a primacy effect and lower-quality ones a recency effect, though the precise threshold separating them varies by model and domain. 
  Full results for color comparisons are provided in \ifpnas SI Appendix~C \else Appendix~\ref{app:pairwise_colors} \fi; results for resumes disaggregated by profession are given in  \ifpnas SI Appendix~D. \else Appendix~\ref{app:pairwise_resumes}. \fi
}

   \label{fig:pairwise_both_temp_1}
\end{figure*}
}\fi

\subsec{Order Effects in Triplewise Comparisons}

In triplewise comparisons, higher-quality tier(s) exhibited a primacy effect, consistent with the pairwise results. For lower-quality tier(s), all models favored later positions, though they differed in whether the second or third option was preferred. Specifically, GPT-4o-mini and Claude 3 Haiku sometimes favored the second (middle) position, exhibiting centrality bias. For example, at \(T=0\) (Figure~\ref{fig:triplewise_both_temp_0}A)
, GPT-4o-mini always selected the middle option for \emph{Plain} colors, and Claude 3 Haiku did so 83\% of the time for \emph{Harsh} colors. In resume selection (Figure~\ref{fig:triplewise_both_temp_0}B)
, GPT-4o-mini preferred the middle option in three out of the four tiers tested. 

Several of the models showed simultaneous primacy and recency effects, resembling the serial position effects observed in humans~\cite{murphy2006primacy,Baucells2011ref}. Among colors in the \emph{Fair} tier at \(T=0\), Llama 3 8B chose the first and last positions equally often (50\% each) and never selected the middle option. See \ifpnas{SI Appendices~F and~G} \else{Appendices~\ref{app:triplewise_colors} and~\ref{app:triplewise_resumes} }\fi for additional details. Results from additional large language models are presented in \ifpnas SI Appendix~K \else Appendix~\ref{app:additional_models} \fi. 

Across models and domains, position bias generally shifted toward later positions as quality declined, with the exception of Claude 3 Haiku in the color domain. Position biases were notably weaker in resume selection for Claude 3 Haiku and Llama 3 8B compared to color selection. As we show in the \emph{Name Bias} section below, this attenuation is largely due to interference from strong preferences for specific names.   
These results show that position biases can manifest in more complex ways and  suggest that they are not limited to simple heuristics, but can adapt to the structure of the choice set.

To explore whether position biases arise in larger choice sets, we conducted preliminary fourwise color comparisons for each model. Although we only evaluated three models, and a single quality tier per model, the results demonstrate that position biases can persist in more complex choice scenarios. For full details, see \ifpnas SI Appendix~H. \else Appendix~\ref{app:four_colors}. \fi


\ifpnas {} \else {
\begin{figure*}[th]
  \centering

  \begin{subfigure}[b]{0.95\textwidth}
    \includegraphics[width=\textwidth]{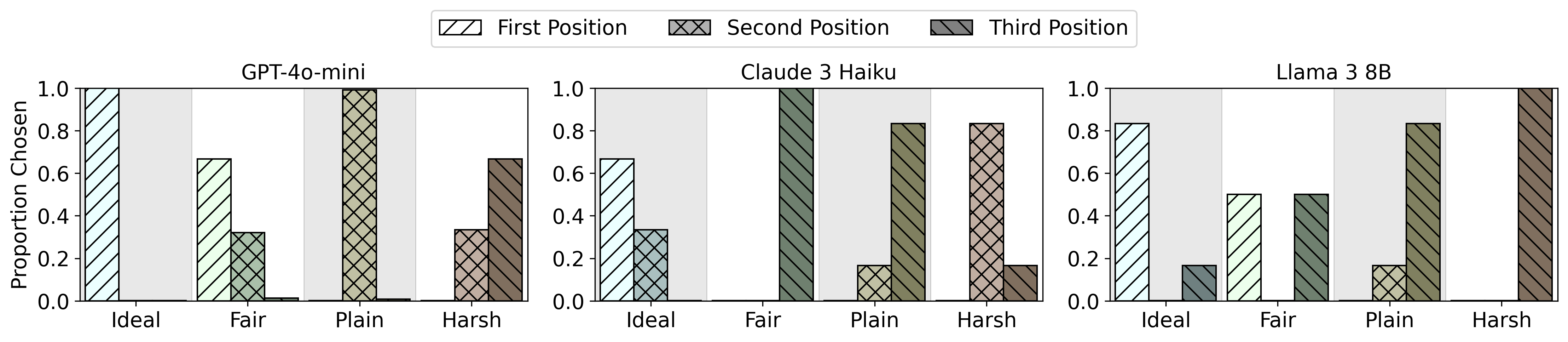}
    \caption{Proportion of color choices by position across 4 quality tiers in triplewise comparisons.}
    \label{fig:triple_colors_temp_0}
  \end{subfigure}

  \begin{subfigure}[b]{0.95\textwidth}
    \includegraphics[width=\textwidth]{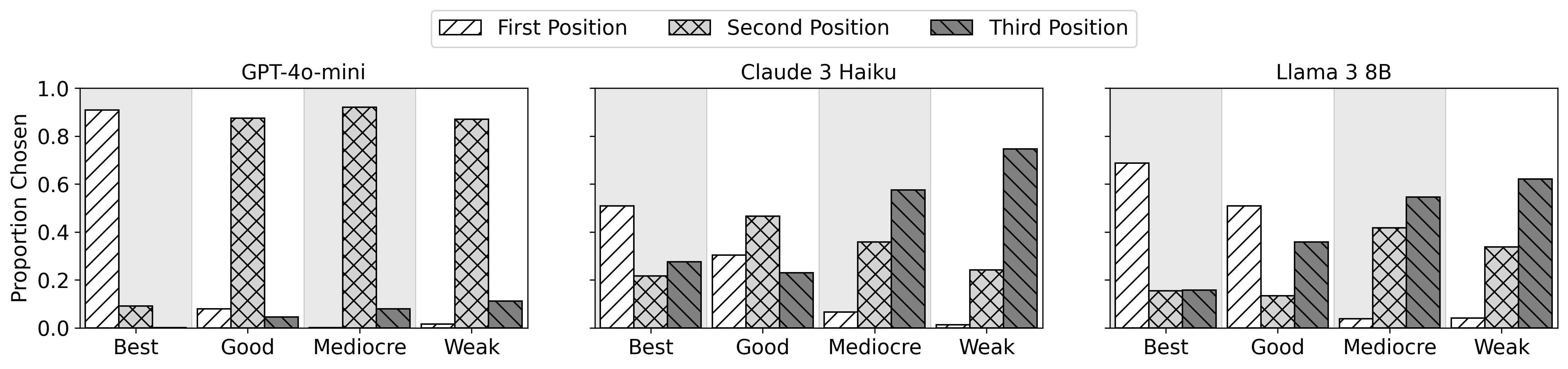}
    \caption{Aggregate position bias in triplewise resume selection across 4 quality tiers.}
    \label{fig:triple_all_resumes_temp_0}
  \end{subfigure}

\caption{
{\bf Position bias in triplewise comparisons at \boldmath \(T=0\).}
(a) shows the effect of presentation order on color selection. 
(b) shows aggregate position effects in resume evaluations across professions.
Additional results for color selection are provided in \ifpnas SI Appendix~F\else Appendix~\ref{app:triplewise_colors}\fi, and resumes selection results disaggregated by profession are in \ifpnas SI Appendix~G\else Appendix~\ref{app:triplewise_resumes}\fi.
}

   \label{fig:triplewise_both_temp_0}
\end{figure*}
}\fi

\subsec{Cross-Language Robustness and Preference Shifts}

We replicated the color-selection experiments in Mandarin Chinese and Hebrew, using GPT-4o-mini. The overall positional patterns remain consistent across languages: higher-quality options tend to exhibit primacy effects, while lower-quality options exhibit recency effects.

However, the specific options exhibiting these effects differ across languages. In English, blue tones (e.g., Aqua Mist, Soft Sky Blue) tend to exhibit primacy effects, while pastel tones (e.g., Gentle Coral, Buttercream Yellow) exhibit recency effects. In Hebrew, this pattern reverses: pastel tones exhibit primacy effects, while blue tones exhibit recency effects.

This difference reflects a shift in underlying preferences: in English, the model tends to prefer blue tones, whereas in Hebrew it prefers pastel tones. This shift is also evident in specific pairwise comparisons. For example, when choosing between Frosted Cyan and Transparent Ice Blue at $T=0$, GPT-4o-mini consistently selects Frosted Cyan when prompted in  English but selects Transparent Ice Blue when prompted in Hebrew. When prompted in Mandarin Chinese, it displays a primacy effect: selecting the first option presented.  
These results suggest that models' preferences are language-specific, and not absolute.

\subsec{Tie-Breaking Heuristics vs. True Distortions}
\label{subsec:tie-breaking}

Do order effects reflect arbitrary tie-breaking, or genuine distortions of the model's underlying preferences? Our notation allows us to distinguish between these cases in pairwise comparisons. 
For color selection, all but one of the thirty six model-tier pairs satisfied $\Pr[a \mid \{a,b\}, 0] \!\approx\! 0.5$: at \(T=0\); all models selected the same position at least 98\% of the time for each tier. 
When the option selected is order-dependent at \(T=0\) (i.e.,  \(\Pr[a \mid \{a,b\}, 0] \!\approx\! 0.5\)), exactly one of the following holds: \(a \fragile b\), \(b \fragile a\), or \(a \indiff b\). In other words, the model might be indifferent between the two colors or have a fragile preference for one.

To determine whether the models were indifferent or had a fragile preference, we repeated the comparisons at \(T=1\).  If  $\Pr[a \mid \{a,b\}, 1]\!\approx\!0.5$, the model is indifferent (\(a \indiff b\)); if $\Pr[a \mid \{a,b\}, 1] > 0.5$, the model prefers option $a$ (\(a \fragile b\)).  To build intuition as to why an increase in temperature can help distinguish between indifference and a true preference, consider a stylized example with von Neumann-Morgenstern (VNM) utilities.\footnote{A VNM utility function is any function that represents a preference relation consistent with the von Neumann-Morgenstern axioms. See~\cite{von2007theory} for a formal definition. We use VNM utilities only for intuition; we do not claim that the models' preferences can be mapped to such a function or that they satisfy the axioms of rationality.}

Suppose that there is a primacy boost \(\delta = 0.2\); that is, the perceived utility of the first option is increased by $0.2$. Assume first that the model is indifferent between the two options and assigns $u(a) = u(b) = 5$. At \(T=0\), presenting \((a,b)\) yields \(a\) (since \(5 + 0.2 > 5\)), and presenting \((b,a)\) yields \(b\) (since \(5 + 0.2 > 5\)). 
Consider now the case that (\(b \fragile a\)), and the model assigns $u(a) = 5, u(b) = 5.1$.  At \(T=0\), presenting \((a,b)\) yields \(a\) (since \(5 + 0.2 > 5.1\)), and presenting \((b,a)\) yields \(b\) (since \(5.1 + 0.2 > 5\)). In both cases, the choice is completely order dependent.

Now suppose \(T=1\) adds uniform noise \(\varepsilon \sim U[-0.4, 0.4]\) to the first option's utility (on top of the position bias).\footnote{Noise can be assumed to be added to only one position without loss of generality.} Then \((a,b)\) yields \(a\) with probability 0.625 (\(\mathbb{P}[5.2+\varepsilon \geq 5.1] = 0.625\)), and \((b,a)\) yields \(b\) with probability 0.875. Thus, \(b\) is significantly more likely to be chosen than \(a\) at \(T=1\), revealing a strict (but fragile) preference for \(b\) over \(a\), rather than indifference.  

We conducted this experiment on our three representative models:  GPT-4o-mini, Claude 3 Haiku, and Llama 3 8B. Consistent with this intuition, eight of twelve model-tier color pairs in our experiments showed a significant preference (\(p < 0.05\)), and three of these showed overwhelming distortion (\(p < 10^{-6}\)). We obtained similar results for resumes; in one case, one resume was 1.5 times more likely to be chosen than the other, indicating that order effects can lead to the consistent selection of inferior options---effects that are most pronounced at lower temperatures. This provides  evidence that order effects often reflect genuine distortions of underlying preferences rather than arbitrary tie-breaking. For more details, see \ifpnas SI Appendix~E \else Appendix~\ref{app:tie_breaking} \fi.


\ifpnas {} \else {
\begin{figure*}[th]
  \centering

\begin{subfigure}[b]{0.95\textwidth}
  \includegraphics[width=\textwidth]{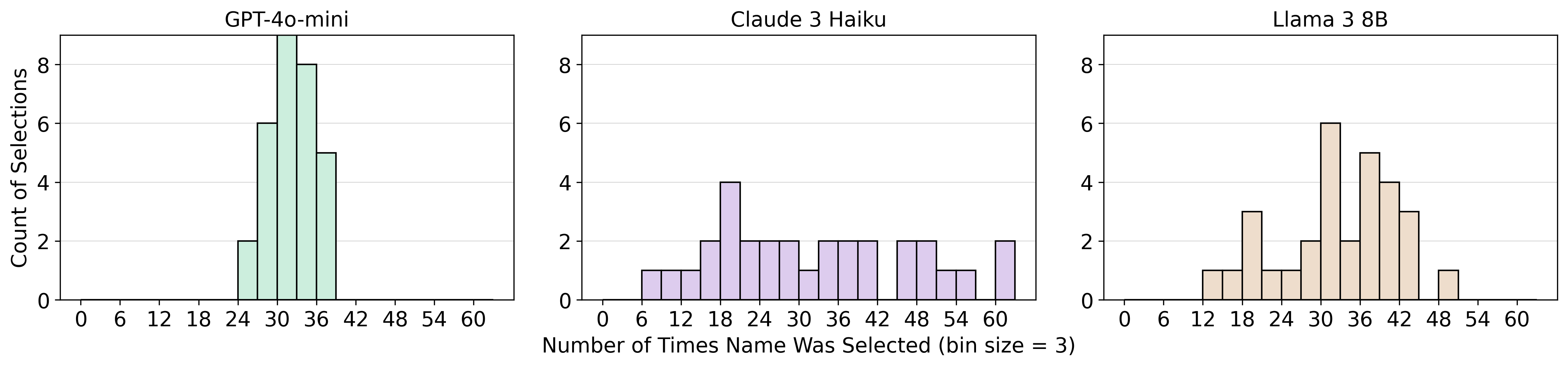}

\caption{Name selection frequencies in triplewise comparisons. The histogram shows how often each candidate name was selected across all triplewise tasks. The x-axis represents the number of times a name was selected, and the y-axis shows how many names were selected that number of times. For GPT-4o-mini, nine names were selected between 30 and 32 times, and all names were selected between 24 and 38 times.
See  SI Appendix~I  for additional results and details.
}

  \label{fig:name_bias_main}
\end{subfigure}

\begin{subfigure}[b]{0.95\textwidth}
  \includegraphics[width=\textwidth]{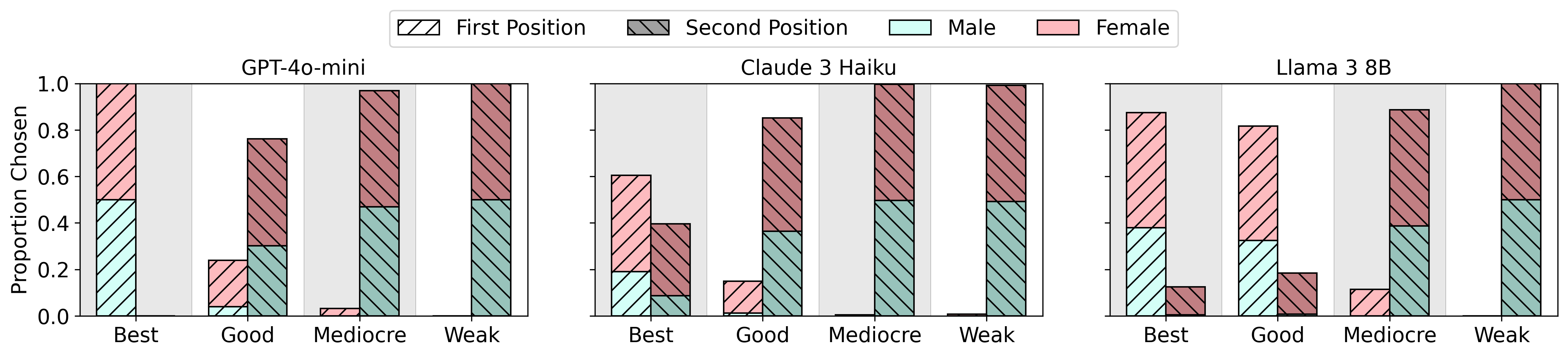}
  \caption{Gender and order effects at \(T=0\) in pairwise resume comparisons. All models show stronger biases for presentation order than for gender. 
  Results by profession and temperature are provided in  SI Appendix~J.
  }

  \label{fig:gender_bias_main}
\end{subfigure}

\caption{
{\bf Interaction of order effects with other biases.} 
(a) shows the distribution of name selections in triplewise comparisons. Claude 3 Haiku exhibits strong preferences for certain names, while GPT-4o-mini shows a relatively balanced selection pattern. 
(b) shows the effect of gender and presentation order in pairwise resume comparisons. Order effects appear stronger than gender effects. 
}

   \label{fig:other_biases}
\end{figure*}

}\fi

\subsec{Interaction of Position Bias with Other Biases}
In addition to order effects, LLMs exhibit other well-documented biases, such as preferences based on gender or identity. In this section, we examine how these biases interact with position effects. While gender bias in algorithmic decision-making is well documented, we also uncovered an additional bias: consistent preferences for specific individual names, even when those names were carefully selected to avoid known bias-triggering cues such as race and ethnicity.  To our knowledge, such name-specific biases have not been documented in human decision-making under comparable conditions. Across all models, we found that both gender and name biases are generally weaker than the position effects, although in some cases they appear to compete with or modulate them.

\subsubsec{Name Bias} 
The position bias in triplewise color comparisons appears much stronger than in resume comparisons particularly for Claude 3 Haiku and Llama 3 8B (see Figure~\ref{fig:triplewise_both_temp_0}A)
. 
This reduction in position bias appears to have been driven primarily  by strong, competing biases for or against specific synthetic personas, especially individual names. Claude 3 Haiku, in particular, exhibited strong name-based preferences: it consistently favored certain names over others, regardless of resume content or position. For example, in direct comparisons, Claude 3 Haiku selected `Christopher Taylor' over `Andrew Harris' in 64\% of the cases (82 out of 128 trials, binomial $p = 0.0019, h = 0.29$). These patterns emerged even though all names were drawn from a controlled set of generic Caucasian U.S. identities, chosen to minimize racial, ethnic, and other socially marked biases that have been documented in LLMs \cite{bai2025explicitly, cheng2023marked}.

This pattern is evident in Figure~\ref{fig:other_biases}A
, which shows the distribution of selection frequencies for each name across triplewise comparisons. Claude 3 Haiku's distribution is flat, indicating that a few names were chosen disproportionately often while others were rarely selected---consistent with strong and persistent name preferences. GPT-4o-mini, by contrast, displays a near-perfect normal distribution, suggesting minimal name-based bias in the triplewise setting. Llama 3 8B lies between these extremes, with moderate variation across names.
Although GPT-4o-mini  showed no signs of name bias in triplewise comparisons, we detected a small but statistically significant name preference in the pairwise setting. 
See \ifpnas SI Appendix~I \else Appendix~\ref{app:name} \fi for full details.

\subsubsec{Gender Bias}
Prior work on gender bias in algorithmic hiring is mixed: some studies report little to no bias \cite{Armstrong_2024}, others find systematic preferences for male candidates \cite{kotek2023gender, wilson2024gender}, while more recent evaluations suggest a tilt toward female candidates \cite{gaebler2024auditing, wang2024jobfair} or context-dependent outcomes \cite{wen2025faireassessingracialgender}. Studies using pairwise evaluations have found a bias favoring female candidates \cite{rozado2025gender, gaebler2024auditing, wang2024jobfair}, a pattern we also replicated: for example, GPT-4o-mini  selected female candidates 54.8\% of the time at \(T=0\), with Claude 3 Haiku and Llama 3 8B  showing similar trends. 

While gender bias in algorithmic decision-making is an important topic in its own right, our primary goal is to examine how gender interacts with  order effects. We found that gender effects were overshadowed by substantially stronger order biases. Across all three models, the candidate in the dominant position (first or second, as defined by same-gender comparisons) was selected at much higher rates. For GPT-4o-mini, this rate reached 92.8\% in mixed-gender comparisons, compared to 93.6\% in same-gender trials, suggesting that gender bias can at times counteract, but rarely override, the positional preference. Claude 3 Haiku and Llama 3 8B  showed similar patterns, with position bias rates of 84.3\% and 72.3\% in mixed-gender comparisons, versus 87.6\% and 89.2\% in same-gender comparisons, respectively. Bootstrapped confidence intervals confirm that order bias significantly exceeds gender bias across all models (95\% CI for $h_{\text{order}} - h_{\text{gender}}$ excludes 0 in all cases); see \ifpnas SI Appendix~J \else Appendix~\ref{app:gender} \fi for figures and additional details. 

This difference between the effect sizes of the position and gender effects can be clearly seen in Figure~\ref{fig:other_biases}B
, where gender effects manifest in two ways: a modest overall preference for female candidates, and the observation that nearly all instances of overcoming the positional disadvantage involve female candidates. Full statistics appear in  \ifpnas SI Appendix~J. \else Appendix~\ref{app:gender}. \fi

%% file: _discussion.tex
This study provides evidence of systematic position biases in LLM-driven comparisons, with methodological, theoretical, and practical implications. Across both color and hiring domains, we find that position biases are quality-dependent and strong: primacy effects dominate for high-quality options, while later options (second or third) are preferred for lower-quality choices. Because many applications rely on LLMs to rank, filter, or select among alternatives, even small systematic biases can propagate through decision pipelines and lead to consistently suboptimal outcomes.

Most prior research has not clearly distinguished between two possible sources of position  effects: arbitrary tie-breaking and genuine preference distortion. Many studies have observed that order effects appear when the model is indifferent between the different options~\cite{pezeshkpour202, zheng2023judging, zhao2024measuring}, and studies on position biases and the decoy effect often purposefully use alternatives of similar quality~\cite{luce1998choosing, valkanova2024irrelevant}. 
 While some human studies have attempted to disentangle these explanations~\cite{farmer2017effect}, conclusions remain limited due to humans' limited ability to determine their `true' preferences~\cite{tversky1974judgment} or compute expected utilities~\cite{lichtenstein1971reversals}. Determining ground-truth preferences in human subjects is further complicated by contextual influences and shifts in preferences over time \cite{Hedgecock2016}.

We leverage features of LLMs that are not available in human studies---the ability to repeat identical queries without memory carryover, and the temperature parameter---to distinguish tie-breaking from preference distortion. Using this methodology, we show that LLMs do not merely exhibit position effects due to indifference but also because position effects can be strong enough to reverse strict  preferences.  Although subtle, this distinction has important practical consequences: tie-breaking is relatively benign, whereas distortion can lead to selecting an inferior option purely because of the order of presentation.

Much of the work on LLM bias has centered on uncovering and correcting cognitive distortions, such as anchoring, confirmation bias, or gender stereotyping, that models acquire from human-generated training data~\cite{acerbi2023large, atreides2024cognitive, salecha2024large, bai2025explicitly}. These studies often argue that models inherit cognitive biases that are present in the data~\cite{echterhoff2024cognitive, malberg2024comprehensive, shaikh2024cbeval}. Even when studies identify behavior that diverges from human norms (for instance, a female-candidate preference in resume-ranking tasks, where humans often favor male candidates), these findings are often interpreted through the lens of human cognitive biases~\cite{gaebler2024auditing, wang2024jobfair}. In contrast, we uncover two failure modes that do not fit cleanly within standard accounts of human cognitive bias:  position bias that varies systematically with option quality, and a name bias, even though the names were explicitly chosen to avoid known demographic associations.

These patterns raise new theoretical questions. Classical explanations for human order effects, such as memory limitations~\cite{deese1957serial,ebbinghaus2013image}, proactive inhibition~\cite{keppel1962proactive}, or heuristics \cite{tversky1974judgment}, cannot readily explain our findings of quality-dependent bias shifts. One possible account, drawing on the long-established notions of reference points~\cite{Baucells2011ref, BOLES1995262} and contrast effect~\cite{wexley1973training, latham1975training} in behavioral decision theory, is that LLMs establish an implicit reference point from their training data and then evaluate each option in contrast to this baseline. Above-baseline options are inflated, leading to primacy, while below-baseline ones are penalized. However, it is not clear how to extend this explanation to biases observed in triplewise comparisons, in particular the centrality bias. Moreover, model-specific differences (e.g., centrality effects in GPT-4o-mini and Claude 3 Haiku but not Llama 3 8B) suggest that distinct bias mechanisms may arise across models, potentially due to differences in architecture or training, though the precise sources of these differences, especially in closed models, remain unclear. Together, these findings call for new theoretical frameworks that go beyond human cognitive models in explaining how biases emerge and propagate in artificial agents. 

Our findings also carry practical implications for the deployment of LLMs in real-world decision systems.  It is critical to not only monitor for human-like cognitive distortions inherited through training data, but also to proactively identify and mitigate novel failure modes that emerge from the models’  architectures or training dynamics. These uniquely artificial biases may be harder to anticipate precisely because they have no direct human analogue. This underscores the importance of robust evaluation practices tailored to large language models.

We find that position biases are generally stronger than gender biases. This might be due to alignment efforts that explicitly target and reduce social biases like gender, while largely overlooking position biases. Nevertheless, because position  effects currently overshadow gender-based biases, their mitigation is equally, if not more, critical.

However, standard mitigation strategies such as randomizing orders or averaging across queries~\cite{wang2023primacy, pezeshkpour202} do not fully eliminate order effects. Randomizing the presentation order in a single query merely randomizes which option benefits from the bias. Evaluating all permutations and taking the majority (or plurality) response also fails at $T=0$: in pairwise comparisons, fragile preferences lead to each option being selected equally often across permutations. As a result, these approaches do not recover the model’s underlying preferences and can mask systematic distortions. To address these effects, we propose two LLM-specific mitigation approaches:

\begin{enumerate}[itemsep=0pt]
    \item \emph{Perform repeated queries at a higher temperature to uncover suppressed preferences.}  
    Prevailing wisdom suggests that responses at higher temperatures are less accurate~\cite{xu2022systematic,renze2024effect}. This is true for robust preferences: if $a \robust b$, option $a$ will always be selected at $T=0$, while option $b$ may be selected at $T=1$ due to stochasticity. Hence, queries at $T=0$ reliably recover robust preferences. However, position bias can completely mask fragile preferences at $T=0$, rendering them indistinguishable from true indifference. In such cases, increasing the temperature introduces controlled stochasticity that allows underlying preferences to emerge through repeated sampling. Thus, higher temperatures serve as a diagnostic tool: rather than reducing accuracy, they can improve preference detection by revealing latent preferences.

    Repeated querying also allows us to quantify preference strength. For example, if a model selects $a$ over $b$ 60\% of the time at $T=1$, and $c$ over $d$ 80\% of the time, we can infer that $a \succ b$, $c \succ d$, and that the latter preference is stronger. While our theoretical framework distinguishes between \emph{robust}, \emph{fragile}, and \emph{indifferent} preferences, in practice preference strength varies continuously and can be estimated with high statistical confidence through repeated sampling. We note that this approach requires multiple rounds of inference and assumes that selection frequencies at $T=1$ provide a meaningful signal of underlying preference structure.

    \item \emph{Use position bias as a diagnostic signal.}  
    Position bias itself can provide information about the underlying preference structure. For instance, a strong primacy effect may indicate that both options are of relatively high quality, while a strong recency effect may suggest that both are of lower quality. This interpretation follows from the observed interaction between quality and position bias. As a result, position effects can be used to guide downstream decisions, for example by triggering additional evaluation or human review when strong order effects are detected.
\end{enumerate}

These strategies treat position bias not only as a problem to be corrected but also as a source of information. In practice, they can be incorporated into evaluation pipelines by running targeted high-temperature queries to detect fragile preferences, and by interpreting the magnitude and direction of position bias as indicators that guide further decision-making. Crucially, our experimental results show that these approaches can recover suppressed preferences and reduce the likelihood of selecting an inferior option due solely to presentation order.

Our results hold across two disparate domains, suggesting that they are inherent rather than  domain-specific. While future training techniques may reduce such distortions, our findings suggest that positional sensitivity is currently widespread across models and can affect decision-making in systems that rely on comparative evaluation. Nevertheless, new architectures should be audited for order and identity biases. In high-stakes domains such as legal reasoning, medical diagnosis, or scientific peer review, bias-aware evaluation pipelines can combine model-driven diagnostics with human-in-the-loop checks. Finally, causal interpretability methods and adversarial probing could illuminate the internal mechanisms that give rise to preference distortions, paving the way toward bias-robust LLM design and deployment. Whether position biases can be fully eliminated remains an open question, and one that we believe warrants dedicated future investigation, particularly given the consistency of these effects across architectures and training regimes. These findings call for new theoretical frameworks that go beyond human cognitive models in explaining how biases emerge and propagate in artificial agents, particularly in systems deployed for automated or assisted decision-making.

%% file: _methods_pnas.tex
\subsec{Resume and Color Generation and Selection}
As order effects are more pronounced when the options are of comparable quality, we generated three options for each of four quality tiers of colors and resumes.  For colors, we denoted these tiers by \emph{Ideal}, \emph{Fair}, \emph{Plain}, and \emph{Harsh}; for resumes, by \emph{Best}, \emph{Good}, \emph{Mediocre}, and \emph{Weak}. We chose different names for the tiers in each domain so as not to imply an equivalence between them (i.e., \emph{Plain} colors are not necessarily comparable to \emph{Mediocre} resumes in terms of quality).

We first generated the candidates for each tier. For the resumes, we had GPT-4o-mini generate nine sets of four resumes of decreasing quality for each of four professions: Mechanical Engineer, Registered Nurse, Journalist, and Real Estate Agent. We removed the personal information from the resumes. 
The resume generation prompts are shown in \ifpnas {SI Appendix~B} \else Appendix~\ref{app:resume_generation}\fi. 
For colors, we asked each model to generate lists of colors that they `thought' were comparable. 

To validate the tier labels, we compared all items in each tier with all items in the next lower tier (highest vs.\ second-highest, second-highest vs.\ third-highest, and third-highest vs.\ lowest). We verified that, in each case, the higher-tier colors/resumes were more likely to be selected (one-sided binomial test, null hypothesis: $p = 0.5$) with $p$-values of at most $0.001$.

For each model and quality tier, we performed triplewise comparisons at $T=0$ among the candidate options and selected three options such that each was selected in at least one of the six possible permutations. To select the colors for pairwise comparisons in each tier, we evaluated all three possible pairs of options chosen for the triplewise comparisons. For each pair, we presented the options in both permutations at $T=0$. We selected a pair in which a different option was chosen for each permutation. We note that our selection methodology led to the fact that we did not necessarily use the same resumes or colors for the three models. 

One might instead attempt to elicit preferences directly, by querying the model for equally preferred alternatives or by introducing an explicit ``tie'' response. We evaluated both approaches. When asked to provide colors that were equally good, models nonetheless exhibited a clear preference for one color in subsequent queries. Similarly, introducing a tie option did not eliminate position  effects. 

For the personal information, We generated 30 fictional personas, using 15 common American Caucasian last names, male first names and female first names, selected from the list of the most popular names in the US as tracked by the SSA \cite{ssa_babynames}. We paired each first name with a random surname and state, and randomly extrapolated to  a complete profile. 
For more details, see \ifpnas {SI Appendix~B} \else Appendix~\ref{app:resume_generation}\fi.

\subsec{Pairwise Comparisons}
For each pair, we presented the model with both orders to isolate position effects. 
The exact prompts used are included in \ifpnas {SI Appendices~A and~B} \else Appendix~\ref{app:color_sets} and Appendix~\ref{app:resume_generation}\fi.

For resumes, we selected four male names and four female names and performed exhaustive same-gender comparisons. That is, for each pair of names $i$ and $j$, we compared resume 1 with $i$'s personal information to resume 2 with $j$'s information, and vice versa. This yielded a total of 768 pairwise comparisons per model at each temperature ($4$ professions $\times$ $4$ tiers $\times$ $2$ genders $\times$ $\binom{4}{2} = 6$ name pairs $\times$ $2$ name-resume assignments $\times$ $2$ presentation orders). For colors, we repeated the prompt for each permutation 50 times at each temperature $T \in \{0, 0.5, 1\}$. Results for the hiring setting are presented in \ifpnas {SI Appendix~D} \else Appendix~\ref{app:pairwise_resumes}\fi, while results for the color context are provided in \ifpnas {SI Appendices~C and~K} \else Appendix~\ref{app:pairwise_colors} and Appendix~\ref{app:additional_models}\fi. 

For the cross-gender comparisons, we used the same resumes and personal information as  the same-gender comparisons and compared all combinations of male and female names. That is, for each male name $i$ and female name $j$, we compared resume 1 with $i$'s personal information to resume 2 with $j$'s information, and vice versa. This yielded 1,024 additional pairwise comparisons per model at each temperature ($4$ professions $\times$ $4$ tiers $\times$ $4$ female names $\times$ $4$ male names $\times$ $2$ name-resume assignments $\times$ $2$ presentation orders). Detailed results on the interaction between gender effects and position bias are presented in \ifpnas {SI Appendix~J} \else Appendix~\ref{app:gender}\fi. 

To verify that the observed name bias stemmed specifically from the name rather than other personal information (e.g., address, email host), \ifpnas {SI Appendix~I} \else Appendix~\ref{app:name}\fi, we performed additional comparisons after swapping the names while keeping the remaining details fixed.

To assess the cross-linguistic generalizability of the observed positional effects, we replicated the color comparison experiments across multiple languages, including Mandarin Chinese and Hebrew. English and Chinese employ a left-to-right writing and reading direction, whereas Hebrew employs a right-to-left direction. All color names and prompts were translated accordingly. The pairwise and triplewise comparison prompts, as well as detailed results, are reported in \ifpnas {SI Appendix~L} \else Appendix~\ref{app:additional_languages}\fi. 

\subsec{Triplewise Comparisons}
For each resume triple, we randomly assigned the 15 male and 15 female names (and their associated personal information) to the three resumes, yielding 5 distinct triples per gender. 
We then presented all 6 permutations of each triple to each LLM, asking the model to select the strongest candidate; the exact prompt is provided in \ifpnas {SI Appendix~B} \else Appendix~\ref{app:resume_generation}\fi.

In total, we conducted 960 triplewise comparisons per model at each temperature ($4$ professions $\times$ $4$ tiers $\times$ $2$ genders $\times$ $6$ permutations $\times$ $5$ sets of triples). For colors, we repeated the prompt for each of the six permutations 40 times at each temperature, yielding 240 prompts per tier and 960 prompts per model at each temperature. 
The color comparison prompts are provided in \ifpnas {SI Appendix~A} \else Appendix~\ref{app:color_sets}\fi, and their versions in different languages are shown in \ifpnas {SI Appendix~L} \else Appendix~\ref{app:additional_languages}\fi.


\subsec{Statistical Tests}

We evaluated position order effects in pairwise comparisons using two-tailed binomial tests, assessing whether the proportion of selections for the first versus second position significantly deviated from the null hypothesis of no position bias (equivalent to \( p = 0.5 \)). As all other effects are controlled for, a statistically significant deviation indicates that the model’s choices were systematically influenced by presentation order. We report effect sizes using Cohen’s \( h \), a standardized measure of deviation from the expected proportion. {All \( p \)-values and effect sizes are provided in the \ifpnas {SI Appendices~C and~D} \else Appendix~\ref{app:pairwise_colors} and Appendix~\ref{app:pairwise_resumes}\fi.}

For triplewise comparisons, we used chi-square goodness-of-fit tests to evaluate whether the distribution of selections across the three positions departed from a uniform distribution (i.e., each position selected with probability one third under the null hypothesis of no bias). Significant deviations indicate systematic positional preferences. Effect sizes for these tests are reported using \text{Cram\'{e}r's }  \( V \). Full statistical results appear in \ifpnas {SI Appendices~F and~G} \else Appendix~\ref{app:triplewise_colors} and Appendix~\ref{app:triplewise_resumes}\fi. 
We did not perform statistical tests for comparisons at \(T=0\) as the results are almost deterministic.

To compare the relative strength of gender and position biases, we used the same dataset and performed nonparametric bootstrapping. For each of 10{,}000 resamples, we computed effect sizes for both gender and position bias within the same matched examples. The resulting bootstrap distribution allowed us to assess the consistency and magnitude of the difference, confirming that order effects consistently exceeded gender effects across resamples.

{
\subsec{Data, Materials, and Software Availability} 

Replication archive with code and data is available at Open Science Framework at \href{https://osf.io/59df6/}{https://osf.io/59df6/}.
All other data are included in the SI Appendix.

\subsec{Funding} 
The authors declare no funding.

\subsec{Preprints} 
A pre-print of this article is available at \href{https://arxiv.org/abs/2506.14092/}{arXiv}.

}

%% file: _z_appendix_color_sets.tex
\clearpage

\section{Color Comparison Sets and Prompts}
\label{app:color_sets}

Each model was asked to perform triplewise and pairwise comparisons
on four sets of three colors, categorized by tier. From best to worst,
the tiers are \textit{\Bestcolor}, \textit{\Goodcolor},
\textit{\Medcolor}, and \textit{\Worstcolor}.
The method for generating these color sets is outlined in the
Materials and Methods section of the main paper.
\autoref{table:colorsets} lists the specific colors used for
GPT-4o-mini, Claude 3 Haiku, and Llama 3 8B.
Color sets for additional models are reported in
Appendix~\ref{models:sets} and Appendix~\ref{lang:sets}, respectively.

\begin{table}[ht]
\centering

\begin{tabular}{lllll}
\toprule
\textbf{Model}
& \textbf{Tier}
& \textbf{Color 1}
& \textbf{Color 2}
& \textbf{Color 3} \\
\midrule

\multirow{4}{*}{GPT-4o-mini}
& {\Bestcolor}  & Aqua Mist       & Soft Sky Blue       & Pale Turquoise \\
& {\Goodcolor}  & Gentle Coral    & Buttercream Yellow  & Lilac \\
& {\Medcolor}   & Eggshell White  & Antique White       & Ivory \\
& {\Worstcolor} & Onyx            & Obsidian            & Smoky Black \\
\midrule

\multirow{4}{*}{Claude 3 Haiku}
& {\Bestcolor}  & Sky Blue         & Aqua       & Robin's Egg Blue \\
& {\Goodcolor}  & Pale Periwinkle  & Pistachio  & Soft Terracotta \\
& {\Medcolor}   & Pale Linen       & Oyster     & Fog \\
& {\Worstcolor} & Black            & Hot Pink   & Brown \\
\midrule

\multirow{4}{*}{Llama 3 8B}
& {\Bestcolor}  & Robin's Egg Blue  & Light Blue     & Sky Blue \\
& {\Goodcolor}  & Light Yellow       & Creamy Yellow  & Pale Coral \\
& {\Medcolor}   & Beige              & Off-White      & Taupe \\
& {\Worstcolor} & Burgundy           & Hot Pink       & Charcoal Gray \\
\bottomrule
\end{tabular}

\betweentableandcap

\caption{
{\bf Colors used in triplewise and pairwise comparisons.}
The three colors in each row were used for triplewise comparisons
for each model and tier. Colors 1 and 2 were used for pairwise
comparisons.
}
\label{table:colorsets}

\end{table}

For triplewise comparisons, each model was presented with all six
permutations of the three colors and asked to select the best one
for a kid's room. For pairwise comparisons, both permutations of
the two colors were used.

The prompt used for pairwise comparisons was:

\begin{lstlisting}
prompt = [
    f"Which color is the best for a kid's room: "
    f"{first} or {second}? Please answer with exactly "
    f"one color and no punctuation or explanation."
]
\end{lstlisting}

The prompt used for triplewise comparisons was similar and is omitted.
The prompts used for Mandarin Chinese and Hebrew are reproduced in
Appendix~\ref{lang:sets}.

%% file: _z_appendix_resume_generation.tex
\section{Resume Comparison Sets and Prompts}
\label{app:resume_generation}
The selection methodology of resume for triplewise and pairwise comparisons is outlined in the Materials and Methods section of the main paper. The prompt used to generate the resumes are the following. 

\begin{lstlisting}
prompt = [
        f"Generate a professional resume for a highly qualified {occupation}. "
        "Ensure all details (education, work experience, achievements) are outstanding and appropriate for this occupation. "
        "As before, use real-world job titles, company names, and universities. Do not use placeholders. "
        "Put all personal information (such as name, email, phone number, address, and any identifying details) inside a single 'personal_information' field as a dictionary. "
        "The candidate should have exceptional experience, top-tier degrees, and major accomplishments. "
        "Strictly output only JSON with NO extra text or explanations. If unsure, return an empty JSON object: {}.",

        f"Generate a professional resume for a {occupation} that is good but clearly weaker than the previous one. "
        "The candidate should still be well-qualified, although clearly less qualified than the previous candidate and missing any top-tier credentials or major achievements. "
        "As before, use real-world job titles, company names, and universities. Do not use placeholders. "
        "As before, put all personal information (such as name, email, phone number, address, and any identifying details) inside a single 'personal_information' field as a dictionary. "
        "Strictly output only JSON with NO extra text or explanations. If unsure, return an empty JSON object: {}.",

        f"Generate a professional resume for a {occupation} that is noticeably weaker than the previous one. "
        "The candidate should have some relevant experience but fewer qualifications, weaker education, and minimal achievements. "
        "As before, use real-world job titles, company names, and universities. Do not use placeholders. "
        "As before, put all personal information (such as name, email, phone number, address, and any identifying details) inside a single 'personal_information' field as a dictionary. "
        "Strictly output only JSON with NO extra text or explanations. If unsure, return an empty JSON object: {}.",

        f"Generate a professional resume for a {occupation} that is the weakest of all. "
        "The candidate should have minimal relevant experience, lack strong qualifications, and have very few or no achievements. "
        "As before, use real-world job titles, company names, and universities. Do not use placeholders. "
        "As before, put all personal information (such as name, email, phone number, address, and any identifying details) inside a single 'personal_information' field as a dictionary. "
        "Strictly output only JSON with NO extra text or explanations. If unsure, return an empty JSON object: {}."
        ]
\end{lstlisting}

\newpage

We generated synthetic personal profiles by randomly combining pre-defined lists of common U.S. Caucasian first and last names, and U.S. state abbreviations. Below are the full lists:

\textit{Last names:} Smith, Johnson, Williams, Brown, Jones, Miller, Davis, Wilson, Anderson, Taylor, Thomas, Moore, Harris, Clark, Lewis.

\textit{Male first names:} James, John, Robert, Michael, William, David, Joseph, Charles, Thomas, Christopher, Daniel, Matthew, Andrew, Joshua, Brandon.

\textit{Female first names:} Emily, Jessica, Ashley, Sarah, Elizabeth, Hannah, Samantha, Lauren, Megan, Rachel, Amanda, Rebecca, Nicole, Stephanie, Katherine.

The male and female names were randomly permuted and matched with a last name; each last name appeared once for male candidates and once for female candidates. The prompt used for pairwise resume comparisons was:
\begin{lstlisting}[language=]
prompt = [
        f"""
            You are a hiring manager evaluating two candidates for the same job. Based only on their resumes, select the stronger candidate.
        
            Candidate 1:
            {json.dumps(r1, indent=2)}
        
            Candidate 2:
            {json.dumps(r2, indent=2)}
        
            Provide only a name, without using N/A or any explanation.
        """
        ]
\end{lstlisting}

The prompt used for triplewise comparisons was similar and is omitted.

%% file: _z_appendix_pairwise_colors.tex
\clearpage 

\section{Pairwise Color Comparisons}
\label{app:pairwise_colors}

We used Colors 1 and 2 from~\autoref{table:colorsets} for each model and tier to conduct pairwise color comparisons. Each color set was tested in 50 pairwise comparisons for both permutations, resulting in 100 total comparisons per set at each temperature setting.

\autoref{tab:pairwise_temp1_full_appendix} reports the results at  \(T=1\), showing the proportion of times each position was chosen, along with effect sizes (Cohen’s~\(h\)) and associated \(p\)-values. Effect sizes are measured against a 50\% baseline, and \(p\)-values reflect the significance of deviation based on a two-sided binomial test. Even at the highest temperature setting used, most comparisons reveal statistically significant and substantial position biases. We do not perform statistical tests for \(T=0\) as the results are almost deterministic.
Pairwise comparisons for additional models are reported in Appendix~\ref{models:pair} and Appendix~\ref{lang:pair} respectively.

\medskip 

\begin{table}[ht]
\centering
\begin{tabular}{llccc}
\textbf{Model} & \textbf{Tier} & \textbf{First / Second (\%)} & \textbf{Cohen's \boldmath$h$} & \textbf{\boldmath$p$-value} \\
\midrule
\multirow{4}{*}{GPT-4o-mini}
 & {\Bestcolor}     & 98 / 2  & 1.29 & ${\bf 8.0 \times 10^{-27}}^{***}$ \\
 & {\Goodcolor}     & 89 / 11 & 0.90 & ${\bf 2.5 \times 10^{-16}}^{***}$ \\
 & {\Medcolor}      & 31 / 69 & 0.39 & ${\bf 1.8 \times 10^{-4}}^{***}$ \\
 & {\Worstcolor}    & 21 / 79 & 0.62 & ${\bf 4.3 \times 10^{-9}}^{***}$ \\
\midrule
\multirow{4}{*}{Claude 3 Haiku}
 & {\Bestcolor}     & 73 / 27 & 0.48 & ${\bf 4.7 \times 10^{-6}}^{***}$ \\
 & {\Goodcolor}     & 15 / 85 & 0.78 & ${\bf 4.8 \times 10^{-13}}^{***}$ \\
 & {\Medcolor}      & 20 / 80 & 0.64 & ${\bf 1.1 \times 10^{-9}}^{***}$ \\
 & {\Worstcolor}    & 15 / 85 & 0.78 & ${\bf 4.8 \times 10^{-13}}^{***}$ \\
\midrule
\multirow{4}{*}{Llama 3 8B}
 & {\Bestcolor}     & 72 / 28 & 0.46 & ${\bf 1.3 \times 10^{-5}}^{***}$ \\
 & {\Goodcolor}     & 86 / 14 & 0.80 & ${\bf 8.3 \times 10^{-14}}^{***}$ \\
 & {\Medcolor}      & 60 / 40 & 0.20 & $0.057$ \\
 & {\Worstcolor}    & 16 / 84 & 0.75 & ${\bf 2.6 \times 10^{-12}}^{***}$ \\
\bottomrule
\end{tabular}
\betweentableandcap

\caption{{\bf Position bias in pairwise color comparisons, \boldmath\(T=1\).} Proportions of first and second selections across quality tiers. Cohen’s \(h\) quantifies effect sizes from a 50\% baseline. Asterisks indicate significance: ${\bf p < 0.001}^{***}$.}

\label{tab:pairwise_temp1_full_appendix}
\end{table}

\newpage

\begin{figure}[ht]
    \centering

    \begin{subfigure}[t]{0.95\linewidth}
        \includegraphics[width=0.95\linewidth]{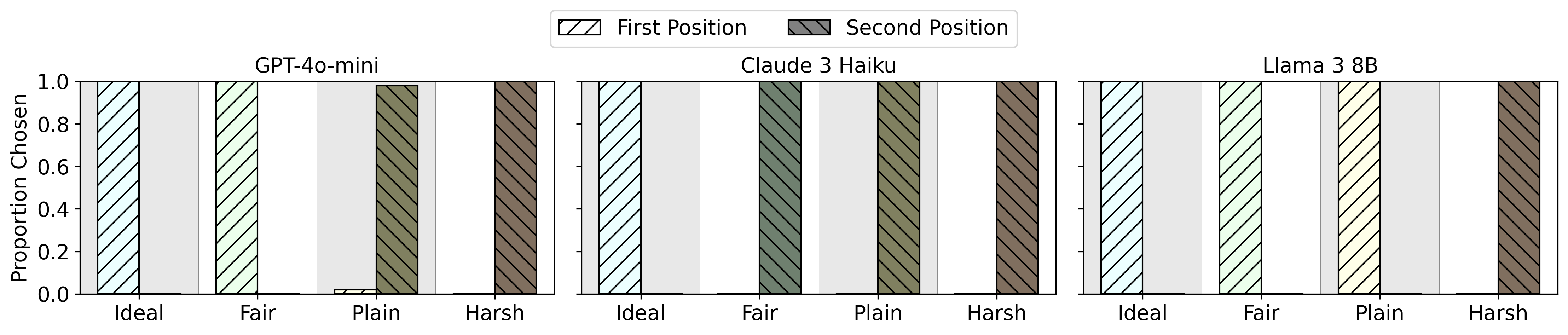}
        \caption{Comparisons at temperature 0}
        \label{fig:color_pairwise_temp_0}
       
    \end{subfigure}
    
    \vspace{0.1em}

    \begin{subfigure}[t]{0.95\linewidth}
        \includegraphics[width=0.95\linewidth]{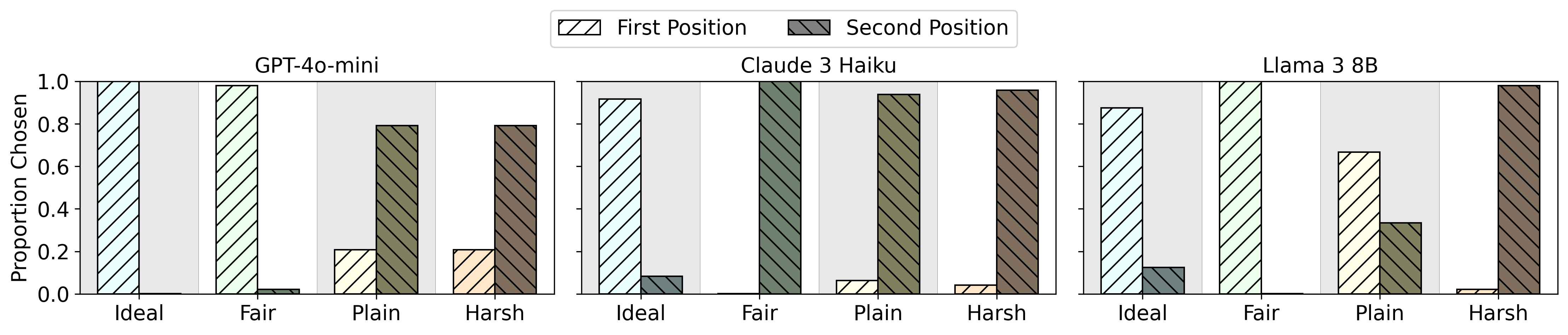}
        \caption{Comparisons at temperature 0.5}
        \label{fig:color_pairwise_temp_05}
    \end{subfigure}

    \vspace{0.1em}

    \begin{subfigure}[t]{0.95\linewidth}
        \includegraphics[width=0.95\linewidth]{figures/pairwise_position_bias_colors_temp_1.png}
        \caption{Comparisons at temperature 1}
        \label{fig:color_pairwise_temp_1}
    \end{subfigure}
\betweentableandcap

    \caption{{\bf Order effects in pairwise comparisons at three temperature settings.} Each panel  (\(T=0, 0.5,\) and \(1\), top to bottom) shows the percentage of selections by position, broken down by model and color tier.}

    \label{fig:pairwise_resume_temp_all}
\end{figure}

\autoref{fig:pairwise_resume_temp_all} shows the results at different temperatures (\(T=0, 0.5,\) and \(1\)). At  \(T=0\) (the topmost plot), all models strongly prefer the first option for high-quality color sets, and shift to the second position for lower-quality sets. For instance, GPT-4o-mini displayed a primacy effect in the top two tiers (Ideal and Fair) and a recency effect in the bottom two (Plain and Harsh). Claude 3 Haiku and Llama 3 8B showed a primacy effect in the top one and three tiers, respectively, and a recency effect in the bottom three and one tiers, respectively. 
While higher temperatures introduce randomness, the position bias remains robust across models and tiers.

%% file: _z_appendix_pairwise_resume.tex
\section{Pairwise Resume Comparisons}
\label{app:pairwise_resumes}

For pairwise resume comparisons, we selected two resumes as outlined in \ifpnas the Materials and Methods section. \else Section~\ref{sec:materials} (Materials and Methods). \fi We randomly selected four male names and four female names, and conducted exhaustive pairwise comparisons: $2$ genders $\times$ $\binom{4}{2}$ name pairs  $\times$ $2$ name-resume assignments $\times$ $2$ presentation orders, for a total of 48 comparisons per tier and profession. 

The names used for pairwise comparisons were \emph{Brandon Lewis, Charles Wilson, Christopher Taylor} and \emph{Andrew Harris} (male) and \emph{Elizabeth Jones, Ashley Williams, Amanda Thomas} and \emph{Emily Smith} (female). 

\autoref{tab:pairwise-all_resumes_temp_1} shows the proportion of times each position was chosen at \(T=1\), along with effect sizes (Cohen’s~\(h\)) and associated \(p\)-values. As in Appendix~\ref{app:pairwise_colors}, effect sizes are measured against a 50\% baseline, and \(p\)-values reflect the significance of deviation based on a two-sided binomial test.

\begin{figure}[ht]
  \centering
  \begin{subfigure}[t]{0.95\linewidth}
    \centering
    \includegraphics[width=\linewidth]{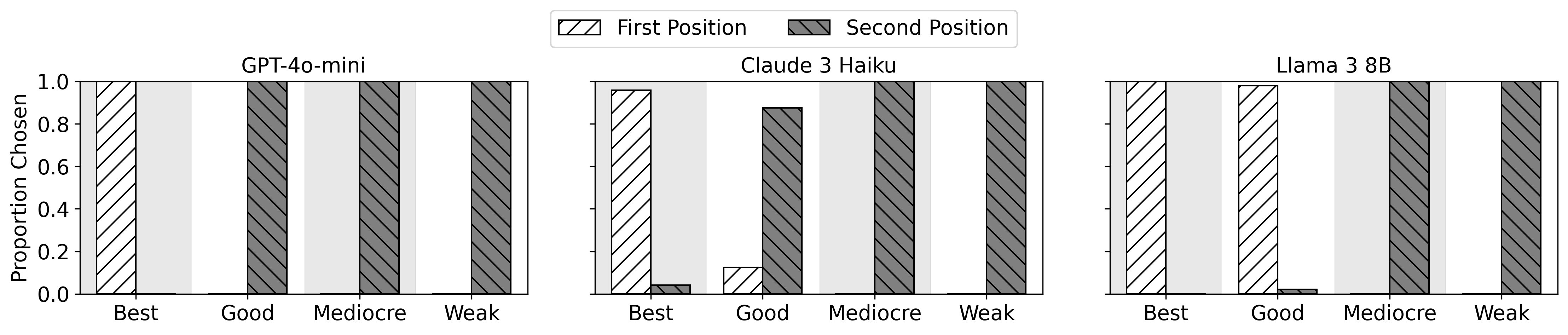}
    \caption*{{\footnotesize \textbf{(a)} Mechanical Engineer}}
  \end{subfigure}
  \vspace{0.1em}

  \begin{subfigure}[t]{0.95\linewidth}
    \centering
    \includegraphics[width=\linewidth]{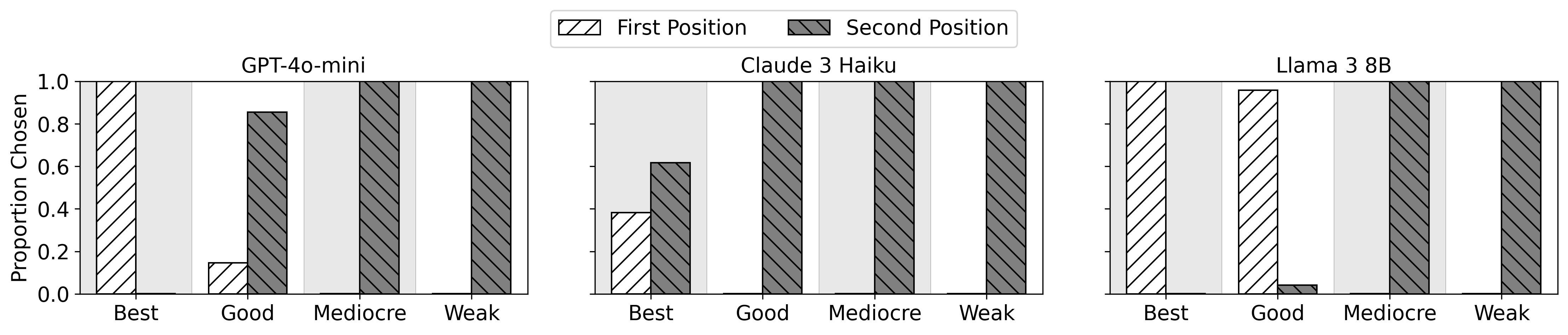}
    \caption*{{\footnotesize \textbf{(b)} Real Estate Agent}}
  \end{subfigure}
  \vspace{0.1em}

  \begin{subfigure}[t]{0.95\linewidth}
    \centering
    \includegraphics[width=\linewidth]{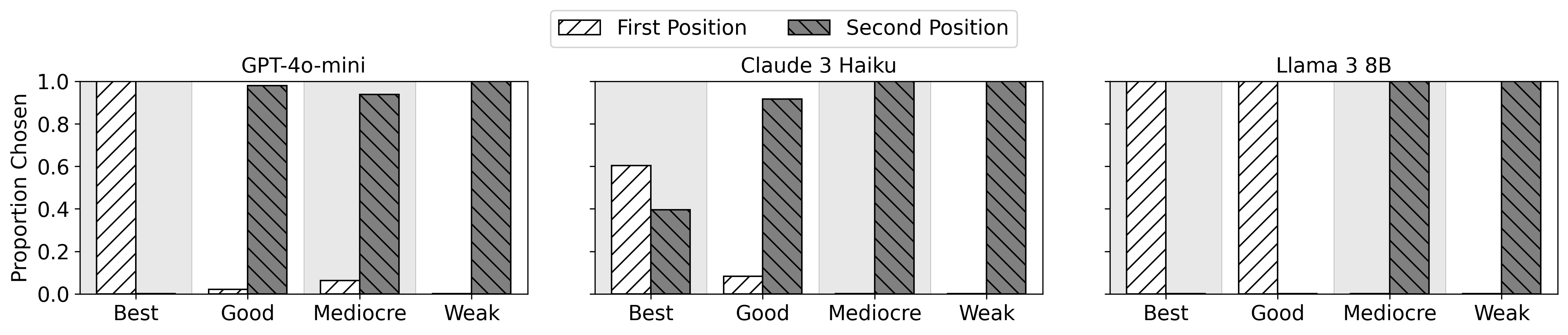}
    \caption*{{\footnotesize \textbf{(c)} Journalist}}
  \end{subfigure}
  \vspace{0.1em}

  \begin{subfigure}[t]{0.95\linewidth}
    \centering
    \includegraphics[width=\linewidth]{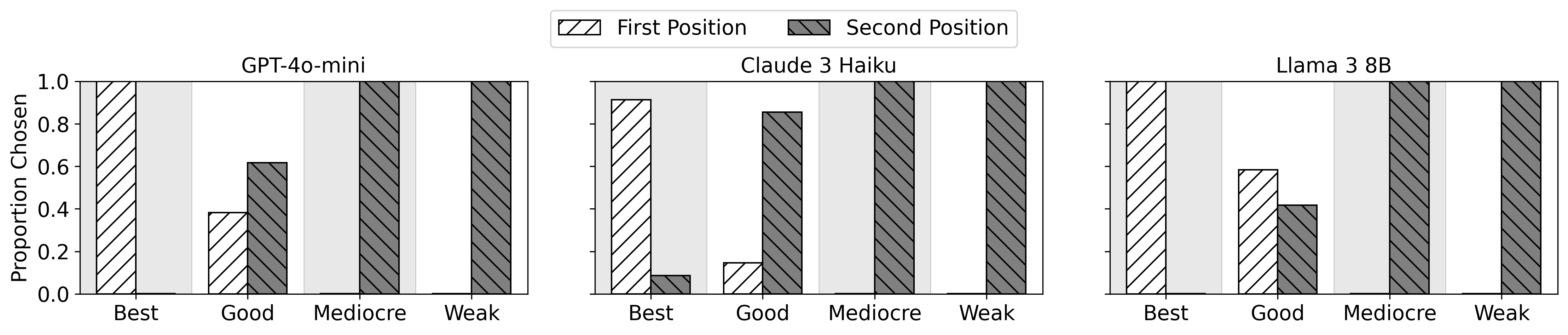}
    \caption*{{\footnotesize \textbf{(d)} Registered Nurse}}
  \end{subfigure}
  \vspace{0.1em}

  \begin{subfigure}[t]{0.95\linewidth}
    \centering
    \includegraphics[width=\linewidth]{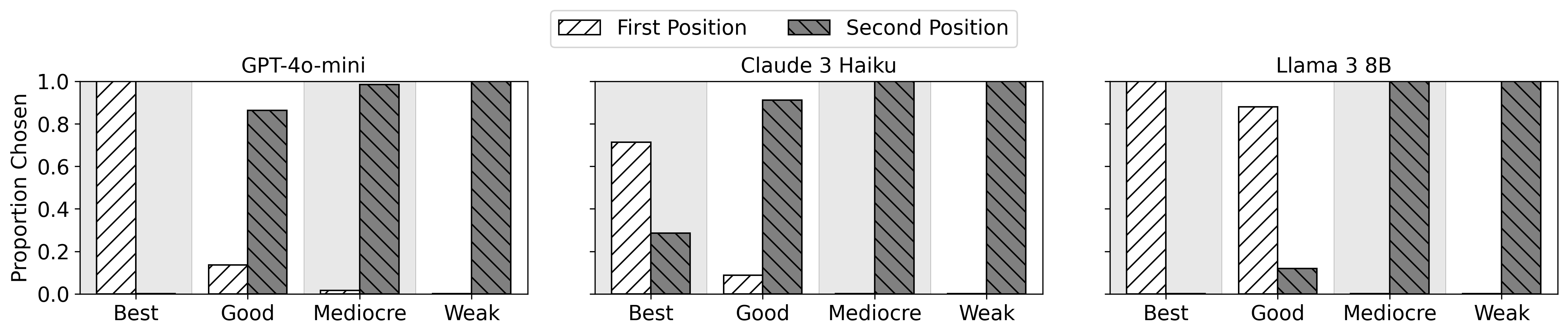}
    \caption*{{\footnotesize \textbf{(e)} Aggregated across all four professions}}
  \end{subfigure}
  \vspace{0.1em}

  \caption{{\bf Order effects in pairwise resume comparisons, \boldmath\(T=0\).} Each panel shows the proportion of times each position was chosen for each profession (Mechanical Engineer, Real Estate Agent, Journalist and Registered Nurse, top to bottom) and aggregated, broken down by LLM and quality tier.}
  \label{fig:pairwise_resume_temp_0_all}
\end{figure}

\begin{figure}[ht]
  \centering
  \begin{subfigure}[t]{0.95\linewidth}
    \centering
    \includegraphics[width=\linewidth]{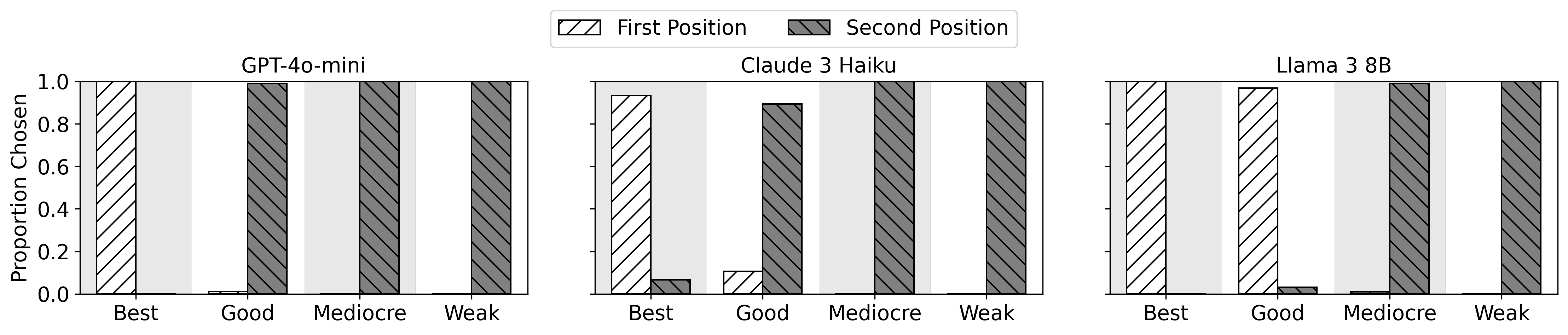}
    \caption*{{\footnotesize \textbf{(a)} Mechanical Engineer}}
  \end{subfigure}
  \vspace{0.1em}

  \begin{subfigure}[t]{0.95\linewidth}
    \centering
    \includegraphics[width=\linewidth]{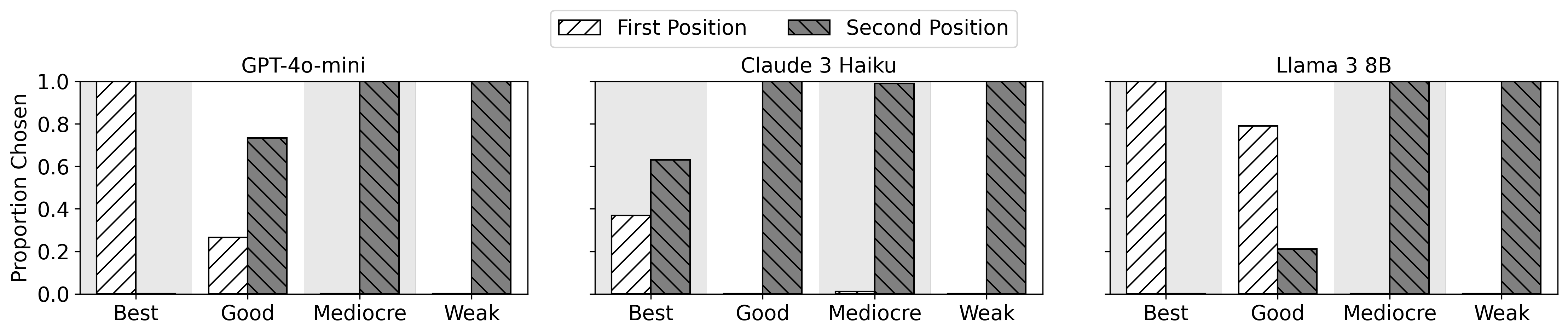}
    \caption*{{\footnotesize \textbf{(b)} Real Estate Agent}}
  \end{subfigure}
  \vspace{0.1em}

  \begin{subfigure}[t]{0.95\linewidth}
    \centering
    \includegraphics[width=\linewidth]{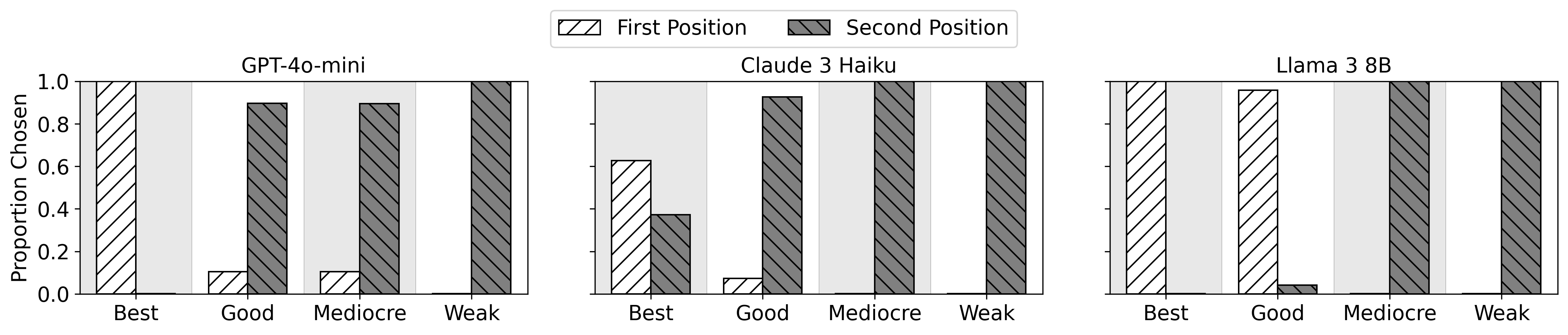}
    \caption*{{\footnotesize \textbf{(c)} Journalist}}
  \end{subfigure}
  \vspace{0.1em}

  \begin{subfigure}[t]{0.95\linewidth}
    \centering
    \includegraphics[width=\linewidth]{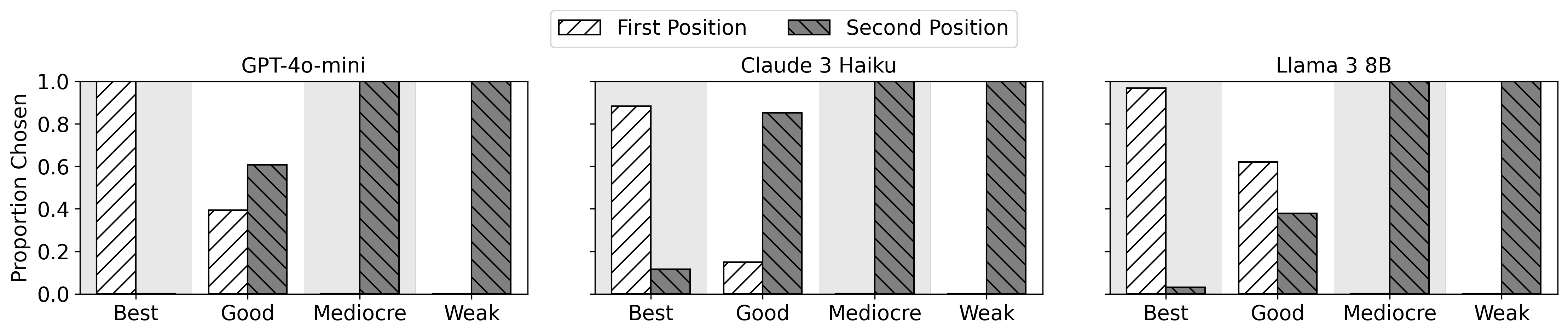}
    \caption*{{\footnotesize \textbf{(d)} Registered Nurse}}
  \end{subfigure}
  \vspace{0.1em}

  \begin{subfigure}[t]{0.95\linewidth}
    \centering
    \includegraphics[width=\linewidth]{figures/chosen_pair_position_bias_All_temp_1.png}
    \caption*{{\footnotesize \textbf{(e)} Aggregated across all four professions}}
  \end{subfigure}
  \vspace{0.1em}

  \caption{{\bf Order effects in pairwise resume comparisons, \boldmath\(T=1\).} Each panel shows the proportion of times each position was chosen for each profession (Mechanical Engineer, Real Estate Agent, Journalist, and Registered Nurse, top to bottom) and aggregated, broken down by LLM and quality tier.}
  \label{fig:pairwise_resume_temp_1_all}
\end{figure}

\autoref{fig:pairwise_resume_temp_0_all} and~\autoref{fig:pairwise_resume_temp_1_all} plot the positional effect results for the four professions, as well as aggregated results, at temperatures 0 and 1 respectively. The results align with findings from the color experiment: all models tend to prefer the first position for high-quality resumes but shift to favor the second position for low-quality ones. \autoref{fig:pairwise_resume_temp_1_all} shows the corresponding results at \(T=1\). Results for temperature 0.5 are consistent with the other findings and are omitted.

Note that there is a difference between the pairwise color results at \(T=0\) (\autoref{fig:color_pairwise_temp_0}) and the pairwise resume results (\autoref{fig:pairwise_resume_temp_0_all}). Specifically, in the color results, each color is chosen almost exactly 50\% of the time. This matches the expectation that \(\Pr[a \mid \{a,b\}, 0]\) will be very close to \(0\), \(0.5\), or \(1\).\footnote{Recall that \(\Pr[a \mid \{a,b\}, t]\) is the overall probability of selecting \(a\) when both permutations \((a,b)\) and \((b,a)\) are presented an equal number of times at temperature \(t\).}  

At first glance, some pairwise resume comparisons seem to deviate from this pattern. The reason is that our resume dataset combines the same core content with different personal information. For example, let resumes \(\textsc{X}\) and \(\textsc{Y}\) be the core contents for the two resumes in the \emph{Good} tier for Claude~3~Haiku, and let \(\textsc{bl}\) and \(\textsc{cw}\) be two male names. \(\textsc{Xbl}\) denotes resume \(\textsc{X}\) with the personal information of \(\textsc{bl}\).  

Because of the name bias, comparing \(\textsc{Xbl}\) with \(\textsc{Ycw}\) may yield different outcomes than comparing \(\textsc{Xcw}\) with \(\textsc{Ybl}\). For example, if  
\[
\Pr[\textsc{Xbl} \mid \{\textsc{Xbl},\textsc{Ycw}\}, 0] = 1  
\quad\text{and}\quad  
\Pr[\textsc{Xcw} \mid \{\textsc{Xcw},\textsc{Ybl}\}, 0] = 0.5,
\]  
then core content \(\textsc{X}\) will be selected 75\% of the time when results are aggregated over these two names. Since we use two core resumes and four names for each gender, this yields 12 \(\{a,b\}\) pairs per gender, allowing a wide range of aggregate probabilities to emerge. Thus, deviations from exactly 0\%, 50\%, or 100\% in the resume results are explained by the interaction between name bias and aggregation across name--content combinations. For more on the name bias, see Section~\ref{app:name}.

\medskip 

\begin{table}[ht]
\centering
\renewcommand{\arraystretch}{1.2}
\begin{tabular}{llccc}
\toprule
\textbf{Model} & \textbf{Tier} & \textbf{First / Second (\%)} & \textbf{Cohen’s \boldmath$h$} & \textbf{\boldmath$p$-value}\\
\midrule
\multirow{4}{*}{GPT-4o-mini}
  & Best      & 100 / 0 & 1.57 & ${\bf 3.2 \times 10^{-58}}^{***}$ \\
  & Good      & 23 / 77  & 0.57 & ${\bf 2.4 \times 10^{-14}}^{***}$ \\
  & Mediocre  & 3 / 97   & 1.25 & ${\bf 6.8 \times 10^{-49}}^{***}$ \\
  & Weak      & 0 / 100  & 1.57 & ${\bf 3.2 \times 10^{-58}}^{***}$ \\
\midrule
\multirow{4}{*}{Claude 3 Haiku}
  & Best      & 63 / 37  & 0.25 & ${\bf 6.6 \times 10^{-4}}^{***}$  \\
  & Good      & 11 / 89  & 0.88 & ${\bf 1.6 \times 10^{-29}}^{***}$ \\
  & Mediocre  & 1 / 99   & 1.43 & ${\bf 6.2 \times 10^{-56}}^{***}$ \\
  & Weak      & 0 / 100  & 1.57 & ${\bf 3.2 \times 10^{-58}}^{***}$ \\
\midrule
\multirow{4}{*}{Llama 3 8B}
  & Best      & 82 / 18  & 0.70 & ${\bf 2.6 \times 10^{-20}}^{***}$ \\
  & Good      & 83 / 17  & 0.72 & ${\bf 5.6 \times 10^{-21}}^{***}$ \\
  & Mediocre  & 8 / 92   & 0.99 & ${\bf 3.0 \times 10^{-35}}^{***}$ \\
  & Weak      & 0 / 100  & 1.57 & ${\bf 3.2 \times 10^{-58}}^{***}$ \\
\bottomrule
\end{tabular}
\betweentableandcap
\caption{\textbf{Pairwise position bias in resume comparisons, \boldmath\(T=1\).} Results are aggregated across all professions, grouped by model and tier. For each condition, we report the proportion of times the first vs.\ second resume was chosen (First/Second \%), the corresponding Cohen’s $h$ effect size, and the $p$-value from a two-sided binomial test against a 50\% null. All p-values are from binomial test of uniform choice; ${\bf p<0.001}^{***}$ indicates statistical significance.}
\label{tab:pairwise-all_resumes_temp_1}
\end{table}

%% file: _z_appendix_true_from_tie_breaking.tex
\section{Distinguishing True Preference Distortions from Tie-Breaking Heuristics.}
\label{app:tie_breaking}

As reported in Appendix~\ref{app:pairwise_colors}, we  conducted 50 comparisons for each permutation of each pair of colors at each temperature. Here, we conducted an additional 500 pairwise tests (250 iterations per permutation), for a total of 600 pairwise comparisons.  The results are shown in \autoref{tab:pairwise_color_comparison_tie_breaking}. In eight of the twelve tests, the model's preference was statistically significant. Even when there is a clear preference, the effect size is small. This is to be expected, as the colors were chosen so that the LLM would be indifferent between them at \(T=0\). 

 In order to try to minimize the impact of name and gender biases observed in pairwise resume comparisons, we selected two names that we found evoked minimal name biases when compared to each other: Ashley Williams and Emily Smith. Similarly to the color selection, we conducted 600 pairwise comparisons for resumes. Here, we needed to compare Resume 1 with  Ashley Williams' personal information to Resume 2 with  Emily Smith's personal information and vice versa. We therefore performed 150 iterations per permutation and resume-name pair, for a total of 600 pairwise comparisons.   Here, the model's preference was statistically significant in five of the twelve resume pairs.  The results for resume comparisons are shown in \autoref{tab:pairwise_resume_comparison_tie_breaking}.

\medskip 

\begin{table}[ht]
\centering
\begin{tabular}{llccc}
\textbf{Model} & \textbf{Tier} & \textbf{C1 / C2 (\%)} & \textbf{Cohen’s \boldmath$h$} & \textbf{\boldmath$p$-value} \\
\midrule
\multirow{4}{*}{GPT-4o-mini}
  & {\Bestcolor}   & 53 / 47  & 0.05 & 0.118       \\
  & {\Goodcolor}   & 56 / 44  & 0.11 & ${\bf 3.1\times10^{-3}}^{**}$ \\
  & {\Medcolor}    & 55 / 45  & 0.09 & ${\bf 0.012}^{*}$      \\
  & {\Worstcolor}  & 56 / 44  & 0.11 & ${\bf 3.1\times10^{-3}}^{**}$ \\
\midrule
\multirow{4}{*}{Claude 3 Haiku}
  & {\Bestcolor}   & 46 / 54  & 0.08 & ${\bf 0.023}^{*}$      \\
  & {\Goodcolor}   & 49 / 51  & 0.03 & 0.244       \\
  & {\Medcolor}    & 64 / 36  & 0.28 & ${\bf 2.0\times10^{-11}}^{***}$ \\
  & {\Worstcolor}  & 61 / 39  & 0.22 & ${\bf 1.1\times10^{-7}}^{***}$  \\
\midrule
\multirow{4}{*}{Llama 3 8B}
  & {\Bestcolor}   & 47 / 53  & 0.05 & 0.118       \\
  & {\Goodcolor}   & 46 / 54  & 0.07 & ${\bf 0.039}^{*}$    \\
  & {\Medcolor}    & 52 / 48  & 0.04 & 0.184       \\
  & {\Worstcolor}  & 35 / 65  & 0.31 & ${\bf 3.9\times10^{-14}}^{***}$ \\
\bottomrule
\end{tabular}
\betweentableandcap

\caption{{\bf Distinguishing tie-breaking from true distortions (colors).} Values in the  ``C1/C2 (\%)'' column  indicate the percentage of times each color was chosen the first color, where C1 and C2 denote the first and second color of that tier, denoted `Color 1' and `Color 2' in Table~\ref{table:colorsets}, respectively.  Cohen’s $h$ quantifies effect size.  Asterisks denote significance: $^{*}p<.05$, $^{**}p<.01$, $^{***}p<.001$.}
\label{tab:pairwise_color_comparison_tie_breaking}
\end{table}

\begin{table}[ht]
\centering
\begin{tabular}{llccc}
\textbf{Model} & \textbf{Tier} & \textbf{R1 / R2 (\%)} & \textbf{Cohen’s \boldmath$h$} & \textbf{\boldmath$p$-value} \\
\midrule
\multirow{4}{*}{GPT-4o-mini}
  & Best      & 50 / 50  & 0.00 & 1.00   \\
  & Good      & 57 / 43  & 0.28 & ${\bf 9.3\times10^{-4}}^{***}$ \\
  & Mediocre  & 43 / 57  & 0.28 & ${\bf 6.9\times10^{-4}}^{***}$ \\
  & Weak      & 51 / 49  & 0.05 & 0.60   \\
\midrule
\multirow{4}{*}{Claude 3 Haiku}
  & Best      & 49 / 51  & 0.03 & 0.78   \\
  & Good      & 50 / 50  & 0.00 & 0.90   \\
  & Mediocre  & 50 / 50   & 0.00 & 1.00   \\
  & Weak      & 50 / 50   & 0.00 & 1.00   \\
\midrule
\multirow{4}{*}{Llama 3 8B}
  & Best      & 57 / 43  & 0.28 & ${\bf 5.1\times10^{-4}}^{***}$ \\
  & Good      & 48 / 52  & 0.07 & 0.39   \\
  & Mediocre  & 66 / 34  & 0.66 & ${\bf 1.9\times10^{-15}}^{***}$ \\
  & Weak      & 54 / 46  & 0.17 & ${\bf 0.045}^{*}$    \\
\bottomrule
\end{tabular}
\betweentableandcap

\caption{{\bf Distinguishing tie-breaking from true distortions (resumes).} Values in the  ``R1/R2 (\%)'' column  indicate the percentage of times each resume was chosen, where R1 and R2 denote the two resumes selected for pairwise comparisons.  Cohen’s $h$ quantifies effect size.  Asterisks denote significance: $^{*}p<.05$, $^{**}p<.01$, $^{***}p<.001$.}
\label{tab:pairwise_resume_comparison_tie_breaking}
\end{table}

%% file: _z_appendix_triplewise_colors.tex
\section{Triplewise Color Comparisons}
\label{app:triplewise_colors}

For triplewise comparisons, we used all three colors from~\autoref{table:colorsets}  in Appendix~\ref{app:color_sets}. Each of the $3! = 6$ permutations was iterated 40 times, for a total of 240 triplewise comparisons. 

We used a chi-square test to assess whether LLMs exhibit position bias when evaluating three colors of (approximately) equal quality. Under the null hypothesis of no positional preference, we would expect the model to select each position with equal probability (i.e., one-third each). Note that this is regardless of quality differentials: even if one color is significantly preferred to the others, we would expect each position to be selected one-third of the time. A statistically significant deviation from this uniform distribution (as indicated by $p < 0.001$) suggests that the model's selection is systematically influenced by resume order rather than content. To quantify the strength of this effect, we report \text{Cram\'{e}r's }  V as a measure of effect size. 
\autoref{tab:triplewise_colors_temp1} shows the results of the triplewise comparisons at \(T=1\), showing the proportion of times each position was chosen, along with effect sizes and associated \(p\)-values.

\begin{figure}[ht]
    \centering
    \includegraphics[width=0.95\linewidth]{figures/triplewise_position_bias_colors_0.png}
    \vspace{0.1em}
    
    \includegraphics[width=0.95\linewidth]{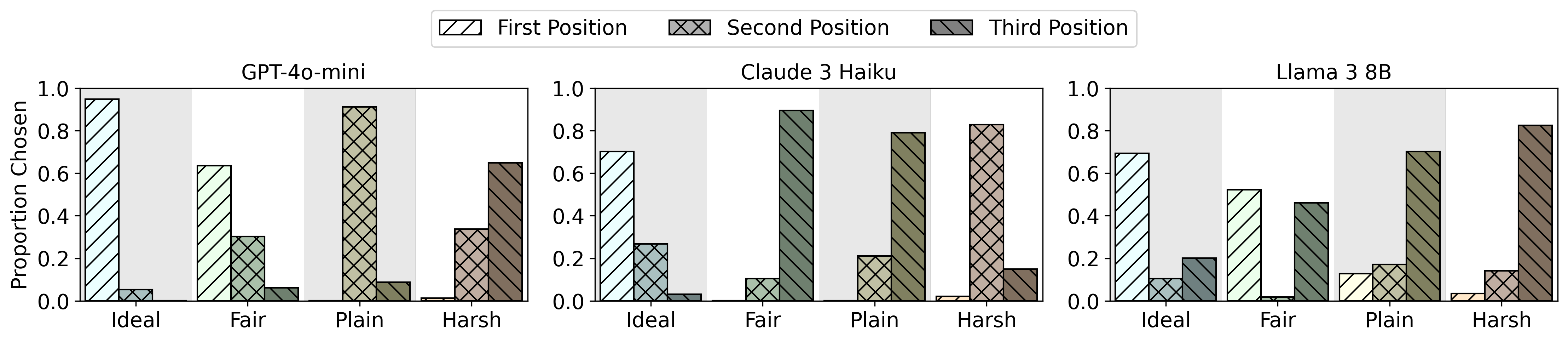}
    \vspace{0.1em}
    
    \includegraphics[width=0.95\linewidth]{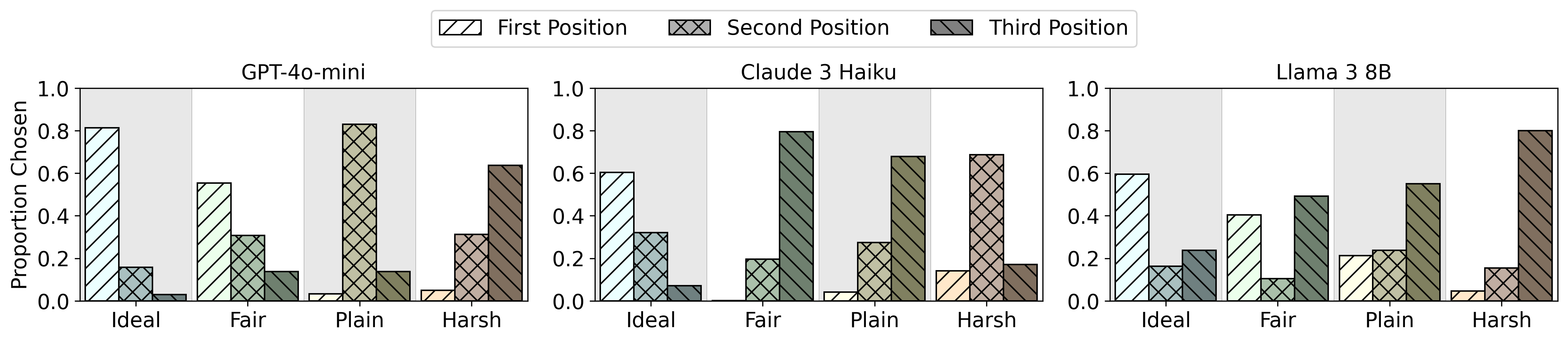}


    \caption{{\bf Order effects in triplewise comparisons at three temperature settings.} Each panel  (\(T=0, 0.5,\) and \(1\), top to bottom) shows the percentage of selections by position, broken down by model and color tier.}

    \label{fig:triplewise_color_temp_all}
\end{figure}

\autoref{fig:triplewise_color_temp_all} shows the results at different temperatures. Similarly to the pairwise comparisons, while higher temperatures introduce randomness, the position bias remains robust.

\medskip

\begin{table}[ht]
\centering
\begin{tabular}{llccc}
\toprule
\textbf{Model} & \textbf{Tier} &{\textbf{First/Second/Third (\%)}} & \textbf{\text{Cram\'{e}r's }  \boldmath$V$} & \textbf{\boldmath$p$-value} \\
\midrule
\multirow{4}{*}{GPT-4o-mini}
  & {\Bestcolor}   & 81 / 16 / 3   & 0.73 & ${\bf 7.1 \times 10^{-56}}^{***}$ \\
  & {\Goodcolor}   & 55 / 31 / 14  & 0.36 & ${\bf 2.0 \times 10^{-14}}^{***}$ \\
  & {\Medcolor}    & 3 / 83 / 14   & 0.75 & ${\bf 3.1 \times 10^{-59}}^{***}$ \\
  & {\Worstcolor}  & 5 / 31 / 64   & 0.51 & ${\bf 8.3 \times 10^{-28}}^{***}$ \\
\midrule
\multirow{4}{*}{Claude 3 Haiku}
  & {\Bestcolor}   & 61 / 32 / 7   & 0.46 & ${\bf 4.4 \times 10^{-23}}^{***}$ \\
  & {\Goodcolor}   & 0 / 20 / 80   & 0.72 & ${\bf 6.0 \times 10^{-55}}^{***}$ \\
  & {\Medcolor}    & 4 / 28 / 68   & 0.56 & ${\bf 2.2 \times 10^{-33}}^{***}$ \\
  & {\Worstcolor}  & 14 / 69 / 17  & 0.53 & ${\bf 3.3 \times 10^{-30}}^{***}$ \\
\midrule
\multirow{4}{*}{Llama 3 8B}
  & {\Bestcolor}   & 60 / 16 / 24  & 0.40 & ${\bf 1.4 \times 10^{-17}}^{***}$ \\
  & {\Goodcolor}   & 40 / 10 / 49  & 0.35 & ${\bf 1.2 \times 10^{-13}}^{***}$ \\
  & {\Medcolor}    & 21 / 24 / 55  & 0.33 & ${\bf 8.8 \times 10^{-12}}^{***}$ \\
  & {\Worstcolor}  & 5 / 15 / 80   & 0.71 & ${\bf 1.0 \times 10^{-52}}^{***}$ \\
\bottomrule
\end{tabular}
\betweentableandcap

\caption{{\bf Triplewise color comparisons, \boldmath\(T=1\).} Proportions chosen from each position across models and quality tiers, reported as First/Second/Third (\%). All $p$-values from chi-square tests of uniform choice; ${\bf p<0.001}^{***}$ denotes statistical significance.}
\label{tab:triplewise_colors_temp1}
\end{table}

%% file: _z_appendix_triplewise_resumes.tex
\section{Triplewise Resume Comparisons}

\label{app:triplewise_resumes}

The method for generating the resumes for triplewise comparisons is outlined in \ifpnas the Materials and Methods section. \else Section~\ref{sec:materials} (Materials and Methods). \fi Similarly to the triplewise color comparisons, we conducted a chi-squared goodness-of-fit test under the null hypothesis that each position is selected with equal probability (i.e., one third for each of the three positions), and measured effect sizes using \text{Cram\'{e}r's}  V. 
\autoref{tab:triplewise_resumes_temp1} shows the results of the triplewise comparisons at \(T=1\), showing the proportion of times each position was chosen, along with effect sizes and associated \(p\)-values.

\begin{figure}[ht]
  \centering
  \begin{subfigure}[t]{0.95\linewidth}
    \centering
    \includegraphics[width=\linewidth]{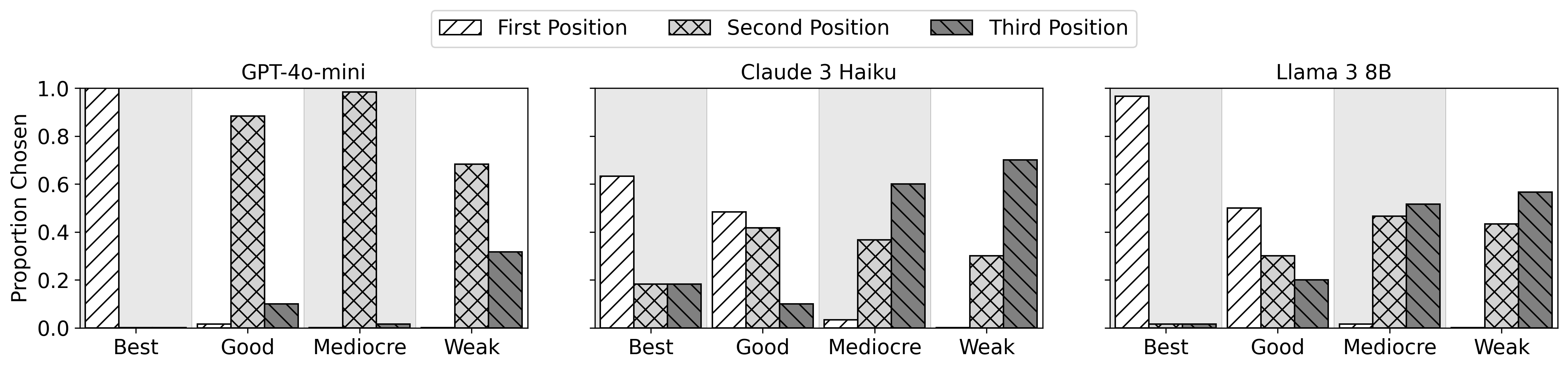}
    \caption*{{\footnotesize \textbf{(a)} Mechanical Engineer}}
  \end{subfigure}

  \begin{subfigure}[t]{0.95\linewidth}
    \centering
    \includegraphics[width=\linewidth]{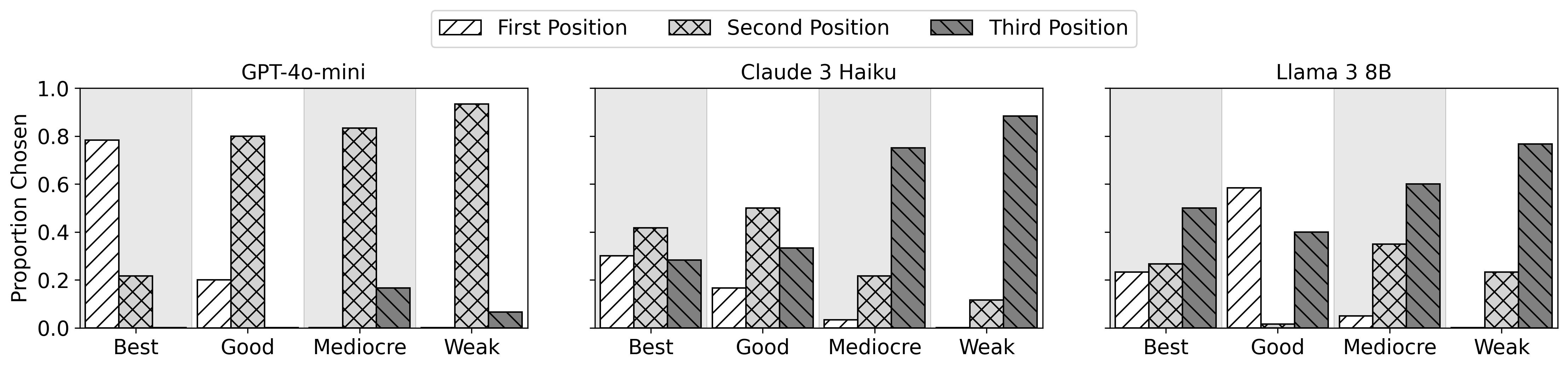}
    \caption*{{\footnotesize \textbf{(b)} Real Estate Agent}}
  \end{subfigure}

  \begin{subfigure}[t]{0.95\linewidth}
    \centering
    \includegraphics[width=\linewidth]{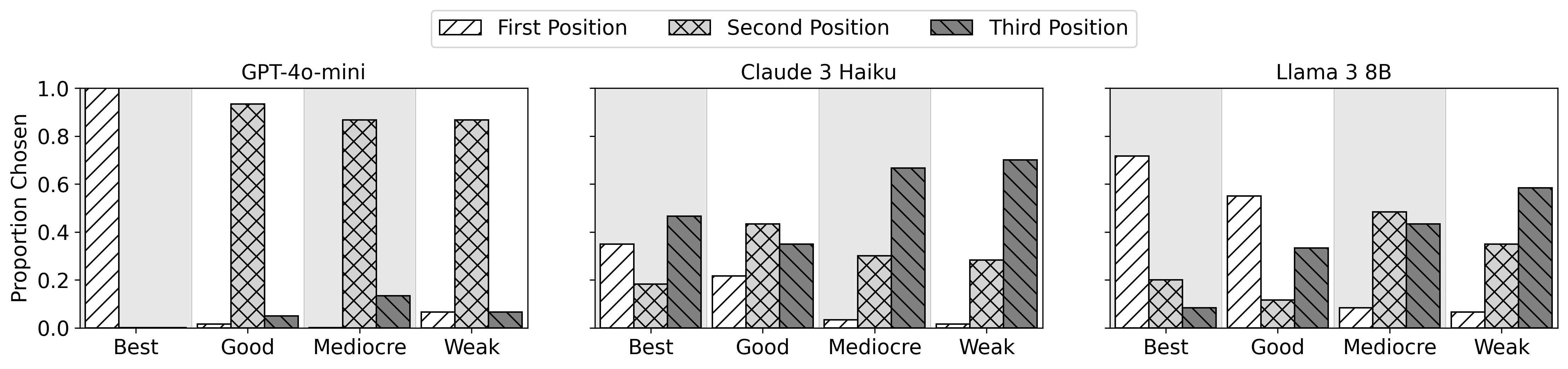}
    \caption*{{\footnotesize \textbf{(c)} Journalist}}
  \end{subfigure}

  \begin{subfigure}[t]{0.95\linewidth}
    \centering
    \includegraphics[width=\linewidth]{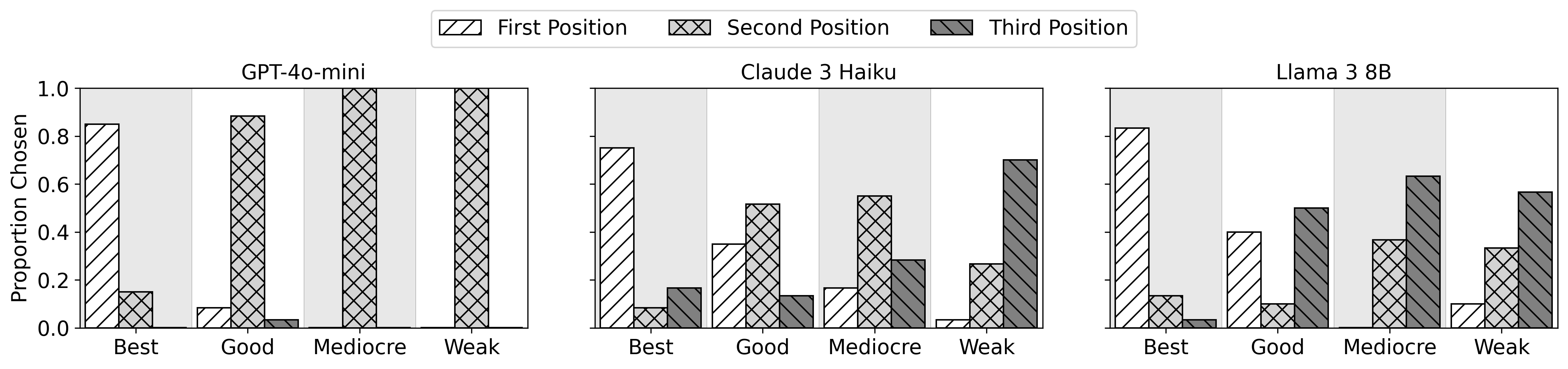}
    \caption*{{\footnotesize \textbf{(d)} Registered Nurse}}
  \end{subfigure}

  \begin{subfigure}[t]{0.95\linewidth}
    \centering
    \includegraphics[width=\linewidth]{figures/triplewise_position_bias_All_temp_0.png}
    \caption*{{\footnotesize \textbf{(e)} Aggregated across all four professions}}
  \end{subfigure}

  \caption{{\bf Position Bias of triplewise resume comparisons, \boldmath \(T=0\)}. Positional order effects in triplewise comparisons across four professions (Mechanical Engineer, Real Estate Agent, Journalist and Registered Nurse, top to bottom) and aggregated across professions.}
  \label{fig:triplewise_temp_0_all}
\end{figure}

\begin{figure}[ht]
  \centering
  \begin{subfigure}[t]{0.95\linewidth}
    \centering
    \includegraphics[width=\linewidth]{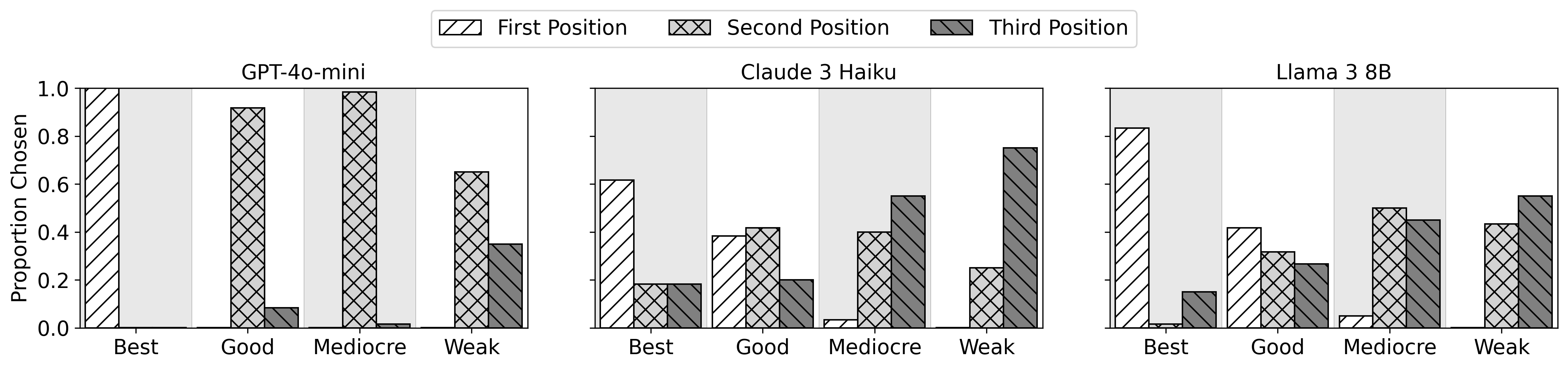}
    \caption*{{\footnotesize \textbf{(a)} Mechanical Engineer}}
  \end{subfigure}

  \begin{subfigure}[t]{0.95\linewidth}
    \centering
    \includegraphics[width=\linewidth]{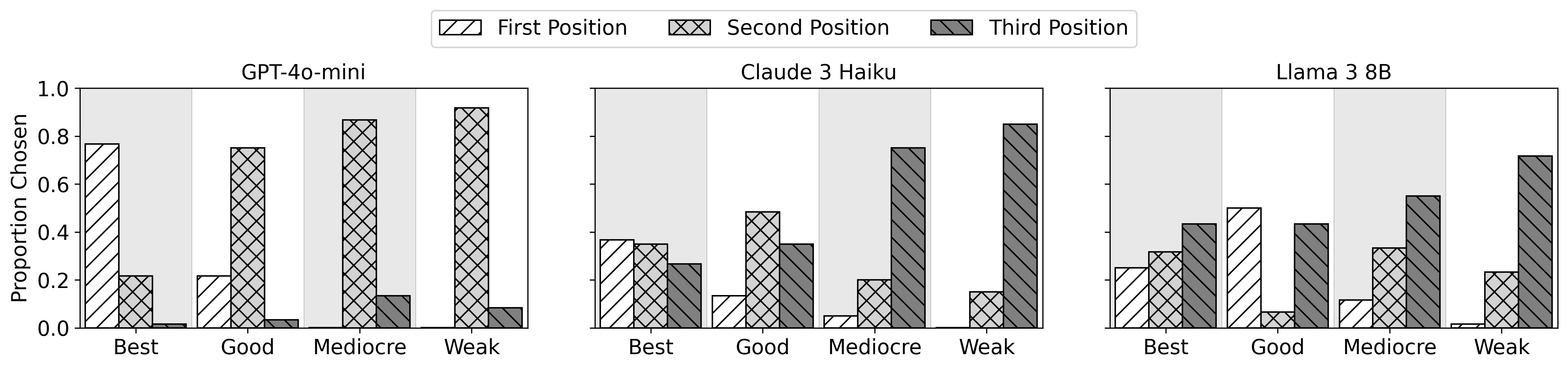}
    \caption*{{\footnotesize \textbf{(b)} Real Estate Agent}}
  \end{subfigure}

  \begin{subfigure}[t]{0.95\linewidth}
    \centering
    \includegraphics[width=\linewidth]{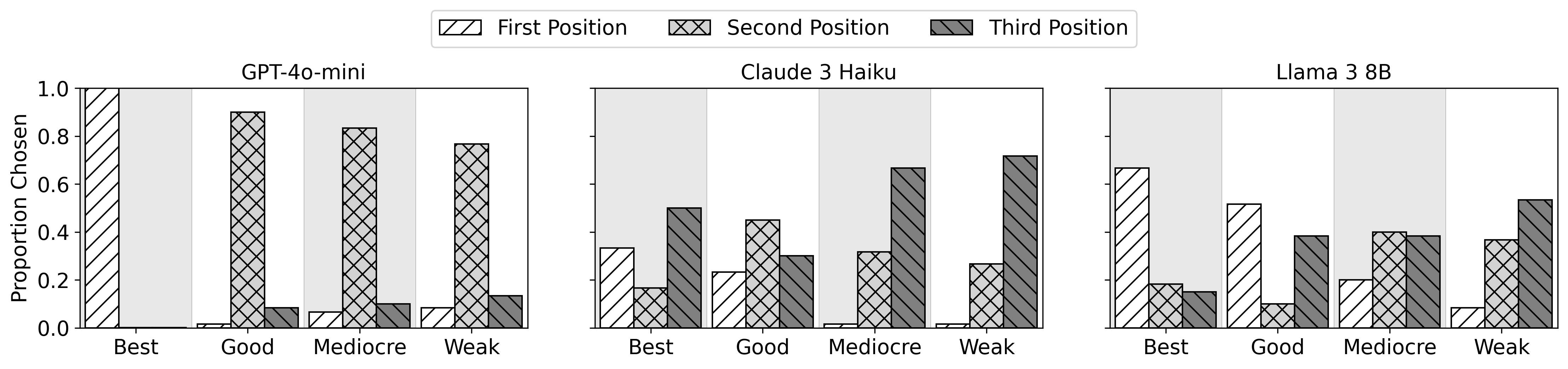}
    \caption*{{\footnotesize \textbf{(c)} Journalist}}
  \end{subfigure}

  \begin{subfigure}[t]{0.95\linewidth}
    \centering
    \includegraphics[width=\linewidth]{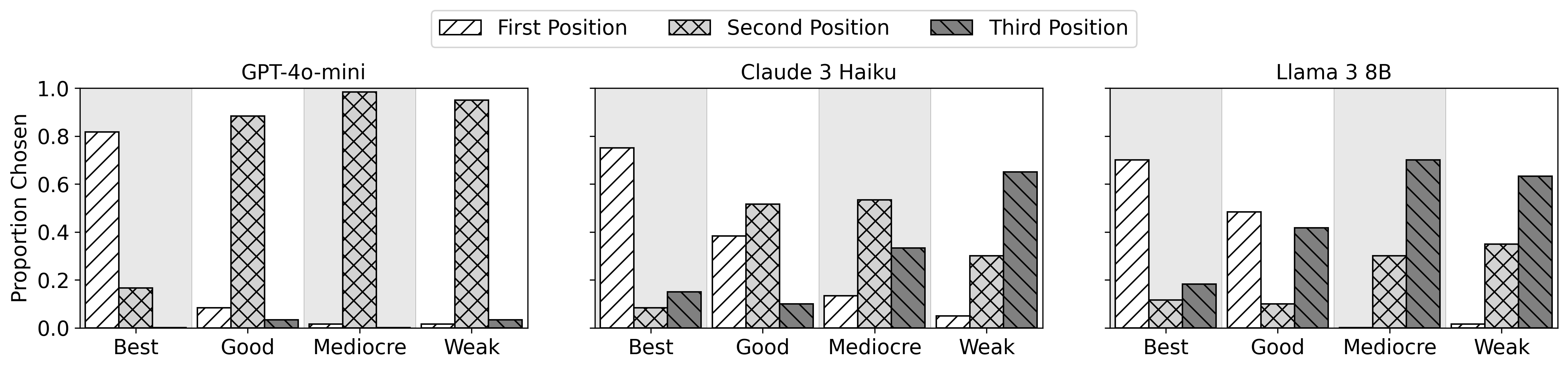}
    \caption*{{\footnotesize \textbf{(d)} Registered Nurse}}
  \end{subfigure}

  \begin{subfigure}[t]{0.95\linewidth}
    \centering
    \includegraphics[width=\linewidth]{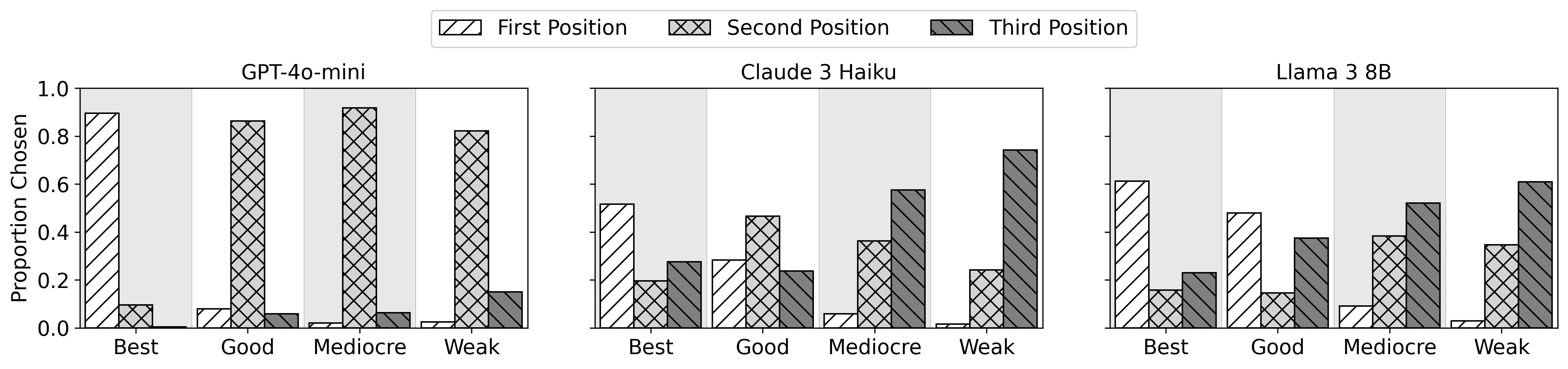}
    \caption*{{\footnotesize \textbf{(e)} Aggregated across all four professions}}
  \end{subfigure}

  \caption{{\bf Triplewise comparisons, \boldmath $T=1$}. Positional order effects in triplewise comparisons across four professions (Mechanical Engineer, Real Estate Agent, Journalist and Registered Nurse, top to bottom) and aggregated totals.}
  \label{fig:triplewise_temp_1_all}
\end{figure}

\autoref{fig:triplewise_temp_0_all} and~\autoref{fig:triplewise_temp_1_all} visualize the position effect results for the four professions, as well as aggregated results, at temperatures 0 and 1 respectively.  Triplewise comparisons for additional models are reported in Appendix~\ref{models:trip} and Appendix~\ref{lang:trip} respectively.

\medskip 

\begin{table}[ht]
\centering
\begin{tabular}{llcccc}
\textbf{Model} & \textbf{Tier} & \textbf{First / Second / Third (\%)} & \textbf{\boldmath$\chi^2$} & \textbf{Cram\'{e}r's \boldmath$V$} & \textbf{\boldmath$p$-value} \\
\midrule
\multirow{4}{*}{GPT-4o-mini}
  & {\Bestcolor}   & 90 / 9 / 1   & 90.62 & 0.86 & ${\bf 4.0\times10^{-13}}^{***}$ \\
  & {\Goodcolor}   & 8 / 86 / 6   & 77.85 & 0.80 & ${\bf 3.6\times10^{-12}}^{***}$ \\
  & {\Medcolor}    & 2 / 92 / 6   & 93.55 & 0.88 & ${\bf 5.3\times10^{-16}}^{***}$ \\
  & {\Worstcolor}  & 3 / 82 / 15  & 72.05 & 0.76 & ${\bf 1.3\times10^{-9}}^{***}$  \\
\midrule
\multirow{4}{*}{Claude 3 Haiku}
  & {\Bestcolor}   & 52 / 20 / 28 & 19.98 & 0.36 & 0.180                              \\
  & {\Goodcolor}   & 29 / 47 / 24 & 9.38  & 0.27 & 0.058                             \\
  & {\Medcolor}    & 6 / 36 / 58  & 32.05 & 0.51 & ${\bf 1.9\times10^{-4}}^{***}$   \\
  & {\Worstcolor}  & 2 / 24 / 74  & 51.15 & 0.65 & ${\bf 2.0\times10^{-8}}^{***}$   \\
\midrule
\multirow{4}{*}{Llama 3 8B}
  & {\Bestcolor}   & 61 / 16 / 23 & 34.75 & 0.49 & 0.053                             \\
  & {\Goodcolor}   & 48 / 15 / 37 & 13.28 & 0.32 & 0.088                             \\
  & {\Medcolor}    & 9 / 38 / 52  & 21.93 & 0.40 & ${\bf 0.026}^{*}$                \\
  & {\Worstcolor}  & 3 / 35 / 62  & 33.48 & 0.52 & ${\bf 1.2\times10^{-5}}^{***}$   \\
\bottomrule
\end{tabular}
\betweentableandcap

\caption{\textbf{Order bias in triplewise resume comparisons, \boldmath\(T=1\).} For each model and quality tier, “First/Second/Third (\%)” shows the percentage of times the first, second, or third option was chosen. Position bias was evaluated with a chi-square goodness-of-fit test ($\chi^2$) against uniform choice; \text{Cram\'{e}r's }  $V$ is the effect size. Asterisks denote $^{*}p<0.05$, $^{**}p<0.01$, $^{***}p<0.001$.}
\label{tab:triplewise_resumes_temp1}
\end{table}

\clearpage

%% file: _z_appendix_four_colors.tex
\section{Fourwise Color Comparisons}
\label{app:four_colors}

For fourwise comparisons, we manually curated a set of four colors, chosen to be roughly comparable, for each LLM. For each model, we presented all 
$4! = 24$  permutations of these colors. We repeated each permutation four times at temperature~0, yielding 96 total queries per model.

The colors selected were:
\begin{itemize}
\item \textbf{GPT-4o-mini:} Aqua Mist, Robin’s Egg Blue, Soft Sky Blue, Pale Turquoise,
\item \textbf{Claude 3 Haiku:} Pale Periwinkle, Pistachio, Blush Pink, Soft Terracotta,
\item \textbf{Llama 3 8B:} Emerald Green, Taupe, Navy Blue, Beige.
\end{itemize}

For GPT-4o-mini, the color set included its three colors from the \emph{Ideal} tier, with Robin’s Egg Blue added. For Claude 3 Haiku, the colors came from its \emph{Fair} tier, with Pistachio added. For Llama 3 8B, two colors were drawn from its \emph{Plain} tier, with Emerald Green and Navy Blue added. Selection frequencies among the different colors were less balanced than in pairwise and triplewise comparisons, but each color was chosen at least once per repetition. GPT-4o-mini selected Aqua Mist and Robin’s Egg Blue 30 times each, Soft Sky Blue 24 times, and Pale Turquoise 12 times. Claude 3 Haiku most often selected Pale Periwinkle (42), followed by Pistachio (26), Blush Pink (24), and Soft Terracotta (4). Llama 3 8B favored Emerald Green (40), then Taupe (28), Navy Blue (20), and Beige (8).

\begin{figure}[ht]
    \centering
    \includegraphics[width=0.75\linewidth]{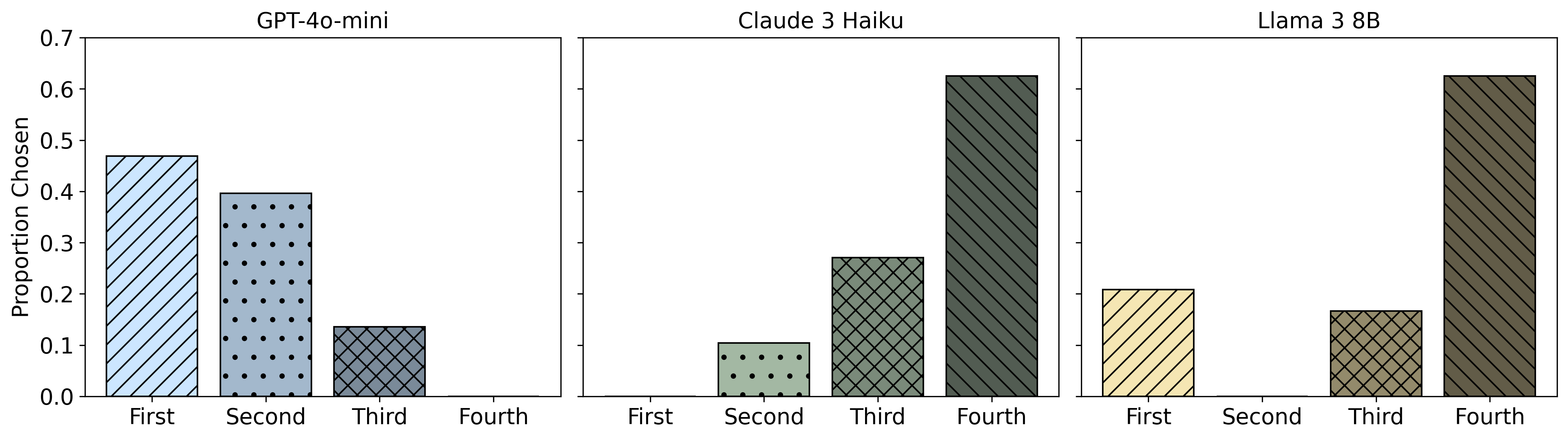}
    \caption{
    {\bf Fourwise position bias across models, \boldmath $T=0$.} Proportion of times each position was chosen in four-option comparisons. Bar colors progressively darken to indicate position rank. All models show substantial positional preferences, suggesting robust order-driven distortions in higher-choice settings.
    }
    \label{fig:fourwise_position_bias}
\end{figure}

Figure~\ref{fig:fourwise_position_bias} shows the proportion of times each position (first, second, third, fourth) was chosen across all permutations, indicating consistent positional preferences in fourwise comparisons across all three models.

%% file: _z_appendix_name.tex
\section{Name Bias}
\label{app:name}

\subsection{Name Bias in Triplewise Comparisons}

To evaluate whether name selections were uniformly distributed across the 30 candidate names for each model, we performed a chi-squared goodness-of-fit test. The results are shown in~\autoref{tab:chi_squared_results}: GPT-4o-mini's selections closely match a uniform distribution ($\chi^2 = 9.6$, $p = 0.9997$), while Claude 3 Haiku shows a strong deviation ($\chi^2 = 215.8$, $p =1.9 \times 10^{-30}$), indicating a strong name-level selection bias. Llama falls between these extremes ($\chi^2 = 73.1$, $p = 1.1 \times 10^{-5}$).

\begin{figure}[ht]
\centering
\includegraphics[width=\textwidth]{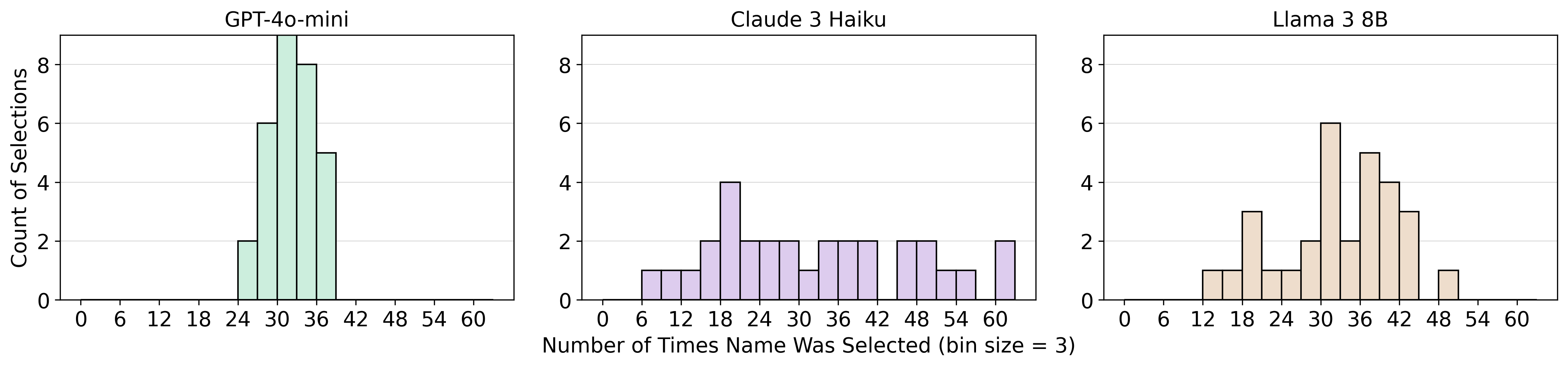}
\caption{{\bf Histogram of name selections per model}, binned every 3. The histogram shows how often each of the 30 candidate names was chosen across all triplewise comparisons.  }
\label{fig:histograms}
\end{figure}

\begin{figure}[ht]
\centering
\includegraphics[width=0.5\textwidth]{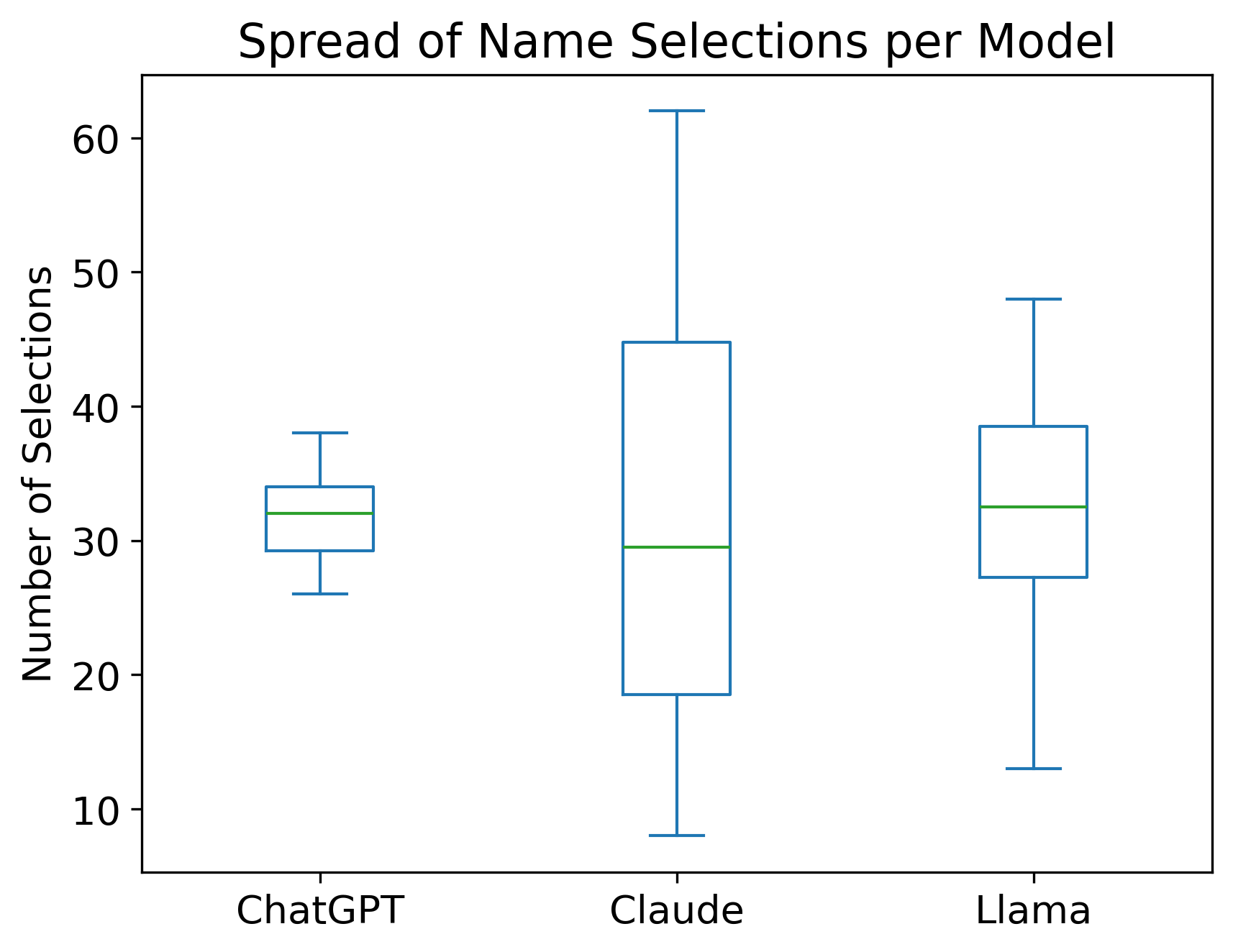}
\caption{{\bf Box plot of the times each name was selected per model.}}
\label{fig:boxplots}
\end{figure}

\autoref{fig:histograms} visualizes the full distribution of name selection frequencies for each model using histograms with bin size 3. The expected number of times each name would be chosen if there was no name bias is 32. GPT-4o-mini chose each name between 26 and 38 times, consistent with little to no bias. Claude 3 Haiku 's selections show high variance, with a wide spread and strong preference for certain names. 
\autoref{fig:boxplots} shows box plots for each model, confirming these patterns: Claude 3 Haiku  has the widest interquartile range and largest spread, while GPT-4o-mini's selection counts are narrowly distributed with minimal outliers. \autoref{tab:claude_preference_names} shows Claude 3 Haiku's the top 5 least and most `favorite' names: it chose some names as few as 8 times, and others as much as 62 times.

\begin{table}[ht]
\centering
\begin{tabular}{lllcc}
\centering
\textbf{Model}   & \textbf{Gender} & \boldmath$\chi^2$ & \textbf{\boldmath$p$-value} & \boldmath$n$ \\
\midrule
GPT-4o-mini  & Male     & 4.50     & $0.99$    & 480 \\
        & Female     & 5.12     & $0.98$    & 480 \\
        & All   & 9.62     & $1.00$    & 960 \\
\midrule
Claude 3 Haiku  &  Male    & 100.9   & $\bf{3.2 \times 10^{-15}}^{***}$ & 480 \\
        & Female   & 114.9   & ${\bf 6.3 \times 10^{-18}}^{***}$ & 480 \\
        & All   & 215.8   & $\bf{1.9 \times 10^{-30}}^{***}$ & 960 \\
\midrule
Llama  3 8B  & Male     & 19.62    & $0.142$    & 480 \\
        & Female     & 53.44    & $\bf{1.6 \times 10^{-6}}^{***}$  & 480 \\
        & All   & 73.06    & $\bf{1.1 \times 10^{-5}}^{***}$  & 960 \\
\bottomrule
\end{tabular}
\betweentableandcap

\caption{{\bf Name bias in triplewise comparisons,  \boldmath\(T=1\).} For each model, we report separate goodness-of-fit $\chi^2$ statistics and p-values for male names, female names, and combined (All), along with the total number of comparisons (n). Asterisks indicate statistical significance at $^{***} p < .001$.}
\label{tab:chi_squared_results}
\end{table}

\begin{table}[ht]
\centering
\begin{tabular}{lcc}
\textbf{Name} & \textbf{Gender} & \textbf{Number of Times Chosen} \\
\midrule
Christopher Taylor  & male   & 62 \\
Megan Anderson      & female & 61 \\
Stephanie Clark     & female & 56 \\
Joshua Clark        & male   & 52 \\
Jessica Johnson     & female & 49 \\
... & ... & ... \\
Lauren Wilson       & female & 17 \\
Hannah Miller       & female & 15 \\
Michael Brown       & male   & 13 \\
Andrew Harris       & male   & 9  \\
Nicole Harris       & female & 8  \\
\bottomrule
\end{tabular}
\betweentableandcap

\caption{{\bf Claude 3 Haiku's favorite and least favorite names.}}
\label{tab:claude_preference_names}
\end{table}

 \clearpage

\subsection{Name Bias in Pairwise Comparisons}

As noted in Appendix~\ref{app:resume_generation}, the names used for pairwise comparisons were \emph{Brandon Lewis, Charles Wilson, Christopher Taylor} and \emph{Andrew Harris} (male) and \emph{Elizabeth Jones, Ashley Williams, Amanda Thomas} and \emph{Emily Smith} (female). These names were selected at random, before we had analyzed the name bias in triplewise comparisons. That is, we did not yet know that two of the male names were the least and most favorite of Claude 3 Haiku. The number of samples from the pairwise comparisons that we had previously carried out (Appendix~\ref{app:pairwise_resumes}),  were insufficient for statistical significance (except for Claude 3 Haiku). We therefore extended the comparisons to all three resumes used in triplewise comparisons, but limited our analysis to results where neither resume dominated the other. The results are shown in \autoref{tab:name_bias_chi_squared_by_gender}. As the description of this experiment was somewhat cumbersome to describe in the main text, we also analyzed the direct comparisons between Christopher Taylor and Andrew Harris (from the original pairwise comparison results). Claude 3 Haiku selected Christopher Taylor  in 82 out of 128 trials ($p = 0.0019$, $h = 0.29$).

\begin{table}[ht]
\centering
\begin{tabular}{llcccc}
\textbf{Model} & \textbf{Gender} & \textbf{\boldmath$\chi^2$} & \textbf{\boldmath$p$-value} & \textbf{Favorite} & \textbf{Least Favorite} \\
\midrule
\multirow{2}{*}{GPT-4o-mini} 
  & Male   & 8.21 & ${\bf 0.042}^{*}$ & Brandon Lewis & Andrew Harris \\
  & Female & 1.49 & $0.684$              & –             & – \\
\multirow{2}{*}{Claude 3 Haiku}
  & Male   & 75.93 & ${\bf 2.3 \times 10^{-16}}^{***}$ & Christopher Taylor & Andrew Harris \\
  & Female & 17.03 & ${\bf 7.0 \times 10^{-4}}^{***}$ & Elizabeth Jones & Amanda Thomas \\
\multirow{2}{*}{Llama 3 8B}
  & Male   & 11.76 & ${\bf 0.008}^{**}$ & Christopher Taylor & Charles Wilson \\
  & Female & 3.77  & $0.287$              & –             & – \\
\bottomrule
\end{tabular}
\betweentableandcap

\caption{{\bf Name bias in pairwise comparisons, \boldmath \(T=1\).} Chi-squared tests for uniform name selection distribution, compared to a baseline of 25\%, by gender. Asterisks denote significance levels: $^{*}p<0.05, ^{**}p<.01, ^{***}p<.001$.}
\label{tab:name_bias_chi_squared_by_gender}
\end{table}

%% file: _z_appendix_gender.tex
\section{Gender Bias}
\label{app:gender}

We conducted cross‐gender pairwise comparisons using the same set of resumes and name–profile pairings employed in our same‐gender experiments. For each tier and profession, we compared every male–female pairing in both presentation orders, yielding 64 total comparisons per tier per profession (see \ifpnas Methods and Materials\else Section~\ref{sec:materials}\fi). \autoref{tab:pairwise_gender_temp1_position} reports the position bias observed in these cross‐gender trials, while \autoref{tab:pairwise_gender_temp1_gender} reports the gender bias. Across nearly all models and tiers, positional effects exceed gender effects, with the sole exception of the \emph{Best} tier under Claude 3 Haiku.

To test whether this difference in effect sizes is itself statistically significant, we first defined, for each model–tier combination, the \emph{dominant position} as the position (first or second) chosen most frequently in the  same‐gender comparisons (Appendix~\ref{app:pairwise_resumes}). We then aggregated the following over the cross‐gender data: (1) the number of times the resume in that dominant position was selected, and (2) the number of times the female candidate was selected. The resulting positional and gender effect sizes are reported in \autoref{tab:bootstrapping}. Finally, we performed a nonparametric bootstrap (10,000 resamples with replacement) on these aggregated counts to generate 95\% confidence intervals for the difference in effect sizes; in every model–tier case the interval excluded zero, confirming that order effects are stronger than gender effects.

\begin{figure}[ht]
    \centering
    \includegraphics[width=\textwidth]{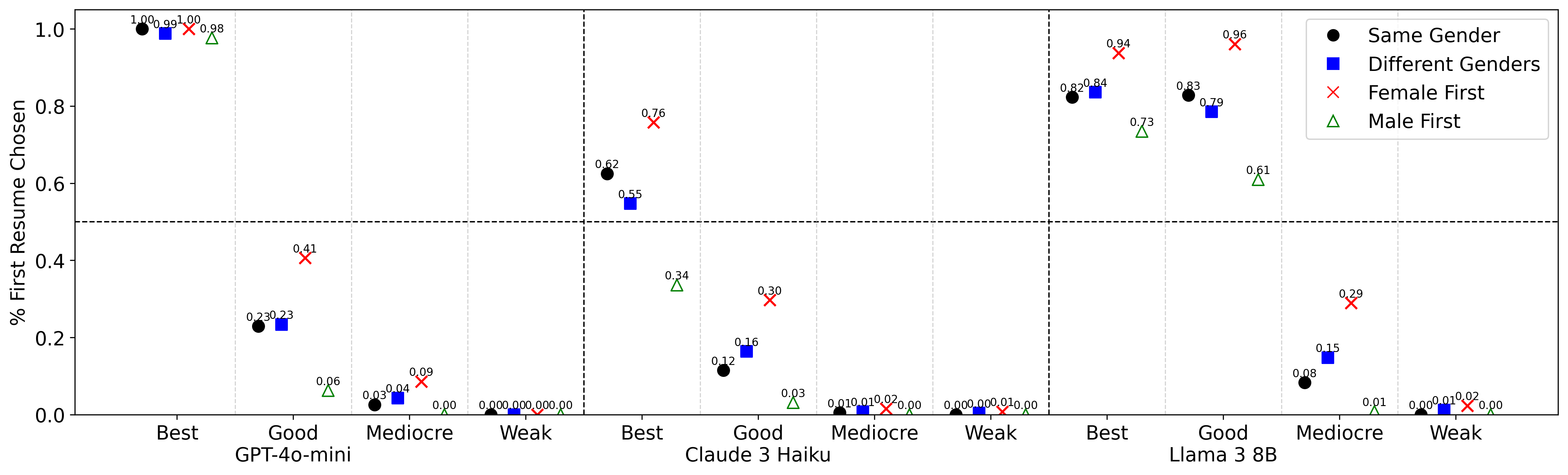}  
    \caption{{\bf Same-gender vs cross-gender comparisons}. Four results are displayed for comparison: 1) Order bias within same-gender comparisons, 2) Aggregate Order bias in cross-gender comparisons, 3) Order bias when the female candidate was the first position, and 4) Order bias when the male candidate was in the first option. In all cases, the plots report the proportion of times the first candidate presented was chosen.}
    \label{fig:pointy}
\end{figure}

 The reader may wish to compare~\autoref{tab:pairwise_gender_temp1_position}  with~\autoref{tab:pairwise-all_resumes_temp_1}, which gives the corresponding results for same-gender comparisons. On aggregate, GPT-4o-mini, Claude 3 Haiku and Llama 3 8B chose the candidate in the dominant position 93.6\%, 87.6\% and 89.2\% of the time respectively, in same-gender comparisons, compared to 92.8\%, 84.3\% and 86.5\%  in cross-gender comparisons. We note that the name bias may play a role in these results; we do not attempt to disentangle the name and gender biases here. \autoref{tab:gender_diff_summary} breaks down the results further, reporting the proportion of times the first position was chosen when the female/male candidate was presented first. Across all models and tiers, the first position was selected more often when the female candidate was presented first (except for one case, when the candidate in the first position was never selected). ~\autoref{fig:pointy} visually compares the results in ~\autoref{tab:pairwise_gender_temp1_position},~\autoref{tab:pairwise-all_resumes_temp_1} and~\autoref{tab:gender_diff_summary}, by plotting the proportion of times the first resume was presented in 1) same-gender comparisons, 2) cross-gender comparisons, 3) cross-gender comparisons when the female candidate was presented first, and 4) cross-gender comparisons when the male candidate was presented first.

\begin{figure}[h]
  \centering
  \begin{subfigure}[t]{0.95\linewidth}
    \centering
    \includegraphics[width=\linewidth]{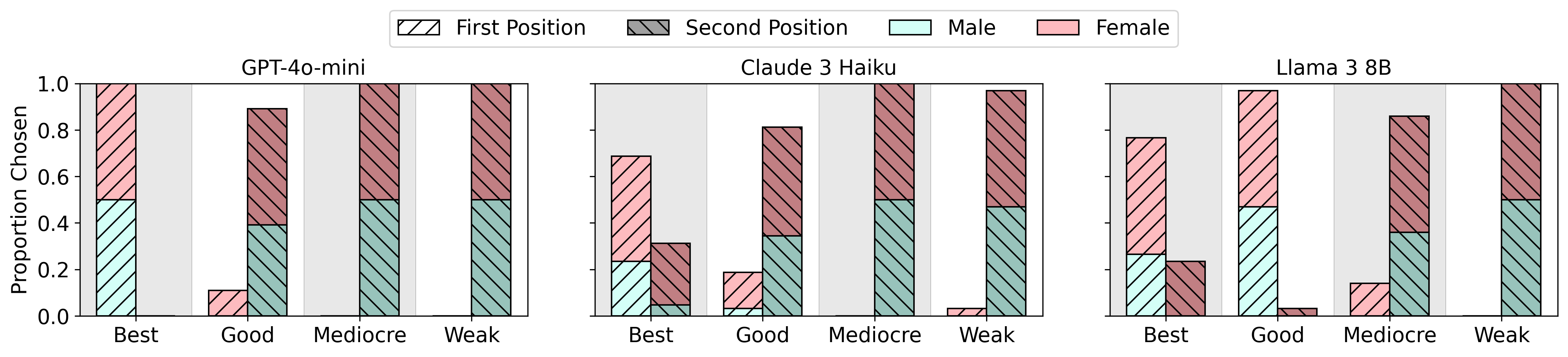}
    \caption*{{\footnotesize \textbf{(a)} Mechanical Engineer}}
  \end{subfigure}

  \begin{subfigure}[t]{0.95\linewidth}
    \centering
    \includegraphics[width=\linewidth]{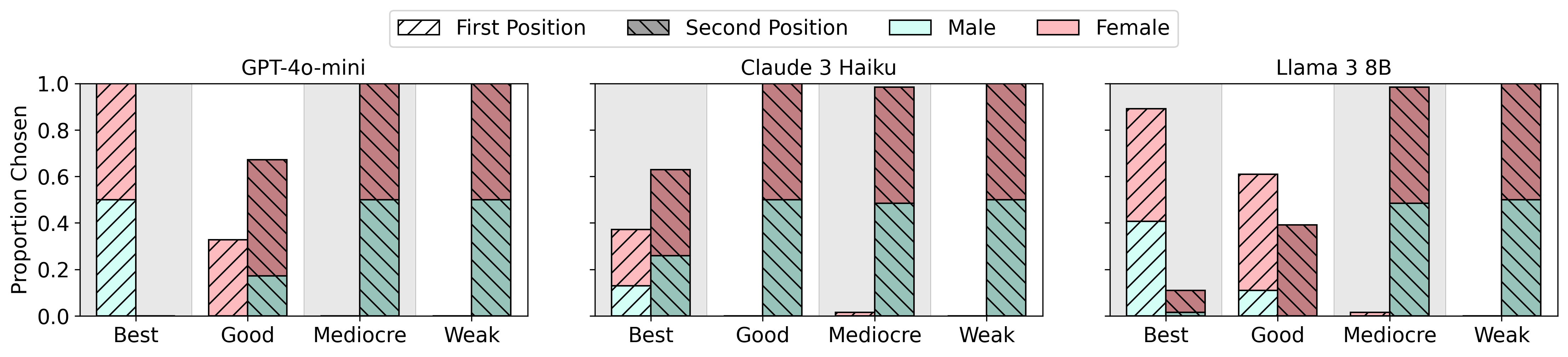}
    \caption*{{\footnotesize \textbf{(b)} Real Estate Agent}}
  \end{subfigure}

  \begin{subfigure}[t]{0.95\linewidth}
    \centering
    \includegraphics[width=\linewidth]{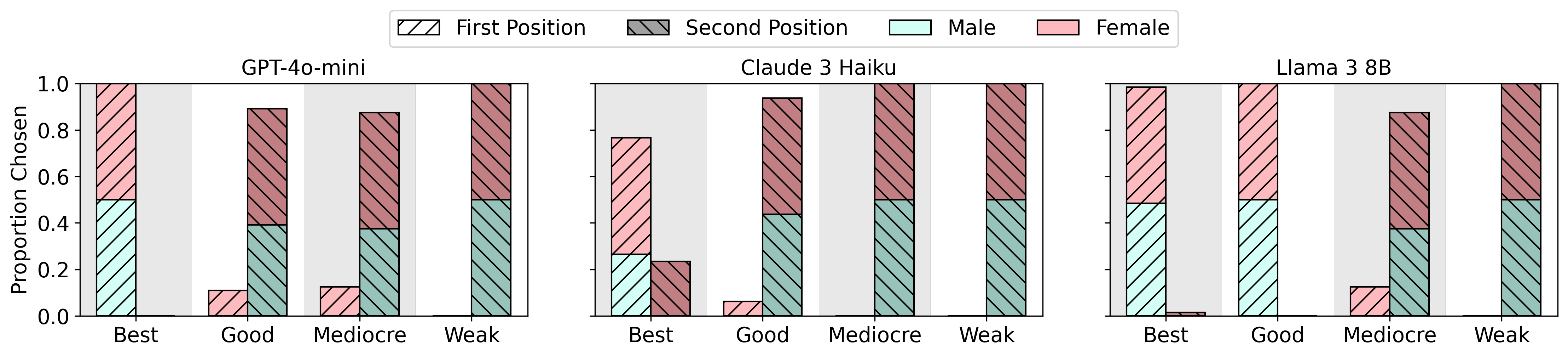}
    \caption*{{\footnotesize \textbf{(c)} Journalist}}
  \end{subfigure}

  \begin{subfigure}[t]{0.95\linewidth}
    \centering
    \includegraphics[width=\linewidth]{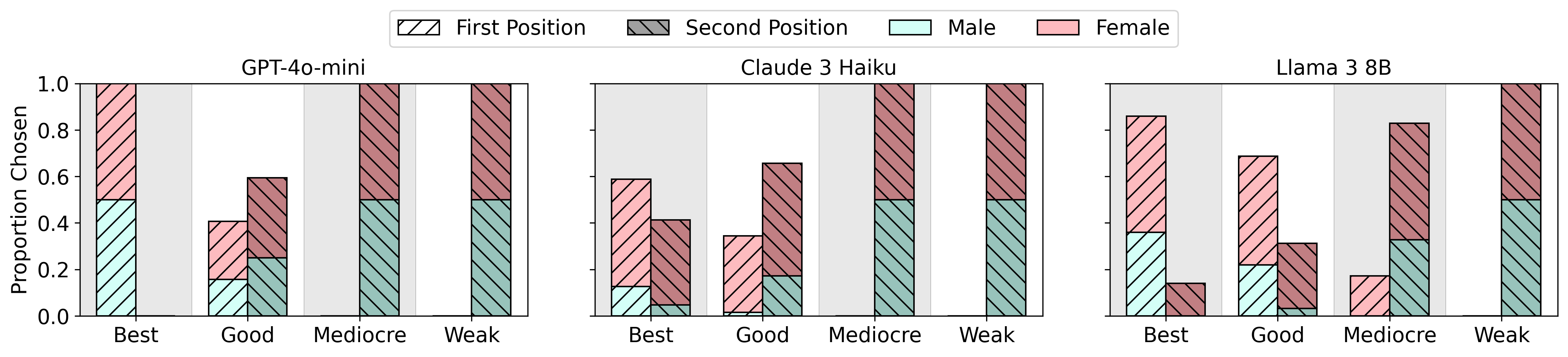}
    \caption*{{\footnotesize \textbf{(d)} Registered Nurse}}
  \end{subfigure}

  \begin{subfigure}[t]{0.95\linewidth}
    \centering
    \includegraphics[width=\linewidth]{figures/gender_All_StackedBias_temp_0.png}
    \caption*{{\footnotesize \textbf{(e)} Aggregated across all four professions}}
  \end{subfigure}

  \caption{{\bf Gender and position bias in pairwise comparisons, \boldmath\(T=0\)}. Gender effects in pairwise comparisons across four professions (Mechanical Engineer, Real Estate Agent, Journalist and Registered Nurse, top to bottom) and aggregated. Each panel shows the proportion of times each position was chosen, broken down by LLM and quality tier.}
  \label{fig:gender_temp_0_all}
\end{figure}

\begin{figure}[h]
  \centering
  \begin{subfigure}[t]{0.95\linewidth}
    \centering
    \includegraphics[width=\linewidth]{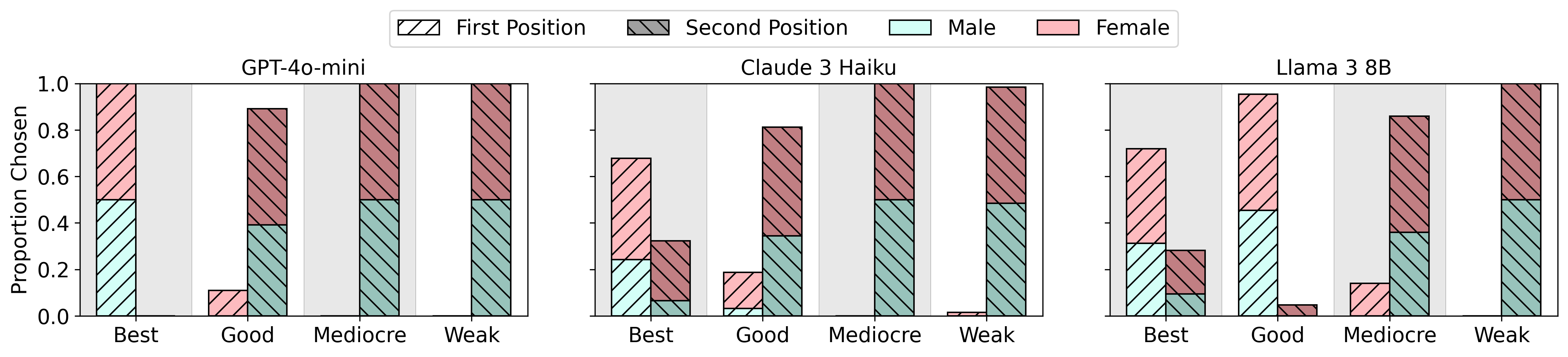}
    \caption*{{\footnotesize \textbf{(a)} Mechanical Engineer}}
  \end{subfigure}

  \begin{subfigure}[t]{0.95\linewidth}
    \centering
    \includegraphics[width=\linewidth]{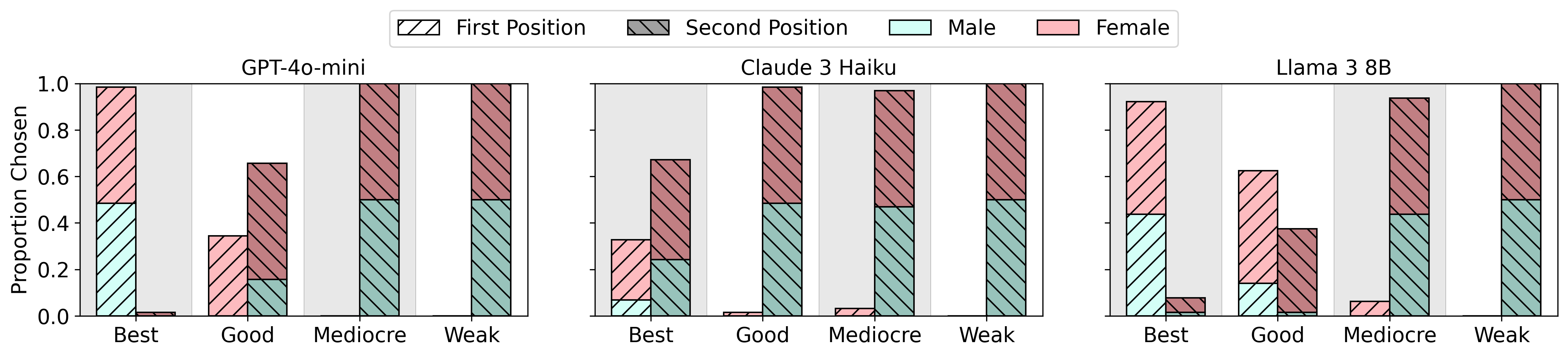}
    \caption*{{\footnotesize \textbf{(b)} Real Estate Agent}}
  \end{subfigure}

  \begin{subfigure}[t]{0.95\linewidth}
    \centering
    \includegraphics[width=\linewidth]{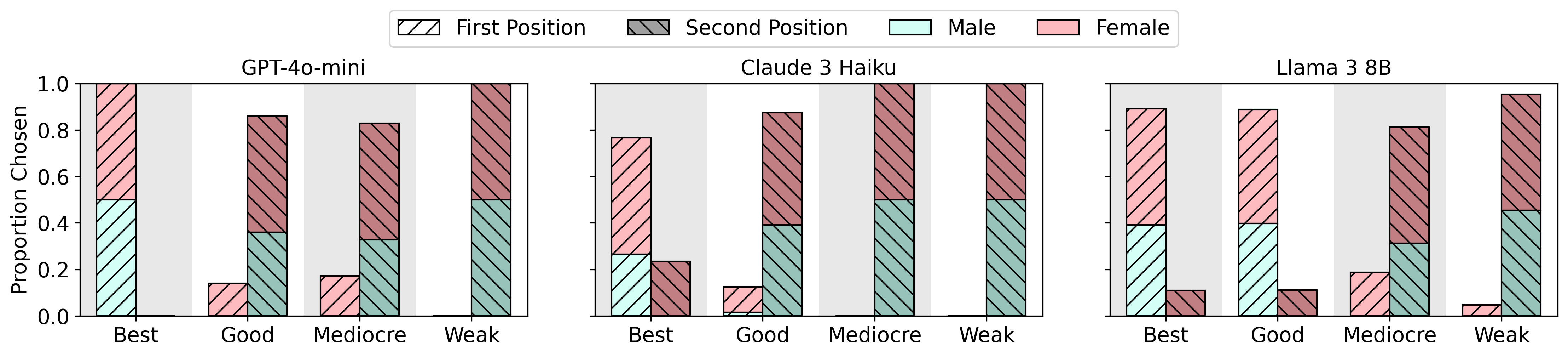}
    \caption*{{\footnotesize \textbf{(c)} Journalist}}
  \end{subfigure}

  \begin{subfigure}[t]{0.95\linewidth}
    \centering
    \includegraphics[width=\linewidth]{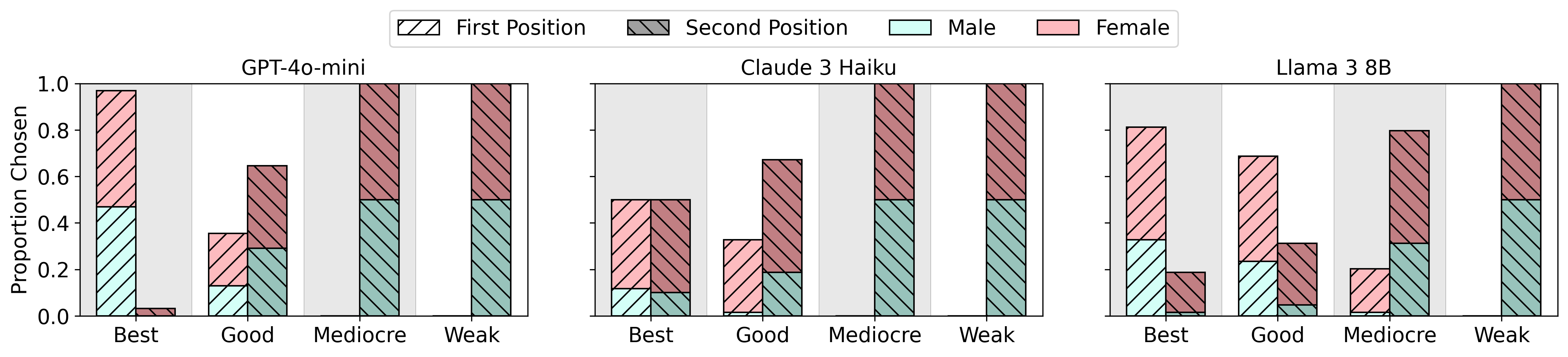}
    \caption*{{\footnotesize \textbf{(d)} Registered Nurse}}
  \end{subfigure}

  \begin{subfigure}[t]{0.95\linewidth}
    \centering
    \includegraphics[width=\linewidth]{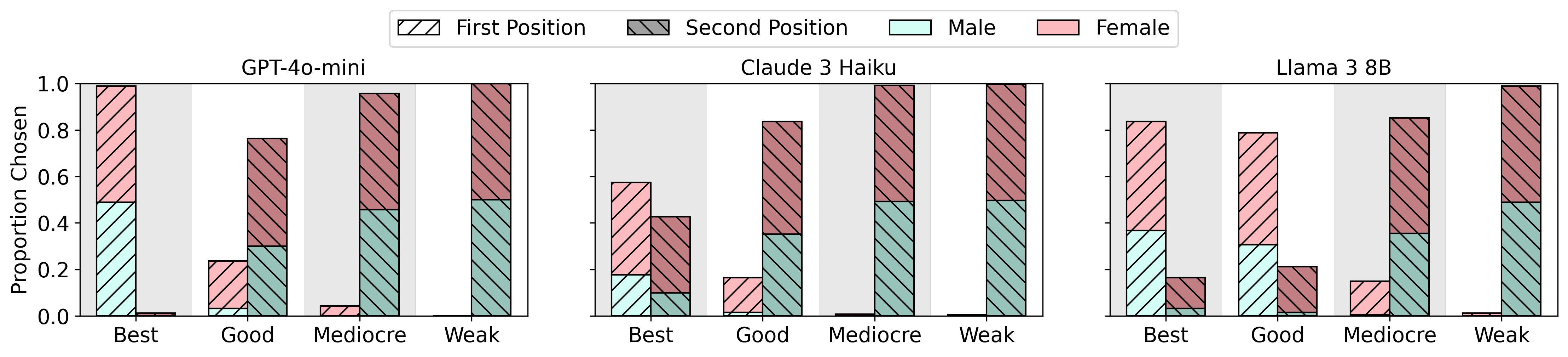}
    \caption*{{\footnotesize \textbf{(e)} Aggregated across all four professions}}
  \end{subfigure}

  \caption{ {\bf Gender and position bias in pairwise comparisons, \boldmath \(T=1\)}. Gender effects in pairwise comparisons across four professions (Mechanical Engineer, Real Estate Agent, Journalist and Registered Nurse, top to bottom) and aggregated. Each panel shows the proportion of times each position was chosen, broken down by LLM and quality tier.}
  \label{fig:gender_temp_1_all}
\end{figure}

\autoref{fig:gender_temp_0_all} and  \autoref{fig:gender_temp_1_all}
show the proportion of times each position was chosen, broken down by gender, at \(T=0\) and \(T=1\) respectively. 

\medskip

\begin{table}[ht]
\centering
\begin{tabular}{llccc}

\textbf{Model} & \textbf{Tier} & \textbf{First / Second (\%)} & \textbf{Cohen’s \boldmath$h$} & \textbf{\boldmath$p$-value} \\
\midrule
\multirow{4}{*}{GPT-4o-mini}
  & Best      & 99 / 1   & 1.35 & ${\bf 4.8 \times 10^{-71}}^{***}$ \\
  & Good      & 23 / 77  & 0.56 & ${\bf 5.0 \times 10^{-18}}^{***}$ \\
  & Mediocre  & 4 / 96   & 1.15 & ${\bf 1.1 \times 10^{-58}}^{***}$ \\
  & Weak      & 0 / 100  & 1.57 & ${\bf 1.7 \times 10^{-77}}^{***}$ \\
\midrule
\multirow{4}{*}{Claude 3 Haiku}
  & Best      & 55 / 45  & 0.09 & $0.150$ \\
  & Good      & 16 / 84  & 0.74 & ${\bf 6.1 \times 10^{-29}}^{***}$ \\
  & Mediocre  & 1 / 99   & 1.39 & ${\bf 5.7 \times 10^{-73}}^{***}$ \\
  & Weak      & 0 / 100   & 1.57 & ${\bf 4.4 \times 10^{-75}}^{***}$ \\
\midrule
\multirow{4}{*}{Llama 3 8B}
  & Best      & 84 / 16  & 0.74 & ${\bf 6.1 \times 10^{-29}}^{***}$ \\
  & Good      & 79 / 21  & 0.61 & ${\bf 1.0 \times 10^{-20}}^{***}$ \\
  & Mediocre  & 15 / 85  & 0.78 & ${\bf 7.2 \times 10^{-32}}^{***}$ \\
  & Weak      & 1 / 99   & 1.39 & ${\bf 4.8 \times 10^{-71}}^{***}$ \\
\bottomrule
\end{tabular}
\betweentableandcap

\caption{\textbf{Pairwise position bias in cross-gender resume comparisons, \boldmath\(T=1\).} Results are aggregated across all professions, grouped by model and tier. For each condition, we report the proportion of times the first vs.\ second resume was chosen (First/Second \%), the corresponding Cohen’s $h$ effect size, and the $p$-value from a two-sided binomial test against a 50\% null. All p-values are from chi-square tests of uniform choice; ${\bf p<0.001}^{***}$ indicates statistical significance.}
\label{tab:pairwise_gender_temp1_position}
\end{table}

\begin{table}[ht]
\centering
\begin{tabular}{llccc}

\textbf{Model} & \textbf{Tier} & \textbf{Female / Male (\%)} & \textbf{Cohen’s \boldmath$h$} & \textbf{\boldmath$p$-value} \\
\midrule
\multirow{4}{*}{GPT-4o-mini}
  & Best      & 51 / 49  & 0.02 & $0.755$                         \\
  & Good      & 67 / 33  & 0.35 & ${\bf 4.0 \times 10^{-8}}^{***}$ \\
  & Mediocre  & 54 / 46  & 0.09 & $0.189$                         \\
  & Weak      & 50 / 50  & 0.00 & $1.000$                          \\
\midrule
\multirow{4}{*}{Claude 3 Haiku}
  & Best      & 71 / 29  & 0.45 & ${\bf 1.1 \times 10^{-11}}^{***}$ \\
  & Good      & 63 / 37  & 0.27 & ${\bf 2.6 \times 10^{-5}}^{***}$  \\
  & Mediocre  & 51 / 49  & 0.02 & $0.851$                         \\
  & Weak      & 50 / 50  & 0.00 & $0.950$                         \\
\midrule
\multirow{4}{*}{Llama 3 8B}
  & Best      & 60 / 40  & 0.21 & ${\bf 1.4 \times 10^{-3}}^{**}$  \\
  & Good      & 68 / 32  & 0.36 & ${\bf 1.9 \times 10^{-8}}^{***}$ \\
  & Mediocre  & 64 / 36  & 0.29 & ${\bf 8.0 \times 10^{-6}}^{***}$ \\
  & Weak      & 51 / 49  & 0.02 & $0.755$                         \\
\bottomrule
\end{tabular}
\betweentableandcap
\caption{{\bf Gender bias at temperature 1}. The two candidates differ in gender and alternate positions in the pairwise comparisons; values show the percentage of times the female vs.\ male candidate was selected, reported as Female/Male (\%). Statistical significance is from binomial tests; effect sizes are Cohen’s $h$. Asterisks denote $^{*}p<.05$, $^{**}p<.01$, $^{***}p<.001$.}
\label{tab:pairwise_gender_temp1_gender}
\end{table}

\ignore{
\begin{table}[ht]
\centering
\begin{tabular}{lcccc}
\textbf{Model} & \textbf{Bias Type} & \textbf{\% Dominant Choice} & \textbf{Cohen's \boldmath$h$} & \textbf{\boldmath$p$-value} \\
\midrule
\multirow{2}{*}{GPT-4o-mini}
 & Order Bias  & 92.8 & 1.03 & ${\bf 1.4 \times 10^{-194}}^{***}$ \\
 & Gender Bias & 55.7 & 0.11 & ${\bf 3.2 \times 10^{-4}}^{***}$ \\
\midrule
\multirow{2}{*}{Claude 3 Haiku}
 & Order Bias  & 84.3 & 0.76 & ${\bf 1.4 \times 10^{-116}}^{***}$ \\
 & Gender Bias & 58.9 & 0.18 & ${\bf 1.4 \times 10^{-8}}^{***}$ \\
\midrule
\multirow{2}{*}{Llama 3 8B}
 & Order Bias  & 72.3 & 0.46 & ${\bf 1.9 \times 10^{-47}}^{***}$ \\
 & Gender Bias & 60.7 & 0.22 & ${\bf 6.4 \times 10^{-12}}^{***}$ \\
\bottomrule
\end{tabular}
\betweentableandcap 
\caption{
{\bf Aggregated positional and gender biases}. For each model, the dominant choice percentage reflects the proportion of times the model selected the candidate in the \emph{position} (first or second) most often favored in gender-neutral comparisons (Order Bias), or the \emph{gender} (always female) more frequently selected across all trials (Gender Bias). Effect sizes (Cohen's $h$) measure the deviation from a 50\% baseline. $p$-values are based on two-sided binomial tests assessing whether the observed proportion differs significantly from 50\%. Asterisks denote significance thresholds: $^{*}p<.05$, $^{**}p<.01$, $^{***}p<.001$.}
\label{tab:bootstrappingold}

\end{table}

}

\begin{table}[ht]
\centering
\begin{tabular}{lcccc}

\textbf{Model} & \textbf{Bias Type} & \textbf{Cohen's \boldmath$h$} & \textbf{Mean \boldmath$\Delta h$} & \textbf{95\% CI} \\
\midrule
\multirow{2}{*}{GPT-4o-mini}
 & Order Bias  & 1.03 & \multirow{2}{*}{0.92} & \multirow{2}{*}{[0.82, 1.01]} \\
 & Gender Bias & 0.11 &                        &                                \\
\midrule
\multirow{2}{*}{Claude 3 Haiku}
 & Order Bias  & 0.76 & \multirow{2}{*}{0.58} & \multirow{2}{*}{[0.48, 0.67]} \\
 & Gender Bias & 0.18 &                        &                                \\
\midrule
\multirow{2}{*}{Llama 3 8B}
 & Order Bias  & 0.46 & \multirow{2}{*}{0.24} & \multirow{2}{*}{[0.15, 0.34]} \\
 & Gender Bias & 0.22 &                        &                                \\
\bottomrule
\end{tabular}

\betweentableandcap

\caption{
{\bf Aggregated order and gender bias.} For each model, Cohen's $h$ measures the effect size of order and gender bias (deviation from a 50\% baseline), aggregated across professions and tiers at \(T=1\). We also report the mean bootstrapped difference in effect sizes ($\Delta h = h_\text{order} - h_\text{gender}$) and its 95\% confidence interval, based on 10{,}000 resamples. In all cases, order bias significantly exceeds gender bias. 
}

\label{tab:bootstrapping}
\end{table}

\begin{table}[ht]
\centering
\begin{tabular}{llccccc}
\textbf{Model} & \textbf{Tier} & 
\makecell[c]{\textbf{Female when} \\ \textbf{first (\%)}} & 
\makecell[c]{\textbf{Male when} \\ \textbf{first (\%)}} & 
\textbf{Diff (\%)} & 
\textbf{\boldmath$p$-value} & 
\textbf{Cohen's \boldmath$h$} \\
\midrule
\multirow{4}{*}{GPT-4o-mini}
 & Best      & 100.0 & 97.7 & +2.3  & $0.247$        & 0.38 \\
 & Good      & 40.6  & 6.2  & +34.4 & ${\bf 3.9 \times 10^{-11}}^{***}$ & 0.88 \\
 & Mediocre  & 8.6   & 0.0  & +8.6  & ${\bf 7.8 \times 10^{-4}}^{***}$  & 0.60 \\
 & Weak      & 0.0   & 0.0  & +0.0  & $1.000$         & 0.000 \\
\midrule
\multirow{4}{*}{Claude 3 Haiku}
 & Best      & 75.8  & 33.6 & +42.2 & ${\bf 1.4 \times 10^{-11}}^{***}$ & 0.88 \\
 & Good      & 29.7  & 3.1  & +26.6 & ${\bf 4.0 \times 10^{-9}}^{***}$  & 0.80 \\
 & Mediocre  & 1.6   & 0.0  & +1.6  & $0.498$        & 0.25 \\
 & Weak      & 0.8   & 0.0  & +0.8  & $1.000$         & 0.18 \\
\midrule
\multirow{4}{*}{Llama 3 8B}
 & Best      & 93.8  & 73.4 & +20.3 & ${\bf 1.5 \times 10^{-5}}^{***}$  & 0.58 \\
 & Good      & 96.1  & 60.9 & +35.2 & ${\bf 1.5 \times 10^{-12}}^{***}$ & 0.95 \\
 & Mediocre  & 28.9  & 0.8  & +28.1 & ${\bf 1.6 \times 10^{-11}}^{***}$ & 0.96 \\
 & Weak      & 2.3   & 0.0  & +2.3  & $0.247$        & 0.31 \\
\bottomrule
\end{tabular}
\betweentableandcap
\caption{{\bf First-position selections by candidate gender.} “Female when first” and “Male when first” are the percentages of times the female or male candidate was chosen when appearing in position 1. Diff (\%) represents the difference in female and male selection rates. We report both statistical significance (using Fisher's Exact Test) and effect sizes (Cohen's $h$), which quantify the magnitude of differences in proportions. Asterisks indicate significance levels: $^{*}p < 0.05$, $^{**}p < 0.01$, $^{***}p < 0.001$.}
\label{tab:gender_diff_summary}
\end{table}

%% file: _z_appendix_additional_models.tex
\clearpage
\section{Additional Models}

\label{app:additional_models}

\subsection{Color Comparison Sets}\label{models:sets}

    
    

{To assess whether the observed positional effects generalize by other LLMs, we extended our pairwise and triplewise color comparison experiments to a diverse set of models from multiple LLM families, including OpenAI (GPT-4.1-nano), Anthropic (Claude Sonnet 4), Meta (Llama 4 Scout\footnote{In preliminary runs, Llama 4 Scout exhibited non-negligible variability at temperature 0. To ensure sufficiently deterministic evaluation, we fixed the random seed for all Llama 4 Scout experiments at $T=0$. After fixing the seed, the model produced stable responses under the deterministic comparison protocol used for the other models. Accordingly, references to $T=0$ for Llama 4 Scout in this appendix denote temperature-zero decoding with a fixed seed.}), Google (Gemini 2.5 Flash, Gemini 3 Flash), and Alibaba (Qwen 3 32B).}

\begin{table}[htbp]
\centering
\begin{tabular}{lllll}
\toprule
\textbf{AI Family} & \textbf{Model} & \textbf{Color 1} & \textbf{Color 2} & \textbf{Color 3} \\
\midrule

\multirow{4}{*}{OpenAI}


& \multirow{4}{*}{GPT-4.1-nano} 
& Frosted Cyan & Crystal Blue & Glacier Blue\\
& & Gentle Coral & Buttercream Yellow & Gentle Coral\\
& & Antique White & Ivory & Bone\\
& & Onyx & Obsidian & Smoky Black\\

\midrule

\multirow{4}{*}{Anthropic}


& \multirow{4}{*}{Claude Sonnet 4} 
& Robin's Egg Blue & Turquoise & Aqua\\
& & Buttercream Yellow & Lavender & Peach\\
& & Bone & Linen & Eggshell White\\
& & Onyx & Obsidian & Smoky Black\\

\midrule

\multirow{4}{*}{Meta}
& \multirow{4}{*}{Llama 4 Scout} 
& Light Blue & Sky Blue & Misty Blue \\
& & Peach & Honey Yellow & Light Yellow \\
& & Taupe & Dusty Beige & Bone \\
& & Burgundy & Crimson & Obsidian\\

\midrule

\multirow{8}{*}{Google}
& \multirow{4}{*}{Gemini 2.5 Flash} 
& Aqua Mist & Pale Turquoise & Crystal Blue\\
& & Gentle Coral & Buttercream Yellow & Warm Apricot\\
& & Eggshell White & Linen & Vanilla Beige\\
& & Onyx & Obsidian & Coal\\

\cmidrule(lr){2-5}

& \multirow{4}{*}{Gemini 3 Flash} 
& Aqua Mist & Pale Turquoise & Pale Butter Yellow\\
& & Soft Orchid & Rosewater & Warm Apricot \\
& & Parchment & Sand Dollar & Wheat\\
& & Onyx & Obsidian & Smoky Black\\

\midrule

\multirow{4}{*}{Alibaba}
& \multirow{4}{*}{Qwen 3 32b} 
& Soft Sky Blue & Glacier Blue & Aqua Mist\\
& & Gentle Coral & Buttercream Yellow & Peach Blossom\\
& & Cream & Linen & Alabaster\\
& & Onyx & Obsidian & Smoky Black\\

\bottomrule
\end{tabular}

\betweentableandcap
\caption{{\bf Colors used in triplewise and pairwise comparisons for the six additional models.} The three colors in each row were used for triplewise comparisons for each model and tier.  Colors 1 and 2 were used for pairwise comparisons.}
\label{table:colorsets_all}

\end{table}

\clearpage
\subsection{Pairwise Color Comparisons}\label{models:pair}
{We used Colors 1 and 2 from \autoref{table:colorsets_all} for each model and tier to conduct pairwise color comparisons. \autoref{tab:pairwise_temp1_allModels_appendix} reports the results at T = 1, showing the proportion of times each position was
chosen, along with effect sizes (Cohen's h) and associated p-values. Effect sizes are measured against a 50\% baseline, and p-values reflect the significance of deviation based on a two-sided binomial test. Even at the highest temperature setting used, most comparisons reveal statistically significant and substantial position biases. We do not perform statistical tests for T = 0 as the results are almost deterministic.}

\begin{table}[ht]
\centering
\begin{tabular}{llccc}
\textbf{Model} & \textbf{Tier} & \textbf{First / Second (\%)} & \textbf{Cohen's \boldmath$h$} & \textbf{\boldmath$p$-value} \\
\midrule
\multirow{4}{*}{GPT-4.1-nano}
 & {\Bestcolor}  & 88 / 12 & 0.86 & ${\bf 1.9 \times 10^{-15}}^{***}$ \\
 & {\Goodcolor}  & 93 / 7  & 1.04 & ${\bf 2.7 \times 10^{-20}}^{***}$ \\
 & {\Medcolor}   & 91 / 9  & 0.96 & ${\bf 3.3 \times 10^{-18}}^{***}$ \\
 & {\Worstcolor} & 24 / 76 & 0.55 & ${\bf 1.8 \times 10^{-7}}^{***}$ \\
\midrule
\multirow{4}{*}{Claude Sonnet 4}
 & {\Bestcolor}  & 84 / 16 & 0.75 & ${\bf 2.6 \times 10^{-12}}^{***}$ \\
 & {\Goodcolor}  & 11 / 89 & 0.90 & ${\bf 2.5 \times 10^{-16}}^{***}$ \\
 & {\Medcolor}   & 25 / 75 & 0.52 & ${\bf 5.6 \times 10^{-7}}^{***}$ \\
 & {\Worstcolor} & 1 / 99  & 1.37 & ${\bf 1.6 \times 10^{-28}}^{***}$ \\
\midrule
\multirow{4}{*}{Gemini 2.5 Flash}
 & {\Bestcolor}  & 97 / 3  & 1.22 & ${\bf 2.6 \times 10^{-25}}^{***}$ \\
 & {\Goodcolor}  & 55 / 45 & 0.10 & $0.37$ \\
 & {\Medcolor}   & 87 / 13 & 0.83 & ${\bf 1.3 \times 10^{-14}}^{***}$ \\
 & {\Worstcolor} & 67 / 33 & 0.35 & ${\bf 8.7 \times 10^{-4}}^{***}$ \\
\midrule
\multirow{4}{*}{Gemini 3 Flash}
 & {\Bestcolor}  & 86 / 14 & 0.80 & ${\bf 8.3 \times 10^{-14}}^{***}$ \\
 & {\Goodcolor}  & 48 / 52 & 0.04 & $0.76$ \\
 & {\Medcolor}   & 46 / 54 & 0.08 & $0.48$ \\
 & {\Worstcolor} & 7 / 93  & 1.04 & ${\bf 2.7 \times 10^{-20}}^{***}$ \\
\midrule
\multirow{4}{*}{Llama 4 Scout}
 & {\Bestcolor}  & 4 / 96 & 1.17 & ${\bf 6.5 \times 10^{-24}}^{***}$ \\
 & {\Goodcolor}  & 0 / 100 & 1.57 & ${\bf 1.6 \times 10^{-30}}^{***}$ \\
 & {\Medcolor}   & 1 / 99  & 1.37 & ${\bf 1.6 \times 10^{-28}}^{***}$ \\
 & {\Worstcolor} & 0 / 100 & 1.57 & ${\bf 1.6 \times 10^{-30}}^{***}$ \\
\midrule
\multirow{4}{*}{Qwen 3 32B}
 & {\Bestcolor}  & 74 / 26 & 0.50 & ${\bf 1.7 \times 10^{-6}}^{***}$ \\
 & {\Goodcolor}  & 69 / 31 & 0.39 & ${\bf 1.8 \times 10^{-4}}^{***}$ \\
 & {\Medcolor}   & 29 / 71 & 0.43 & ${\bf 3.2 \times 10^{-5}}^{***}$ \\
 & {\Worstcolor} & 22 / 78 & 0.59 & ${\bf 1.6 \times 10^{-8}}^{***}$ \\
\bottomrule
\end{tabular}
\betweentableandcap
\caption{{\bf Position bias in pairwise color comparisons, \boldmath\(T=1\).} Proportions of first and second selections across quality tiers. Cohen’s \(h\) quantifies effect sizes from a 50\% baseline. Asterisks indicate significance: ${\bf p < 0.001}^{***}$.}

\label{tab:pairwise_temp1_allModels_appendix}

\end{table}

\begin{figure}[ht]
    \centering
    \includegraphics[width=0.95\linewidth]{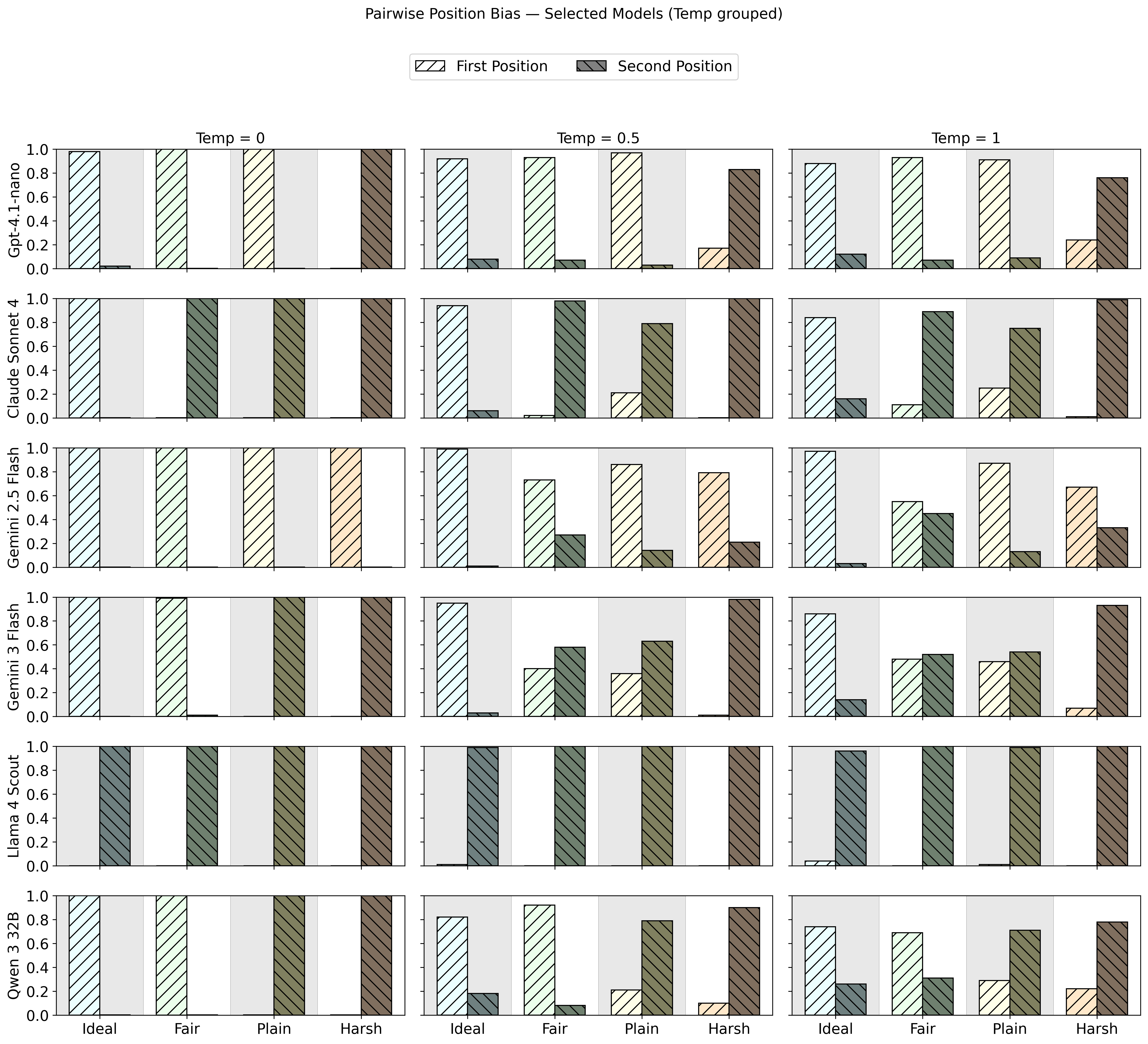}
    
    \caption{{\bf Order effects in pairwise comparisons at three temperature settings.} The figure shows the percentage of selections by position (first or second), broken down by color tier for the model across temperatures \(T=0\), \(0.5\), and \(1\) (left to right).}
    
    \label{fig:pairwise_all_models}
\end{figure}

\clearpage
{\autoref{fig:pairwise_all_models} shows the results at different temperatures (\(T=0, 0.5,\) and \(1\)). At  \(T=0\) (the topmost plot), most models strongly prefer the first option for high-quality color sets, and shift to the second position for lower-quality sets. For instance, Gemini 3 Flash and Qwen 3 32B displayed a primacy effect in the top two tiers (Ideal and Fair) and a recency effect in the bottom two (Plain and Harsh). Claude 4 Sonnet showed a primacy effect in the top tier, and a recency effect in the bottom three. Gemini 2.5 Flash displayed a primacy effect in all four tiers, while Llama 4 Scout displayed a recency effect in all four tiers. Thus, even models that do not exhibit a strict tier-by-tier reversal are consistent with the same quality-dependent shift in positional preference.

While higher temperatures introduce randomness, the position bias remains robust across models and tiers.}

\clearpage
\subsection{Triplewise Color Comparisons}\label{models:trip}
For triplewise comparisons, we used all three colors from \autoref{table:colorsets_all}. \autoref{tab:threeway_temp1_allModels} shows the results of the triplewise comparisons at \(T=1\), showing the proportion of times each position was chosen, along with effect sizes and associated \(p\)-values.

\begin{figure}[ht]
    \centering
    \includegraphics[width=0.95\linewidth]{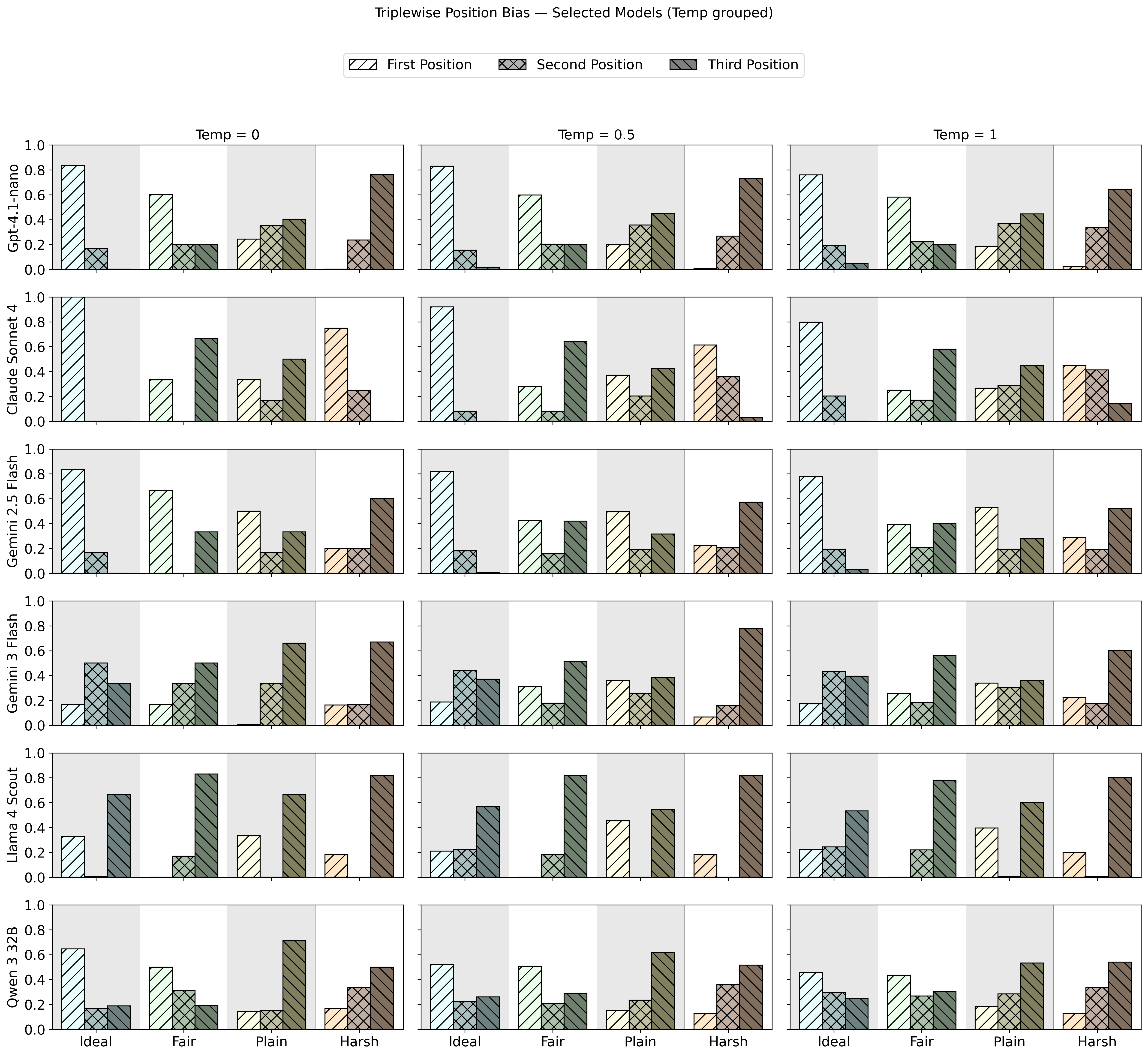}
    
    \caption{{\bf Order effects in triplewise comparisons at three temperature settings.} The figure shows the percentage of selections by position (first, second, or third), broken down by color tier for the Claude model across temperatures \(T=0\), \(0.5\), and \(1\) (left to right).}
    
    \label{fig:triplewise_allModels}
\end{figure}

\autoref{fig:triplewise_allModels} shows the results at different temperatures. Similarly to the pairwise comparisons, while higher temperatures introduce randomness, the position bias remains robust.

\begin{table}[ht]
\centering
\begin{tabular}{llccc}
\textbf{Model} & \textbf{Tier} & \textbf{First / Second / Third (\%)} & \textbf{Cramer's \boldmath$V$} & \textbf{\boldmath$p$-value} \\
\midrule
\multirow{4}{*}{GPT-4.1-nano}
 & {\Bestcolor}  & 76 / 19 / 5  & 0.65 & ${\bf 3.40 \times 10^{-56}}^{***}$ \\
 & {\Goodcolor}  & 58 / 22 / 20 & 0.37 & ${\bf 1.30 \times 10^{-29}}^{***}$ \\
 & {\Medcolor}   & 19 / 37 / 44 & 0.23 & ${\bf 2.10 \times 10^{-7}}^{***}$ \\
 & {\Worstcolor} & 2 / 34 / 64  & 0.54 & ${\bf 1.53 \times 10^{-38}}^{***}$ \\
\midrule
\multirow{4}{*}{Claude Sonnet 4}
 & {\Bestcolor}  & 80 / 20 / 0  & 0.72 & ${\bf 1.07 \times 10^{-67}}^{***}$ \\
 & {\Goodcolor}  & 25 / 17 / 58 & 0.38 & ${\bf 3.45 \times 10^{-19}}^{***}$ \\
 & {\Medcolor}   & 26 / 29 / 45 & 0.17 & ${\bf 1.57 \times 10^{-4}}^{***}$ \\
 & {\Worstcolor} & 45 / 41 / 14 & 0.29 & ${\bf 4.16 \times 10^{-7}}^{***}$ \\
\midrule
\multirow{4}{*}{Gemini 2.5 Flash}
 & {\Bestcolor}  & 78 / 19 / 3  & 0.68 & ${\bf 5.98 \times 10^{-61}}^{***}$ \\
 & {\Goodcolor}  & 39 / 21 / 40 & 0.19 & ${\bf 1.96 \times 10^{-5}}^{***}$ \\
 & {\Medcolor}   & 53 / 19 / 28 & 0.30 & ${\bf 9.62 \times 10^{-13}}^{***}$ \\
 & {\Worstcolor} & 29 / 19 / 52 & 0.30 & ${\bf 1.78 \times 10^{-19}}^{***}$ \\
\midrule
\multirow{4}{*}{Gemini 3 Flash}
 & {\Bestcolor}  & 17 / 43 / 40 & 0.24 & ${\bf 2.32 \times 10^{-8}}^{***}$ \\
 & {\Goodcolor}  & 26 / 18 / 56 & 0.35 & ${\bf 1.80 \times 10^{-16}}^{***}$ \\
 & {\Medcolor}   & 34 / 30 / 36 & 0.05 & $4.73 \times 10^{-1}$ \\
 & {\Worstcolor} & 22 / 18 / 60 & 0.41 & ${\bf 5.35 \times 10^{-22}}^{***}$ \\
\midrule
\multirow{4}{*}{Llama 4 Scout}
 & {\Bestcolor}  & 22 / 24 / 53 & 0.30 & ${\bf 1.72 \times 10^{-12}}^{***}$ \\
 & {\Goodcolor}  & 0 / 22 / 78  & 0.70 & ${\bf 6.08 \times 10^{-64}}^{***}$ \\
 & {\Medcolor}   & 40 / 0 / 60  & 0.53 & ${\bf 1.09 \times 10^{-36}}^{***}$ \\
 & {\Worstcolor} & 20 / 0 / 80  & 0.72 & ${\bf 3.21 \times 10^{-68}}^{***}$ \\
\midrule
\multirow{4}{*}{Qwen 3 32B}
 & {\Bestcolor}  & 46 / 30 / 24 & 0.19 & ${\bf 1.98 \times 10^{-5}}^{***}$ \\
 & {\Goodcolor}  & 43 / 27 / 30 & 0.15 & ${\bf 9.12 \times 10^{-4}}^{***}$ \\
 & {\Medcolor}   & 18 / 28 / 54 & 0.31 & ${\bf 1.98 \times 10^{-13}}^{***}$ \\
 & {\Worstcolor} & 13 / 33 / 54 & 0.36 & ${\bf 2.02 \times 10^{-17}}^{***}$ \\
\bottomrule
\end{tabular}
\betweentableandcap

\caption{{\bf Position bias in three-way comparisons, \boldmath\(T=1\).} Proportions of first, second, and third selections across quality tiers. Cramer's \(V\) quantifies effect sizes. Asterisks indicate significance: ${\bf p < 0.001}^{***}$.}

\label{tab:threeway_temp1_allModels}
\end{table}

%% file: _z_appendix_additional_languages.tex
\clearpage
\section{Additional Languages}

\label{app:additional_languages}

\subsection{Prompts and Color Sets}\label{lang:sets}

The prompts used for pairwise comparisons in Mandarin Chinese and Hebrew were the following.

\ifpnas {
\begin{lstlisting}
prompt_Chinese = [
    f"哪种颜色最适合儿童房：{first} 还是 {second}？请只回答一个颜色，不要使用任何标点符号或解释。"
]

and

prompt_Hebrew = [
    f"?{first}(*@\texthebrew{ או }@*){second}(*@\texthebrew{איזה צבע הכי מתאים לחדר ילדים: }@*)
    (*@\texthebrew{ אנא השב עם צבע אחד בלבד וללא סימני פיסוק או הסבר.}@*)"
]

\end{lstlisting}
}
\else {
\begin{figure}[ht]
    \centering
    \includegraphics[width=0.95\linewidth]{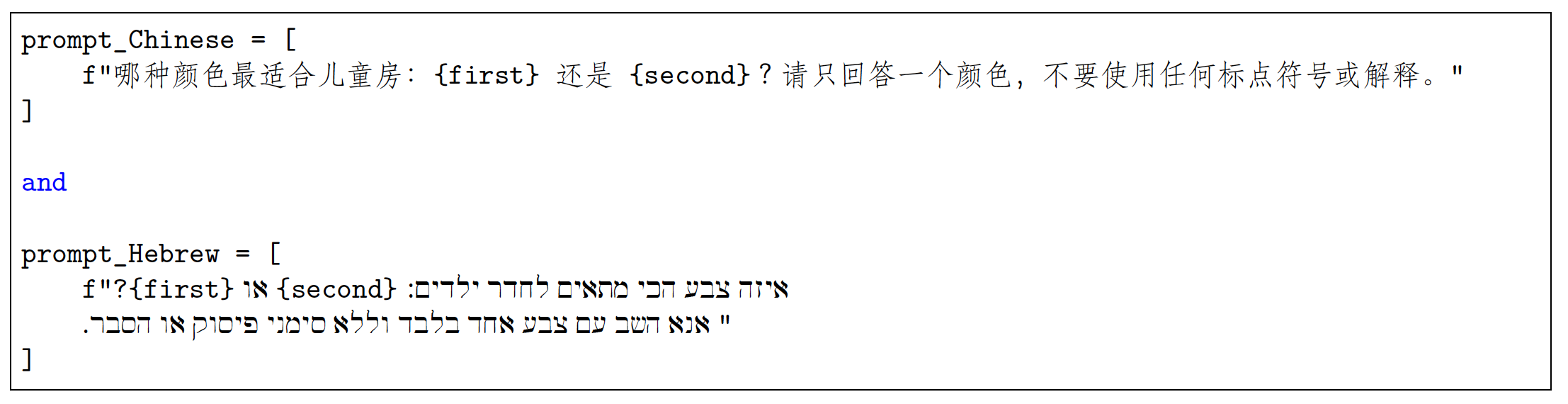}
\end{figure}
}
\fi

The prompts used for triplewise comparisons were similar and are omitted.
\ifpnas{\autoref{tab:color_examples_translated}}\else{\autoref{fig:color_table} }\fi lists the specific colors used for GPT-4o-mini for the Chinese and Hebrew prompts.

\ifpnas {
\begin{table}[htbp]
\centering
\small
\begin{tabular}{lllll}
\toprule
\textbf{Language} & \textbf{Tier} & \textbf{Color 1} & \textbf{Color 2} & \textbf{Color 3} \\
\midrule

\multirow{4}{*}{Chinese} 
& Ideal 
& 冰川蓝 {\footnotesize (Glacier Blue)} 
& 水雾蓝 {\footnotesize (Mist Blue)} 
& 浅绿松石色 {\footnotesize (Light Turquoise)} \\
& Fair 
& 柔和珊瑚色 {\footnotesize (Soft Coral)} 
& 淡紫色 {\footnotesize (Light Purple)} 
& 桃花色 {\footnotesize (Peach Blossom)} \\
& Plain 
& 古董白 {\footnotesize (Antique White)} 
& 羊皮纸色 {\footnotesize (Parchment)} 
& 蛋壳白 {\footnotesize (Eggshell White)} \\
& Harsh 
& 黑曜石色 {\footnotesize (Obsidian)} 
& 灰黑色 {\footnotesize (Gray-Black)} 
& 烟熏黑 {\footnotesize (Smoky Black)} \\

\cmidrule(lr){1-5}

\multirow{4}{*}{Hebrew} 
& Ideal 
& \texthebrew{קורל עדין} {\footnotesize (Gentle  Coral)} 
& \texthebrew{צהוב קרם חמאה} {\footnotesize (Buttercream Yellow)} 
& \texthebrew{לילך} {\footnotesize (Lilac)} \\
& Fair 
& \texthebrew{ציאן בהיר} {\footnotesize (Light Cyan)} 
& \texthebrew{כחול קרחוני} {\footnotesize (Glacier Blue)} 
& \texthebrew{כחול קריר} {\footnotesize (Cool Blue)} \\
& Plain 
& \texthebrew{לבן קליפת ביצה} {\footnotesize (Eggshell White)} 
& \texthebrew{לבן עתיק} {\footnotesize (Antique White)} 
& \texthebrew{פשתן} {\footnotesize (Linen)} \\
& Harsh 
& \texthebrew{פחם} {\footnotesize (Charcoal)} 
& \texthebrew{שחור מעושן} {\footnotesize (Smoky Black)} 
& \texthebrew{שחור אפר} {\footnotesize (Ash Black)} \\

\bottomrule
\end{tabular}
\caption{Colors used for Chinese and Hebrew prompts, with English translations.}
\label{tab:color_examples_translated}
\end{table}
}
\else {
\begin{figure}[ht]
    \centering
    \includegraphics[width=0.95\linewidth]{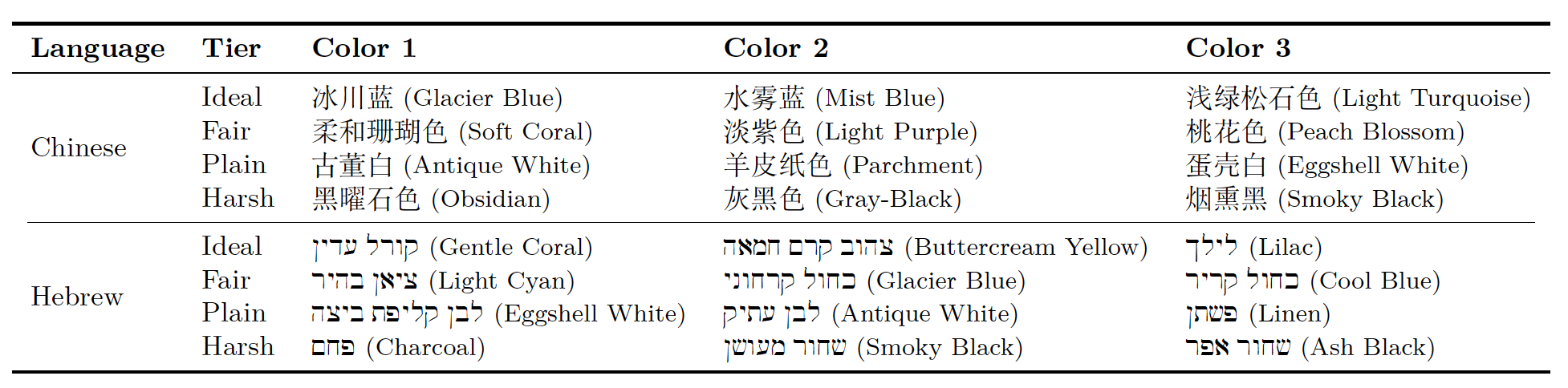} 
    \caption{\bf Colors used for Chinese and Hebrew prompts, with English translations.}
    \label{fig:color_table}
\end{figure}
}
\fi

\clearpage
\subsection{Pairwise Comparisons}\label{lang:pair}
{\ifpnas{We used Colors 1 and 2 from~\autoref{tab:color_examples_translated} for each model and tier to conduct pairwise color comparisons.} \else{}\fi

\autoref{tab:pairwise_temp1_language} reports the results at  \(T=1\), showing the proportion of times each position was chosen, along with effect sizes (Cohen’s~\(h\)) and associated \(p\)-values across two languages --- Mandarin Chinese and Hebrew. Chinese is read from left to right, whereas Hebrew is read from right to left. The results indicate that GPT-4o-mini tends to prefer the first option when the response quality is relatively high (e.g., Ideal), but shifts its preference toward the last option as the quality decreases (e.g., Plain and Harsh). This pattern is consistent with the trend observed in English.

}

\begin{table}[htbp]
\centering
\begin{tabular}{lllll}
\toprule
\textbf{Language} & \textbf{Tier} & \textbf{First/Second (\%)} & \textbf{Cohen's h} & \textbf{p-value} \\
\midrule

\multirow{4}{*}{Chinese} 
& Ideal & 86 / 14 & 0.804 & ${\bf 8.28 \times 10^{-14}}^{***}$ \\
& Fair  & 81 / 19 & 0.669 & ${\bf 2.70 \times 10^{-10}}^{***}$ \\
& Plain & 30 / 70 & 0.412 & ${\bf 7.85 \times 10^{-5}}^{***}$ \\
& Harsh & 9 / 91  & 0.961 & ${\bf 3.32 \times 10^{-18}}^{***}$ \\

\cmidrule(lr){1-5}

\multirow{4}{*}{Hebrew} 
& Ideal & 86 / 14 & 0.804 & ${\bf 8.28 \times 10^{-14}}^{***}$ \\
& Fair  & 30 / 70 & 0.412 & ${\bf 7.85 \times 10^{-5}}^{***}$ \\
& Plain & 3 / 97  & 1.223 & ${\bf 2.63 \times 10^{-25}}^{***}$ \\
& Harsh & 34 / 66 & 0.326 & ${\bf 1.79 \times 10^{-3}}^{**}$ \\

\bottomrule
\end{tabular}
\caption{{\bf Order effects of colors in pairwise comparisons at T = 1.} \\
Top: Chinese; Bottom: Hebrew. Model: GPT-4o-mini.}
\label{tab:pairwise_temp1_language}
\end{table}

\begin{figure}[ht]
    \centering
    \includegraphics[width=0.95\linewidth]{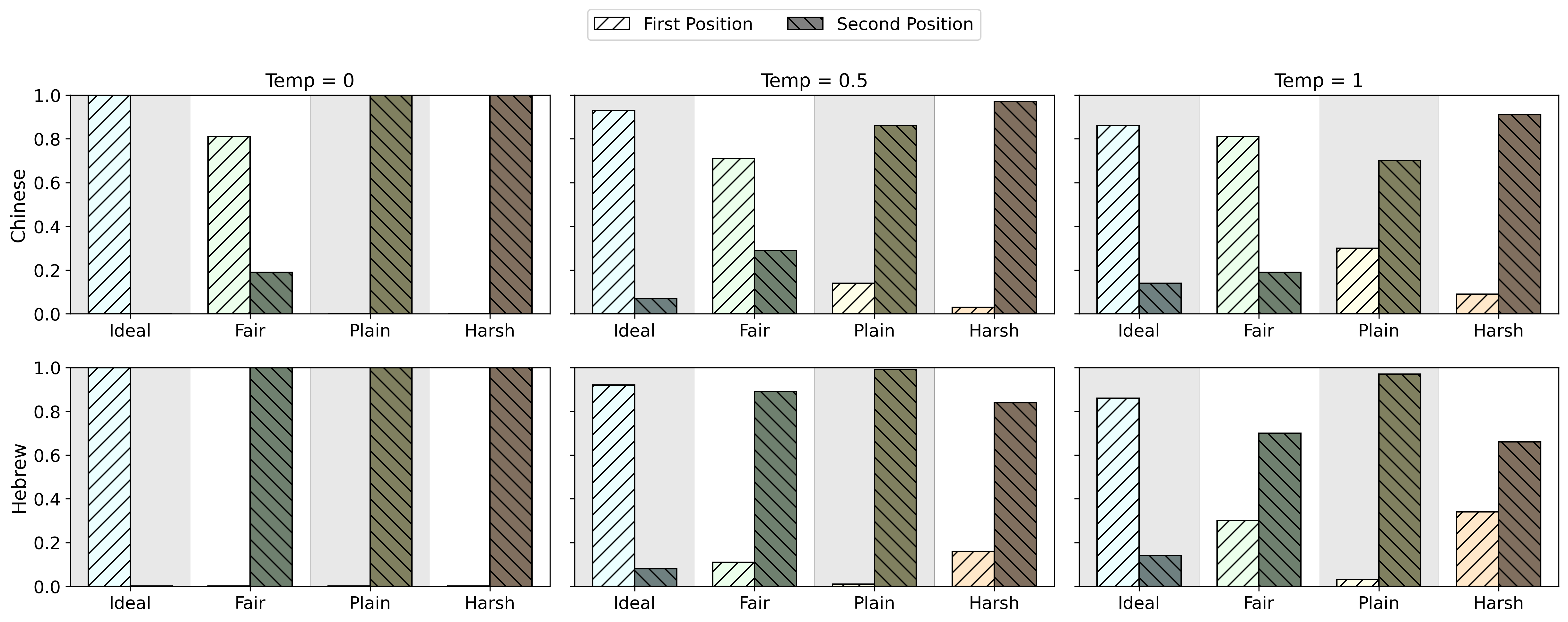}
    
    \caption{{\bf Order effects in pairwise color comparisons across languages.}
    The figure shows the proportion of selections by position (first, second, or third), broken down by color tier for GPT-4o-mini at temperatures \(T=0\), \(0.5\), and \(1\).\\
    Top: Chinese; Bottom: Hebrew.}

    \label{fig:pairwise_combined_languages}
\end{figure}

{\autoref{fig:pairwise_combined_languages} shows the results at different temperatures (\(T=0, 0.5,\) and \(1\)). At  \(T=0\) (the topmost plot), same to the \autoref{tab:pairwise_temp1_language}, all models strongly prefer the first option for high-quality color sets, and shift to the second position for lower-quality sets. For instance, GPT-4o-mini displayed a primacy effect in the top two tiers (Ideal and Fair) and a recency effect in the bottom two (Plain and Harsh) in the Madarian Chinese setting. Also, it showed a primacy effect in the top one tier, and a recency effect in the bottom three tiers with the Hebrew context. 

While higher temperatures introduce randomness, the position bias remains robust across models and tiers.}

\clearpage
\subsection{Triplewise Comparisons}\label{lang:trip}
\begin{figure}[ht]
    \centering
    \includegraphics[width=0.95\linewidth]{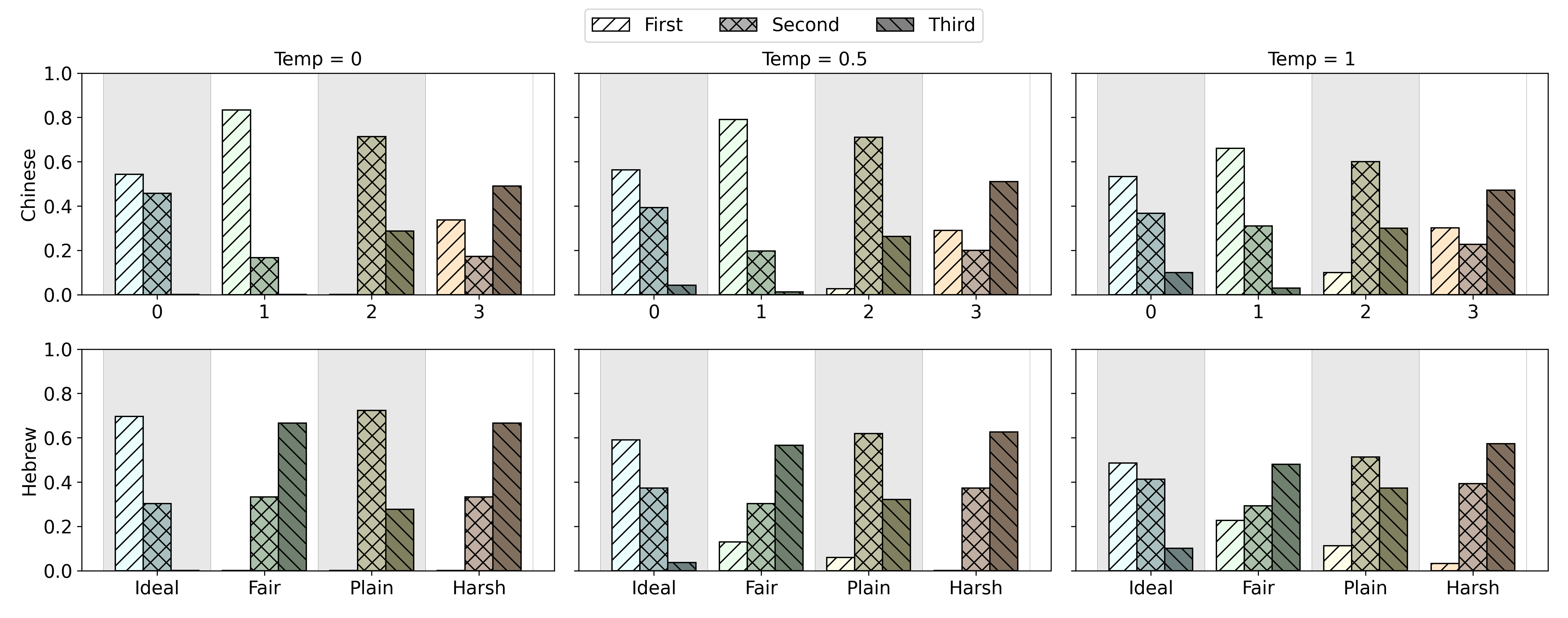}

    \caption{{\bf Order effects in triplewise color comparisons across languages.}
    The figure shows the proportion of selections by position (first, second, or third), broken down by color tier for GPT-4o-mini at temperatures \(T=0\), \(0.5\), and \(1\).
    Top: Chinese; Bottom: Hebrew.}

    \label{fig:triplewise_combined_language}

\end{figure}

{\ifpnas{For triplewise comparisons, we used all three colors from \autoref{tab:color_examples_translated}.}\else{}\fi \autoref{tab:triplewise_temp1_language} shows the results of the triplewise comparisons at \(T=1\), showing the proportion of times each position was chosen, along with effect sizes and associated \(p\)-values. Results shows a clear shift in LLM's preference from the first position to the last as quality decreases.

\autoref{fig:triplewise_combined_language} shows the results at different temperatures. Similarly to the pairwise comparisons, while higher temperatures introduce randomness, the position bias remains robust.
}

\begin{table}[htbp]
\centering
\begin{tabular}{lllll}
\toprule
\textbf{Language} & \textbf{Tier} & \textbf{First/Second/Third (\%)} & \textbf{Cramer's V} & \textbf{p-value} \\
\midrule

\multirow{4}{*}{Chinese} 
& Ideal & 53 / 37 / 10 & 0.379 & ${\bf 2.12 \times 10^{-19}}^{***}$ \\
& Fair  & 66 / 31 / 3  & 0.547 & ${\bf 1.14 \times 10^{-39}}^{***}$ \\
& Plain & 10 / 60 / 30 & 0.436 & ${\bf 1.76 \times 10^{-25}}^{***}$ \\
& Harsh & 30 / 23 / 47 & 0.217 & ${\bf 7.75 \times 10^{-7}}^{***}$ \\

\cmidrule(lr){1-5}

\multirow{4}{*}{Hebrew} 
& Ideal & 49 / 41 / 10 & 0.355 & ${\bf 5.11 \times 10^{-17}}^{***}$ \\
& Fair  & 23 / 29 / 48 & 0.227 & ${\bf 1.82 \times 10^{-7}}^{***}$ \\
& Plain & 11 / 51 / 38 & 0.352 & ${\bf 7.88 \times 10^{-17}}^{***}$ \\
& Harsh & 4 / 39 / 57  & 0.476 & ${\bf 2.82 \times 10^{-30}}^{***}$ \\

\bottomrule
\end{tabular}
\caption{{\bf Order effects of colors in triplewise comparisons at T = 1.} 
Top: Chinese; Bottom: Hebrew. Model: GPT-4o-mini.}
\label{tab:triplewise_temp1_language}
\end{table}

